%% file: main.tex
\documentclass{meta_assets/fairmeta}
\usepackage{amsmath,amssymb,amsthm,mathtools}

\usepackage{pgfplots}
\usepgfplotslibrary{groupplots,fillbetween}
\usepackage{pgfplotstable}
\usetikzlibrary{calc,shapes.geometric}
\usepackage{array}
\usepackage{listings}

\pgfplotsset{
  compat=1.16,
  every axis/.append style={
    font=\footnotesize,
    line width=0.7pt,
    tick style={line width=0.6pt},
  },
  legend style={draw=none,fill=none,font=\scriptsize},
}

\theoremstyle{plain}

\theoremstyle{definition}

\theoremstyle{remark}

\newcommand{\swatch}[2]{{\setlength{\fboxsep}{0pt}\setlength{\fboxrule}{0.3pt}%
  \fcolorbox{#1}{#2}{\rule{1.7mm}{0pt}\rule{0pt}{1.5mm}}}}

\definecolor{cSteps}{RGB}{214,110,20}   % fewer denoising steps
\definecolor{cAttn}{RGB}{31,90,160}     % sparse attention / fewer tokens
\definecolor{cVae}{RGB}{130,50,150}     % replacement decoder
\definecolor{cOne}{RGB}{198,18,48}      % distilled for one step
\definecolor{cFlex}{RGB}{0,130,118}     % chunk-schedule search
\definecolor{cNv}{RGB}{190,30,120}      % NVFP4
\definecolor{darkgreen}{RGB}{0,185,100}  % cross-reference green
\hypersetup{linkcolor=darkgreen}

\DeclareFontShape{T1}{optimistic}{b}{n}{<-> s * [0.88] meta_assets/optimistic}{}
\DeclareFontShape{T1}{optimistic}{bx}{n}{<-> s * [0.88] meta_assets/optimistic}{}
\DeclareFontShape{T1}{optimistic}{b}{sc}{<-> s * [0.88] meta_assets/optimistic}{}
\DeclareFontShape{T1}{optimistic}{bx}{sc}{<-> s * [0.88] meta_assets/optimistic}{}

\def\showtheory{1}

\newcommand{\iftheory}[1]{\ifnum\showtheory=1 #1\fi}

\newcommand{\singlequalstrip}[1]{%
  \def\singlequalfile{#1}\includegraphics[width=0.355\linewidth]\singlequalfile%
}
\newcommand{\singlequaltimes}{%
  \makebox[0.355\linewidth][s]{%
    \makebox[0.116\linewidth][c]{\scriptsize 0 s}\hfill
    \makebox[0.116\linewidth][c]{\scriptsize 2.5 s}\hfill
    \makebox[0.116\linewidth][c]{\scriptsize 5 s}}%
}
\newcommand{\singlequalleftlabel}[1]{%
  \raisebox{0.03405\linewidth}{\raisebox{-0.5\height}{\small\shortstack[r]{#1}}}%
}
\newcommand{\singlequalrightlabel}[1]{%
  \raisebox{0.03405\linewidth}{\raisebox{-0.5\height}{\small\shortstack[l]{#1}}}%
}
\newcommand{\pairedqualcomparison}[8]{%
  \begin{tabular}{@{}r@{\hspace{3pt}}c@{\hspace{5pt}{\color{black!45}\vrule width 0.4pt}\hspace{5pt}}c@{\hspace{3pt}}l@{}}
  & \singlequaltimes & \singlequaltimes & \\
  \singlequalleftlabel{Self Forcing\\17.0 FPS}
      & \singlequalstrip{#1} & \singlequalstrip{#3}
      & \singlequalrightlabel{Causal Forcing\\17.0 FPS} \\[1pt]
  \singlequalleftlabel{SF $+$ \textsc{UnStep}\\49.8 FPS}
      & \singlequalstrip{#2} & \singlequalstrip{#4}
      & \singlequalrightlabel{CF $+$ \textsc{UnStep}\\49.8 FPS} \\[6pt]
  \singlequalleftlabel{Self Forcing\\17.0 FPS}
      & \singlequalstrip{#5} & \singlequalstrip{#7}
      & \singlequalrightlabel{Causal Forcing\\17.0 FPS} \\[1pt]
  \singlequalleftlabel{SF $+$ \textsc{UnStep}\\49.8 FPS}
      & \singlequalstrip{#6} & \singlequalstrip{#8}
      & \singlequalrightlabel{CF $+$ \textsc{UnStep}\\49.8 FPS}
  \end{tabular}%
}

\title{UnStep: Training-Free Acceleration of Causal Video Diffusion
with Fewer Steps Than Distillation}

\author[]{Youssef Mansour}
\author[]{Enis Simsar}
\author[]{Fadime Sener}
\author[]{Markos Georgopoulos}
\author[]{Albert Pumarola}
\author[]{Ali Thabet}
\author[]{Edgar Schoenfeld}

\affiliation[]{Meta Superintelligence Labs}

\abstract{\input{sections/abstract}}

\metadata[Project page]{\url{https://y-mansour.github.io/UnStep/}}
\metadata[Code]{\url{https://github.com/facebookresearch/UnStep}}

\begin{document}

\maketitle

\input{sections/intro}

\input{sections/related_work}

\input{sections/method}

\input{sections/experiments}

\iftheory{\input{sections/theory}}

\input{sections/conclusion}

%\clearpage
\bibliography{references}

\clearpage
\beginappendix
\input{sections/appendix/implementation}
\input{sections/appendix/evaluation}
\input{sections/appendix/extended_results}
\input{sections/appendix/additional_related_work}
\iftheory{\input{sections/appendix/proofs}}
\input{sections/appendix/visualizations}

\end{document}

%% file: sections/abstract.tex
Distilling bidirectional multi-step video diffusion transformers into few-step causal models has become a common approach for streaming video generation. While these few-step students are significantly faster than the teachers they are distilled from, they remain slow for real-time generation. In this work we present \textsc{UnStep}, a training-free wrapper that accelerates few-step causal video models at inference by running them with fewer diffusion transformer (DiT) steps than during distillation and limiting the temporal window retained in the attention KV cache. We propose two inference-only mechanisms to recover quality lost by step reduction and attention windowing: renoising the generated latent frames to a near clean level and reusing the existing clean-cache pass to refine them, and applying truncated SVD to the DiT attention value and output projections. We also accelerate inference with a quality-preserving runtime stack for the DiT and VAE decoder, including more efficient attention calls and KV indexing, fused Triton RoPE with cached coefficients, and VAE decoding with optimized memory layout, precision, and convolution kernels. By reducing computation and optimizing the runtime stack, \textsc{UnStep} sets a new throughput regime for causal video diffusion, by running substantially faster than current methods, reaching 50 FPS on a single H100 without quality loss, and 77 FPS on GB200, all without retraining.

%\textsc{UnStep} is faster than models explicitly trained with reduced steps or limited cached windows, and enables current state-of-the-art causal video diffusion models to stream at over 49 FPS on a single H100 with no quality loss and up to 77 FPS on GB200, all without retraining.

%% file: sections/intro.tex
\section{Introduction}
\label{sec:intro}

%Wan and bidirectional generation
Video diffusion models are now capable of generating high-quality clips, but their computational cost at inference remains a major bottleneck. Wan~2.1 \citep{wan2025} has emerged as a powerful open video foundation model capable of producing videos with significantly higher fidelity than earlier generative video models \citep{ho2022videodiffusion,blattmann2023alignlatents}. Wan employs a bidirectional diffusion transformer \citep{peebles2023dit} that attends over the full clip, so no frame can be emitted until the entire clip has been generated. It also requires 50 denoising steps to generate the video. As a result, it is not suitable for real-time or streaming use cases. 

%Causal distillation
Causal distillation addresses the limitations of multi-step bidirectional models by reducing the number of steps to only a few (e.g., four steps) and converting the bidirectional attention to causal attention. This allows frames to be emitted as they are generated, analogous to autoregressive generation in language models. CausVid \citep{yin2025causvid} is a prominent example of causal distillation: it produces a fast streaming model, but with a notable quality drop relative to the teacher it was distilled from. 

%Self Forcing
Self Forcing \citep{huang2025selfforcing} introduces a new paradigm for causal diffusion distillation. It enables a 4-step causal student to match the performance of the bidirectional 50-step Wan 2.1 teacher \citep{wan2025} by distilling the student on its own rollouts. Self Forcing has therefore become a standard distillation technique in causal diffusion, paving the way for later work to improve quality \citep{zhu2026causalforcing,distillalign}, extend generation to long clips \citep{longlive,rollingforcing}, add interactive control \citep{motionstream}, and make inference faster \citep{helios,zheng2026causalrcm}.

%Other speed attempts
Self Forcing streams at 17 FPS on an H100. To improve throughput, several works distill to fewer denoising steps (e.g. 1 or 2 steps) \citep{oneforcing2026,zhao2026causalforcingpp,zheng2026causalrcm}, use a compact VAE decoder \citep{diagdistill, flashvaed}, or train with limited KV attention windows \citep{lv2026lightforcing, xu2026sparseforcing}. The extra speed usually comes at the cost of quality, and as these methods require training, the speed quality tradeoff is determined during distillation.

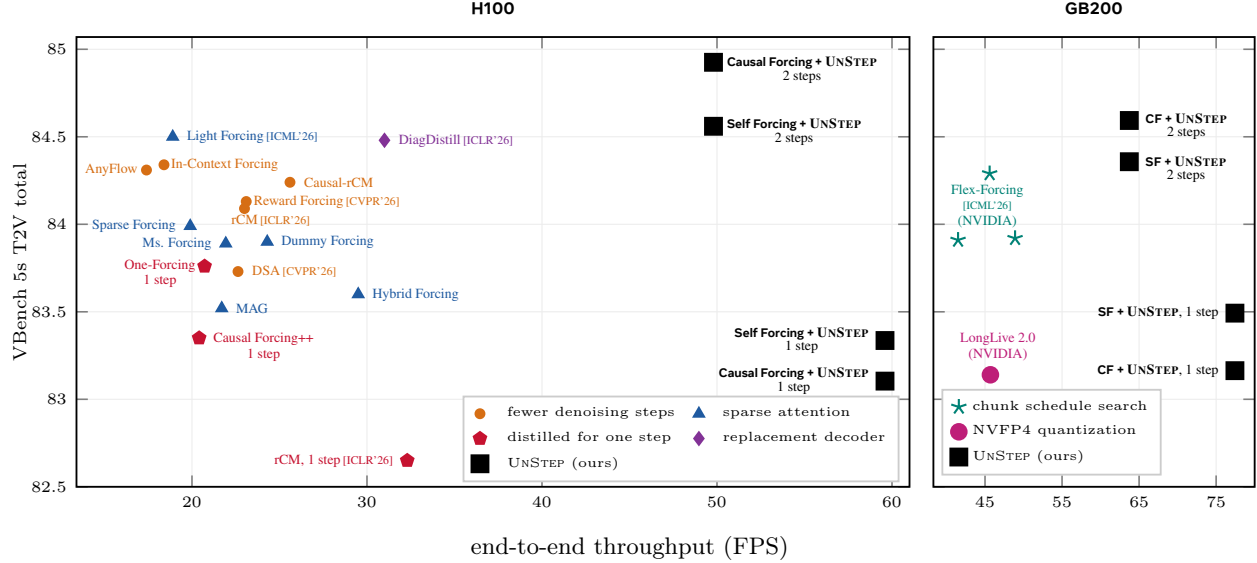
\begin{figure}[t]
\centering
\begin{tikzpicture}
% Filled five-point star for UnStep; the standard PGF "star" is an asterisk.
\pgfdeclareplotmark{unstepstar}{%
  \pgfpathmoveto{\pgfqpointpolar{90}{\pgfplotmarksize}}%
  \pgfpathlineto{\pgfqpointpolar{54}{0.382\pgfplotmarksize}}%
  \pgfpathlineto{\pgfqpointpolar{18}{\pgfplotmarksize}}%
  \pgfpathlineto{\pgfqpointpolar{-18}{0.382\pgfplotmarksize}}%
  \pgfpathlineto{\pgfqpointpolar{-54}{\pgfplotmarksize}}%
  \pgfpathlineto{\pgfqpointpolar{-90}{0.382\pgfplotmarksize}}%
  \pgfpathlineto{\pgfqpointpolar{-126}{\pgfplotmarksize}}%
  \pgfpathlineto{\pgfqpointpolar{-162}{0.382\pgfplotmarksize}}%
  \pgfpathlineto{\pgfqpointpolar{-198}{\pgfplotmarksize}}%
  \pgfpathlineto{\pgfqpointpolar{-234}{0.382\pgfplotmarksize}}%
  \pgfpathclose
  \pgfusepathqfillstroke
}
\pgfplotsset{
  fig1/.style={
    scale only axis, height=5.95cm, ymin=82.5, ymax=85.07,
    grid=major, grid style={gray!15}, ylabel near ticks,
    tick label style={font=\fontfamily{ptm}\fontsize{6}{7}\selectfont},
    ylabel style={font=\scriptsize, yshift=-5pt}, title style={font=\scriptsize, yshift=-1pt},
    clip=false,
  },
  pSteps/.style={only marks, color=cSteps, mark=*,        mark size=1.7pt},
  pAttn/.style={ only marks, color=cAttn,  mark=triangle*, mark size=2.3pt},
  pVae/.style={  only marks, color=cVae,   mark=diamond*,  mark size=2.4pt},
  pOne/.style={  only marks, color=cOne,   mark=pentagon*, mark size=2.5pt},
  pFlex/.style={ only marks, color=cFlex,  mark=star,      mark size=3.0pt, thick},
  pNvfp/.style={ only marks, color=cNv,    mark=otimes*,   mark size=2.9pt, thick},
  pUs/.style={only marks, mark=unstepstar, mark size=3.8pt,
              mark options={draw=orange!80!black, fill=orange!60!yellow, line width=0.55pt}},
}
\tikzset{lbl/.style={font=\fontfamily{ptm}\fontsize{5.25}{5.9}\selectfont, inner sep=1.8pt},
         ldr/.style={draw=cSteps!55, line width=0.3pt}}
\def\vn#1{\,{\fontfamily{ptm}\fontsize{4.4}{5}\selectfont[#1]}}
\def\figunstep{{\bfseries\scshape UnStep}}

% ======================= H100 =======================
\begin{axis}[fig1, name=H, width=11.02cm,
  xmin=13.4, xmax=61, xtick={20,30,40,50,60},
  ylabel={VBench 5s T2V total}, title={\textbf{H100}},
  legend style={at={(0.988,0.01)}, anchor=south east, font=\fontsize{4.6}{5.4}\selectfont,
                legend columns=2, draw=gray!40, fill=white, column sep=5pt, row sep=0.4pt},
  legend cell align=left,
]
\addplot[pSteps, forget plot] coordinates {(25.6,84.24) (23.1,84.13) (23.0,84.09) (22.63,83.73) (18.4,84.34) (17.4,84.31)};
\addplot[pAttn,  forget plot] coordinates {(29.5,83.60) (24.3,83.90) (21.93,83.89) (21.7,83.52) (19.9,83.99) (18.9,84.50)};
\addplot[pVae,   forget plot] coordinates {(31.0,84.48)};
\addplot[pOne,   forget plot] coordinates {(20.41,83.35) (20.72,83.76) (32.3,82.65)};
\addplot[pUs,    forget plot] coordinates {(49.8,84.56) (49.8,84.9256) (59.6,83.3359) (59.6,83.1039)};

\node[lbl, anchor=east, cSteps] at (axis cs:17.21,84.310) {AnyFlow};
\node[lbl, anchor=west, cSteps] at (axis cs:18.49,84.340) {In-Context Forcing};
\node[lbl, anchor=west, cAttn] at (axis cs:19.39,84.500) {Light Forcing\vn{ICML'26}};
\node[lbl, anchor=east, cAttn] at (axis cs:19.41,83.990) {Sparse Forcing};
\node[lbl, anchor=west, cOne, align=center] at (axis cs:20.90,83.300) {Causal Forcing++\\1 step};
\node[lbl, anchor=east, cOne, align=center] at (axis cs:20.53,83.720) {One-Forcing\\1 step};
\node[lbl, anchor=west, cAttn] at (axis cs:22.19,83.520) {MAG};
\node[lbl, anchor=east, cAttn] at (axis cs:21.44,83.890) {Ms. Forcing};
\node[lbl, anchor=west, cSteps] at (axis cs:23.12,83.730) {DSA\vn{CVPR'26}};
\node[lbl, anchor=north, cSteps] at (axis cs:24.50,84.090) {rCM\vn{ICLR'26}};
\node[lbl, anchor=west, cSteps] at (axis cs:23.21,84.130) {Reward Forcing\vn{CVPR'26}};
\node[lbl, anchor=west, cAttn] at (axis cs:24.79,83.900) {Dummy Forcing};
\node[lbl, anchor=west, cSteps] at (axis cs:26.09,84.240) {Causal-rCM};
\node[lbl, anchor=west, cAttn] at (axis cs:29.99,83.600) {Hybrid Forcing};
\node[lbl, anchor=west, cVae] at (axis cs:31.49,84.480) {DiagDistill\vn{ICLR'26}};
\node[lbl, anchor=east, cOne] at (axis cs:31.81,82.650) {rCM, 1 step\vn{ICLR'26}};
\node[lbl, anchor=north west, align=center, xshift=3pt, yshift=5pt] at (axis cs:49.8,84.560)
  {\textbf{Self Forcing + }\figunstep\\[-0.3ex]2 steps};
\node[lbl, anchor=north west, align=center, xshift=3pt, yshift=1pt] at (axis cs:49.8,84.9756)
  {\textbf{Causal Forcing + }\figunstep\\[-0.3ex]2 steps};
\node[lbl, anchor=south east, align=center, xshift=-4pt, yshift=-7pt] at (axis cs:59.6,83.3359)
  {\textbf{Self Forcing + }\figunstep\\[-0.3ex]1 step};
\node[lbl, anchor=south east, align=center, xshift=-4pt, yshift=-7pt] at (axis cs:59.6,83.1039)
  {\textbf{Causal Forcing + }\figunstep\\[-0.3ex]1 step};

\addlegendimage{only marks, color=cSteps, mark=*,         mark size=1.7pt}\addlegendentry{fewer denoising steps}
\addlegendimage{only marks, color=cAttn,  mark=triangle*, mark size=2.3pt}\addlegendentry{sparse attention}
\addlegendimage{only marks, color=cOne,   mark=pentagon*, mark size=2.5pt}\addlegendentry{distilled for one step}
\addlegendimage{only marks, color=cVae,   mark=diamond*,  mark size=2.4pt}\addlegendentry{replacement decoder}
\addlegendimage{pUs}\addlegendentry{\textsc{UnStep} (ours)}
\end{axis}

% ======================= GB200 =======================
\begin{axis}[fig1, name=G, width=4.25cm, at={(H.south east)}, anchor=south west, xshift=9pt,
  xmin=38.3, xmax=80, xtick={45,55,65,75}, yticklabels={}, ylabel={},
  title={\textbf{GB200}},
  legend style={at={(0.025,0.025)}, anchor=south west, font=\fontsize{4.6}{5.4}\selectfont,
                legend columns=1, draw=gray!40, fill=white, row sep=0.4pt},
  legend cell align=left,
]
\addplot[pFlex, forget plot] coordinates {(45.6,84.29) (48.9,83.92) (41.5,83.91)};
\addplot[pNvfp, forget plot] coordinates {(45.7,83.14)};
\addplot[pUs,   forget plot] coordinates {(63.73,84.3575) (63.73,84.5931) (77.44,83.4923) (77.44,83.1636)};

\node[lbl, cFlex, align=center, inner sep=1pt] at (axis cs:45.30,84.105)
  {Flex-Forcing\\{\fontsize{4.2}{4.8}\selectfont[ICML'26]}\\(NVIDIA)};
\node[lbl, anchor=west, cNv, align=center]   at (axis cs:41.04,83.30)
  {LongLive 2.0\\(NVIDIA)};
\node[lbl, anchor=north west, align=center, xshift=4pt, yshift=1pt] at (axis cs:63.73,84.6531)
  {\textbf{CF + }\figunstep\\[-0.3ex]2 steps};
\node[lbl, anchor=north west, align=center, xshift=4pt, yshift=1pt] at (axis cs:63.73,84.4075)
  {\textbf{SF + }\figunstep\\[-0.3ex]2 steps};
\node[lbl, anchor=east, xshift=-4pt] at (axis cs:77.44,83.4923)
  {\textbf{SF + }\figunstep, 1 step};
\node[lbl, anchor=east, xshift=-4pt] at (axis cs:77.44,83.1636)
  {\textbf{CF + }\figunstep, 1 step};

\addlegendimage{only marks, color=cFlex, mark=star,    mark size=3.0pt, thick}\addlegendentry{chunk schedule search}
\addlegendimage{only marks, color=cNv,   mark=otimes*, mark size=2.9pt, thick}\addlegendentry{NVFP4 quantization}
\addlegendimage{pUs}\addlegendentry{\textsc{UnStep} (ours)}
\end{axis}

\node[anchor=north, font=\small] at
  ($(current bounding box.south west)!0.5!(current bounding box.south east)+(0,-4pt)$)
  {end-to-end throughput (FPS)};
\end{tikzpicture}

\caption{Speed in FPS against VBench quality in the H100 comparison (left) and on GB200 (right). In the GB200 plot, SF and CF denote Self Forcing and Causal Forcing. Appendix~\ref{app:speed-sources} lists the details of every point.}
\label{fig:speed}
\end{figure}

%Importance of training-free and introduction of \textsc{UnStep}
The flexibility of choosing the speed quality tradeoff freely at inference motivates a training-free approach, where the same model checkpoint can be directly used without retraining the checkpoint. Toward this goal, we present \textsc{UnStep}, a training-free wrapper that runs on the original Self Forcing checkpoint and boosts its speed from 17 FPS to 50 FPS while slightly raising its quality (Table \ref{tab:othermodels}). The 1-step variant of \textsc{UnStep} runs at 60 FPS with comparable quality to other models trained explicitly with 1 step, while being significantly faster (Figure \ref{fig:speed}). The choice between the 2-step (default) or 1-step variant of \textsc{UnStep} is determined only at inference, without any retraining.

%%%%%%%%%%%%%%%%%%%%%%%%%%%%%%%%%%%%%%%%%%%%%%%%%%%%%%%%%%%%%%%%%%%%%%%%%%%%%%%%%%%%%%%%%%%%%%%%%%%%%%%%%%%%%%%%%%%%%%%%%%%%%%%%%%%%%%%%%%%%%%%%%%%%%%%%%%%%%%%%%%%%%%%%%%%%%%
%%%%%%%%%%%%%%%%%%%%%%%%%%%%%%%%%%%%%%%%%%%%%%%%%%%%%%%%%%%%%%%%%%%%%%%%%%%%%%%%%%%%%%%%%%%%%%%%%%%%%%%%%%%%%%%%%%%%%%%%%%%%%%%%%%%%%%%%%%%%%%%%%%%%%%%%%%%%%%%%%%%%%%%%%%%%%%

%Challenges of training-free
%The training-free constraint poses a challenge: the mismatch between train and inference configurations without training can cause quality drop. For instance, our experiments (§\ref{sec:method}) show that inference-only step reduction degrades quality, whereas Causal Forcing++ \cite{zhao2026causalforcingpp} shows that distilling with 2 steps can match the quality of 4-step distillation. Similarly, Light Forcing \citep{lv2026lightforcing} shows that applying sparse attention only at inference collapses quality, while finetuning can recover it. Therefore, we design \textsc{UnStep} under conservative constraints: we avoid aggressive inference changes that would require finetuning, and any quality lost to acceleration must be recovered using inference-only mechanisms.

%%%more challenges: few-step (causal) vs Multi-step (bidirectional) acceleration
%%%Making a few-step student faster is a harder problem than for a multi-step teacher. At fifty steps the transformer dominates the runtime, and halving the step count nearly halves the runtime, and the decoder does not matter. A four-step student has a fast DiT, and each of the four steps contibutes to quality, and now the surrounding execution and the decoder are first-order costs, so adressing only DiT wont affect the speed as much, as there are other bottlenecks. 

% Unstep explained: runtime stack

\textsc{UnStep} achieves these gains by optimizing the runtime stack for both DiT and VAE, and reducing DiT computation. The runtime optimizations target KV indexing, attention calls, rotary position embeddings (RoPE), and VAE decoding (§\ref{sec:method:runtime}). We keep KV position indices on CPU to avoid synchronization overhead with GPU, and read token counts for FlashAttention calls from query and key tensor shapes without allocating separate GPU tensors. We also compute RoPE rotations using a fused Triton kernel together with cached sine and cosine coefficients, which reduces repeated computation and kernel launches. For VAE decoding, we use FP16 precision and a channels-last 3D memory layout, with convolution autotuning that selects fast kernels for the decoder's tensor shapes. This runtime stack preserves the denoising schedule, attention context, and overall quality while boosting throughput by about 10 FPS (Table \ref{tab:methods}).

% Unstep explained: less steps + attention window
We then reduce DiT computation by using fewer denoising steps (§\ref{sec:method:steps}) and limiting the KV attention window (§\ref{sec:method:attention}). Since these changes degrade quality, we propose two inference-only recovery mechanisms. We refine the clean-cache pass originally used only to update the KV cache for later generation by first renoising the current latent frames to a near clean level, then reusing the same pass to refine them (§\ref{sec:method:clean}). Next, we replace the attention value and output projections by their truncated SVD (§\ref{sec:method:svd}). \iftheory{\S\ref{sec:theory} provides theoretical intuition for why both mechanisms can help under reduced step inference.} Together, these components recover the quality lost while preserving the training-free setting and boosting FPS by a further 18 FPS (Table \ref{tab:methods}). Figure~\ref{fig:method-overview} summarizes the complete \textsc{UnStep} pipeline. %while Figures~\ref{fig:qualitative-otter} and \ref{fig:qualitative-dog}--\ref{fig:qualitative-elephant} show representative outputs before and after applying the wrapper.

\begin{figure}[!t]
\centering
\pairedqualcomparison
  {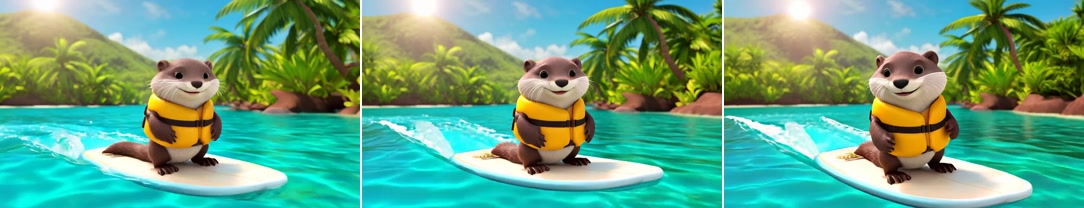}
  {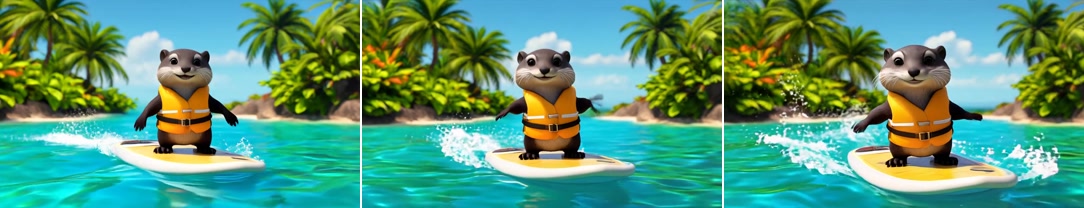}
  {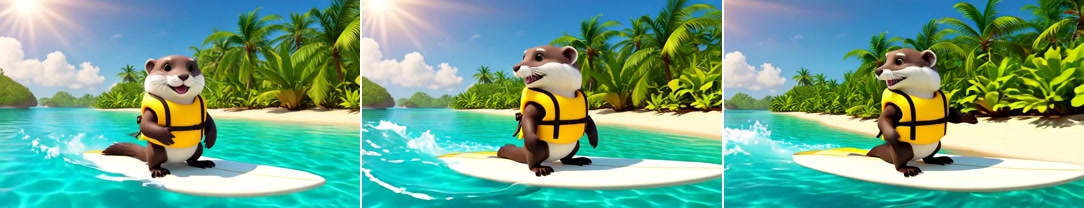}
  {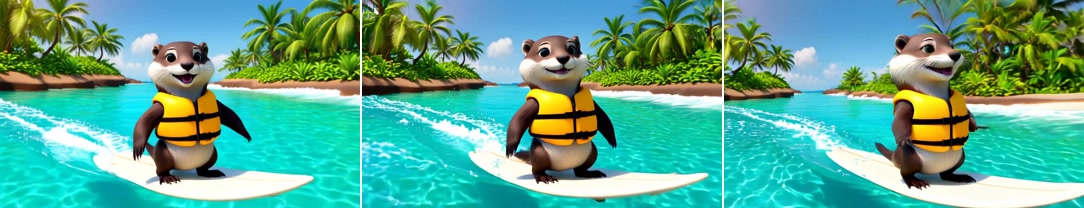}
  {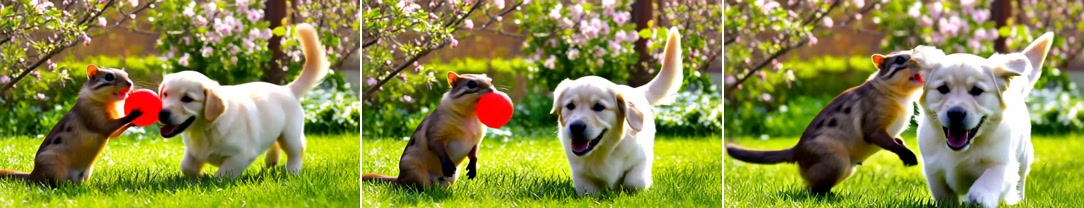}
  {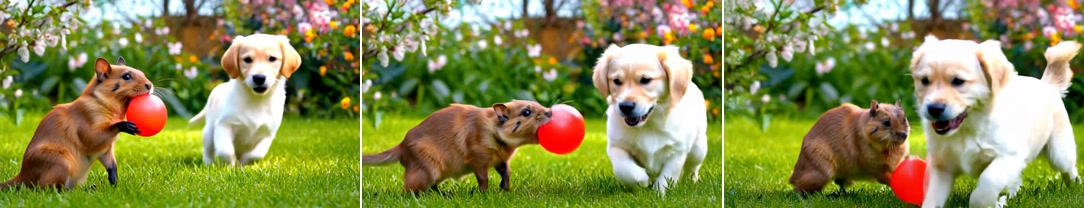}
  {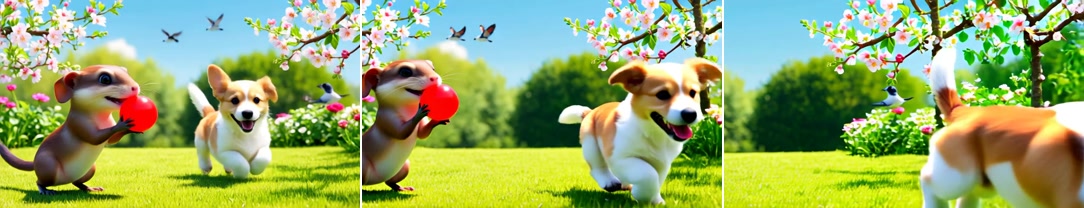}
  {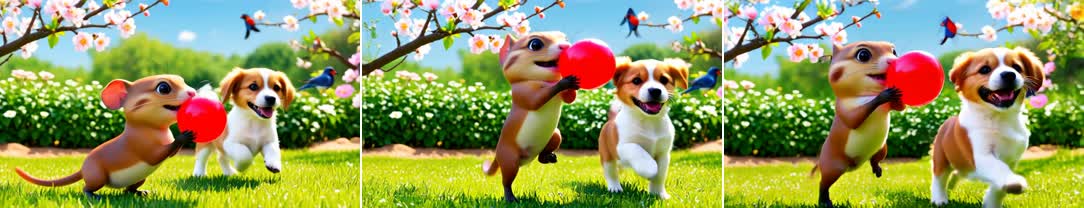}
\caption{Sample videos generated by Self Forcing and Causal Forcing, with and without \textsc{UnStep}, for an otter surfing (top) and a ferret playing ball with a puppy (bottom).}
\label{fig:qualitative-otter}
\label{fig:qualitative-dog}
\end{figure}

%Vbench and speed comparison on H100 and GB200
We take the standard VBench 5s (5-second clips) T2V (text-to-video) suite \citep{huang2024vbench} as the main quality benchmark. To our knowledge, \textsc{UnStep} is the first method to enable Wan2.1-1.3B to stream close to 50 FPS on an H100 while maintaining competitive VBench performance (Figure \ref{fig:speed}---left), and it does so without retraining. The fastest existing methods \cite{diagdistill, hybridforcing} stream at around 30 FPS despite relying on training with smaller decoders, shortened denoising schedules, or sparse attention. More strikingly, on a GB200, \textsc{UnStep} is over 30 FPS faster than NVIDIA's recent LongLive 2.0 \citep{longlive2} with slightly better VBench performance (Figure \ref{fig:speed}---right). This is notable as LongLive 2.0 is specifically optimized for GB200 with NVFP4 quantization. \textsc{UnStep} achieves this unusually large speed jump by targeting bottlenecks in the surrounding runtime stack in addition to the DiT computation reductions, while remaining entirely inference-only.
%, a 4-bit floating-point format designed for Blackwell GPUs, to accelerate inference

% Tailored for one setup, zero-shot on the rest
We design and tune \textsc{UnStep}'s parameters in a single setup: Self Forcing as the causal model, H100 as the GPU accelerator, and VBench 5s T2V as the quality benchmark. We then fix these parameters and design choices and apply \textsc{UnStep} as is, in a zero-shot setting, across several different setups: different checkpoints, including Causal Forcing \citep{zhu2026causalforcing} and LongLive 1.0 \citep{longlive} (§\ref{sec:exp:othermodels}), different GPU, GB200 (§\ref{sec:exp:gb200}), a 1-denoising-step variant (§\ref{sec:exp:onestep}), image-to-video generation on VBench I2V \citep{huang2024vbenchpp} (§\ref{sec:exp:i2v}), and 30-second long clips on VBench Long \citep{huang2024vbenchpp}  (§\ref{sec:exp:long}). Across all setups, the central observation remains the same: \textsc{UnStep} preserves the quality of the original checkpoint, while substantially improving its speed.

%Summary of contributions
\paragraph{To summarize, our key contributions are:}
\begin{itemize}
\itemsep1pt
    \item We introduce \textsc{UnStep}, a training-free inference wrapper that accelerates few-step causal video diffusion models built on Wan 2.1 (§\ref{sec:method}), and enables streaming at 50 FPS on H100 with strong VBench performance, and up to 77 FPS on GB200 (Figure \ref{fig:speed}).
    \item We reduce runtime stack overheads with more efficient attention calls and KV indexing, fused Triton RoPE with cached coefficients, and FP16 VAE decoding with a convolution friendly memory layout and convolution autotuning (§\ref{sec:method:runtime}).
    \item We reduce computation by using fewer DiT denoising steps (§\ref{sec:method:steps}) and a limited temporal KV attention window (§\ref{sec:method:attention}), and recover the resulting quality loss with a refined clean-cache pass (§\ref{sec:method:clean}) and truncated SVD of the attention value/output projections (§\ref{sec:method:svd}).
    \item We show that \textsc{UnStep} allows the speed quality tradeoff to be chosen flexibly at inference from the same checkpoint (§\ref{sec:exp:onestep}) and generalizes across checkpoints (§\ref{sec:exp:othermodels}), GPUs (§\ref{sec:exp:gb200}), image-to-video generation (§\ref{sec:exp:i2v}), and long video generation (§\ref{sec:exp:long}).
\ifnum\showtheory=1
    \item Finally, we develop a toy mathematical model that provides intuition for why clean-cache refinement and spectral truncation can help recover quality under reduced step inference (§\ref{sec:theory}).
\fi
\end{itemize}

%% file: sections/related_work.tex
\section{Related Work}
\label{sec:related}

\paragraph{Few-step causal video models.}
CausVid distilled a bidirectional video diffusion model into an autoregressive generator \citep{yin2025causvid}. Self Forcing reduced the train--test mismatch by distilling on the student's own rollouts \citep{huang2025selfforcing}, whereas Causal Forcing initialized the causal student using an autoregressive ODE teacher before applying asymmetric distribution matching \citep{zhu2026causalforcing}. These methods train few-step causal models through distillation. \textsc{UnStep} instead changes the inference procedure of an already distilled causal checkpoint.

\paragraph{Training-free acceleration of causal video.}
Several causal video approaches change the autoregressive computation without retraining. Block Cascading changes the block schedule \citep{bandyopadhyay2025blockcascading}, while Dummy Head and HeadCast restrict history access for certain attention heads \citep{guo2026dummyhead,headcast}. FlowCache reuses computations across autoregressive flow evaluations, while history-guided caching combines feature reuse with residual correction \citep{flowcache,historycache}. Surprise Forcing jointly decides what history to retain and when to skip computation \citep{surpriseforcing}. \textsc{UnStep} instead uses a fixed shorter schedule and bounded KV window, then recovers their quality loss through the clean-cache pass and V/O spectral edit.

\paragraph{Attention, decoder, and runtime acceleration.}
Trained causal models reduce attention or memory cost through sparse, hybrid, or multiscale context \citep{lv2026lightforcing,xu2026sparseforcing,hybridforcing,msforcing}, while Forcing-KV uses training-free head-dependent KV compression \citep{ji2026forcingkv}. MonarchRT also targets efficient attention for Self Forcing \citep{monarchrt}. DiagDistill and Flash-VAED use small replacement decoders \citep{diagdistill,flashvaed}. \textsc{UnStep} keeps the original checkpoint and VAE weights, but combines inference-only runtime changes across the DiT and decoder in addition to changes to the schedule and attention window.

\paragraph{KV window and spectral edit.}
The limited KV attention window follows the attention sink idea in Streaming\-LLM \citep{xiao2024streamingllm} and its adaptation to causal video generation in LongLive \citep{longlive}. \textsc{UnStep} combines reduced attention window with reduced step inference and clean-cache recovery. Its V/O truncation is motivated by LASER, which showed selective post-training rank reduction can improve some tasks \citep{sharma2024laser}. \textsc{UnStep} replaces selected V/O weight matrices by dense truncated SVD reconstructions to recover quality instead of reducing projection cost. Appendix~\ref{app:related} discusses broader related work.

%% file: sections/method.tex
\section{UnStep}
\label{sec:method}

As shown in Figure~\ref{fig:method-overview}, \textsc{UnStep} accelerates few-step causal video diffusion entirely at inference. We first describe the optimized runtime stack for the DiT and VAE decoder, then show how \textsc{UnStep} reduces DiT computation through fewer denoising steps and a limited temporal KV attention window. Finally, we describe the two inference-only mechanisms used to recover the quality lost by these speedups: the refined clean-cache pass and V/O truncated SVD. \iftheory{\S\ref{sec:theory} gives a toy model intuition for why these two recovery mechanisms can help under reduced step inference.} Appendix~\ref{app:implementation} gives the exact implementation configuration and reports additional ablations.

Table~\ref{tab:methods} summarizes the cumulative ablation of \textsc{UnStep}. C0 starts from original Self Forcing with FlashAttention-3 \citep{shah2024flashattention3} and applies standard \texttt{torch.compile} \cite{ansel2024pytorch2} to the DiT and VAE decoder to optimize execution for the fixed tensor shapes. Each subsequent row adds one component of \textsc{UnStep} and reports its effect on end-to-end throughput on H100 and VBench 5s T2V quality. The rest of this section follows the rows of Table~\ref{tab:methods}, describing the runtime stack, the generation trajectory changes, and the two quality recovery mechanisms.

\begin{figure}[!t]
\centering
\includegraphics[width=\linewidth]{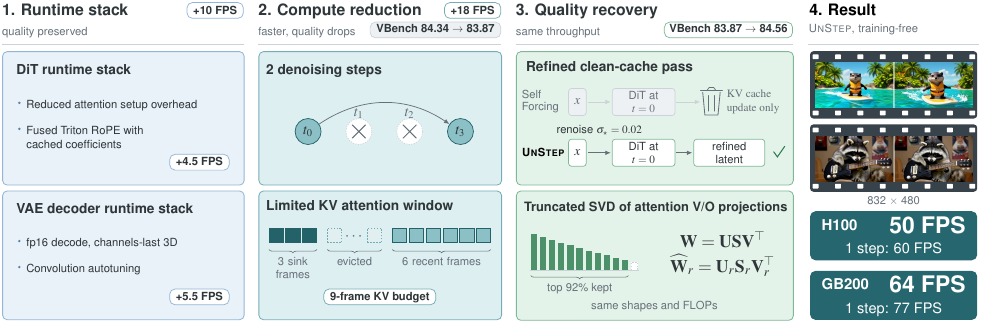}
\caption{Overview of \textsc{UnStep}. First column shows the kernel and runtime optimizations that accelerate the DiT and VAE without degrading quality. Second column shows the reduced DiT computation that further increases speed but reduces quality. Third column shows the quality recovery mechanisms, which add no runtime. All changes are inference-only, with no retraining.}
\label{fig:method-overview}
\end{figure}

\begin{table}[htbp]
\centering
\small
\setlength{\tabcolsep}{5pt}
\begin{tabular*}{\linewidth}{@{\extracolsep{\fill}}lllrrrrc@{}}
\toprule
& & Configuration & FPS & Total & Quality & Semantic & \S \\
\midrule
\multirow{2}{*}{Original} & \multirow{2}{*}{C0}
      & Self Forcing + FA3                      & 17.0 & 84.31 & 85.28 & 80.44 & \multirow{2}{*}{\ref{sec:method}} \\
 &    & $+$ compilation                         & 22.2 & 84.28 & 85.26 & 80.35 & \\
\midrule
\multirow{2}{*}{\shortstack[l]{Runtime\\stack}}
 & C1 & C0 $+$ transformer runtime stack       & 26.7 & 84.34 & 85.24 & 80.72 & \ref{sec:method:runtime} \\
 & C2 & C1 $+$ decoder runtime stack           & 32.2 & 84.34 & 85.26 & 80.68 & \ref{sec:method:runtime} \\
\midrule
\multirow{2}{*}{Speedup}
 & C3 & C2 $+$ two denoising steps               & 44.3 & 83.91 & 84.68 & 80.82 & \ref{sec:method:steps} \\
 & C4 & C3 $+$ limited attention window          & \textbf{49.8} & 83.87 & 84.67 & 80.71 & \ref{sec:method:attention} \\
\midrule
\multirow{2}{*}{\shortstack[l]{Quality\\recovery}}
 & C5 & C4 $+$ clean-cache pass                  & \textbf{49.8} & 84.33 & 84.91 & 81.99 & \ref{sec:method:clean} \\
 & C6 & C5 $+$ V/O truncated SVD                 & \textbf{49.8} & \textbf{84.56} & 85.20 & 82.02 & \ref{sec:method:svd} \\
\bottomrule
\end{tabular*}
\caption{The components of \textsc{UnStep} applied cumulatively. Speed is measured on H100 end-to-end for both DiT and VAE, and  quality on the VBench 5s T2V suite. Base model is Wan2.1-1.3B  distilled to 4 steps with Self Forcing with resolution $832\times480$. FA3 stands for FlashAttention 3.}
\label{tab:methods}
\end{table}

\subsection{The runtime stack}
\label{sec:method:runtime}

The computational cost of operations such as handling KV cache positions, preparing and applying rotary position embeddings (RoPE), setting up attention sequence lengths, and decoding latents with the VAE are usually hidden in multi-step diffusion models, as repeated DiT forward passes dominate runtime. In few-step causal models, these computational costs become visible in end-to-end throughput. In this subsection we target these operations, and show that making them more efficient can significantly improve throughput.

\paragraph{DiT runtime stack:}

Before each self-attention call, Self Forcing uses cache position indices to select where to write the current keys and values and which cached range to pass to attention. These are stored on GPU, so Python reads require CPU--GPU synchronization. These indices depend on token positions and counts, not on the computed keys or values. \textsc{UnStep} therefore maintains them on CPU, which eliminates the CPU--GPU communication overhead. Self Forcing also uses variable-length FlashAttention, which supports different sequence lengths within a batch but requires extra GPU tensors recording their lengths and boundaries, which are rebuilt on every attention call. For single video inference, \textsc{UnStep} uses fixed-length FlashAttention, which reads token counts from the query and key tensor dimensions. This avoids repeated GPU tensor allocations and the operations used to construct them. Together with the runtime changes detailed in Appendix~\ref{app:implementation:dit-runtime}, these changes add 2.3 FPS, reaching 24.5 FPS.

\textsc{UnStep} next replaces the RoPE application path with a fused Triton kernel (Appendix~\ref{app:implementation:dit-runtime}). RoPE encodes token positions by rotating query and key coordinate pairs. Original Self Forcing computes these rotations through separate PyTorch operations that create temporary tensors and launch multiple GPU kernels. \textsc{UnStep} computes the same rotations in one kernel using \texttt{float32} instead of \texttt{float64}, avoiding intermediate writes and repeated launches. Original Self Forcing caches the base RoPE table, but builds the coefficients for each chunk's token positions on every call. \textsc{UnStep} caches these per chunk sine and cosine tensors by chunk shape and starting latent frame index, across layers and denoising steps. The fused RoPE kernel with cached coefficients adds another 2.2 FPS, reaching 26.7 FPS, as shown in C1 in Table~\ref{tab:methods}.

\paragraph{VAE decoder:}

We next optimize the VAE decoder while keeping the same VAE weights and causal temporal behavior. Original Self Forcing runs the VAE in \texttt{bfloat16} and leaves its convolution weights in PyTorch's default memory layout. \textsc{UnStep} casts both the decoder weights and latent inputs to \texttt{fp16}, and uses a channels-last 3D memory layout for convolution weights and decoder inputs. This places the channel values at each spatial and temporal position next to one another in memory. The precision and layout are chosen together to use faster convolution kernels on H100. Together with output preallocation and the decoding changes detailed in Appendix~\ref{app:implementation:vae-runtime}, these changes add 2.8 FPS, reaching 29.5 FPS.

Next, \textsc{UnStep} utilizes convolution autotuning to select an efficient GPU kernel. It divides the computation across GPU threads and access memory differently, so their relative speed depends on input shape, precision, and memory layout. During warmup, it measures the execution time of these alternatives and keeps the fastest tested kernel for subsequent calls with the same configuration. This adds another 2.7 FPS, reaching 32.2 FPS, which corresponds to C2 in Table~\ref{tab:methods}.

\subsection{Running below the distilled step count}
\label{sec:method:steps}
  
We next run the model with a shorter inference schedule than the one it was distilled with. Self Forcing is distilled with four DiT steps per chunk. Following common practice in reduced step causal video models \citep{zhao2026causalforcingpp,oneforcing2026}, we keep the first chunk at the original four steps and reduce later chunks to two denoising steps. The first chunk remains unchanged to provide a high-quality initial anchor for the later chunks to reuse its causal context. As shown in C3 of Table~\ref{tab:methods}, this gives the largest single speedup, 12 FPS, but also causes the main quality drop, reducing VBench total from 84.34 to 83.91. We also evaluate a faster one-step variant, which uses one denoising step for later chunks instead of two (§\ref{sec:exp:onestep}).

\subsection{Limited KV attention window}
\label{sec:method:attention}

Causal models cache past keys and values so that each new chunk can attend to earlier generated chunks without recomputation. In original Self Forcing, this cache grows with the video length, so the attention cost increases as generation continues. \textsc{UnStep} limits this attention cost with a sink and a sliding window cache. Following the same idea as StreamingLLM \citep{xiao2024streamingllm} and its use in video diffusion by LongLive 1.0 \citep{longlive}, we retain the first 3 latent frames as sink tokens and the 6 most recent frames as a sliding window, for a total KV budget of 9 latent frames, evicting older middle tokens from the cache.

This reduces the attention context, increasing throughput by 5.5 FPS as seen in C4 of Table~\ref{tab:methods}. Since the cache size no longer grows with sequence length, the same mechanism also supports long streaming generation (§\ref{sec:exp:long}). This is the last speed oriented change, the remaining components recover the quality lost from using fewer steps and limited attention.

\begin{figure}[!t]
\centering
\pairedqualcomparison
  {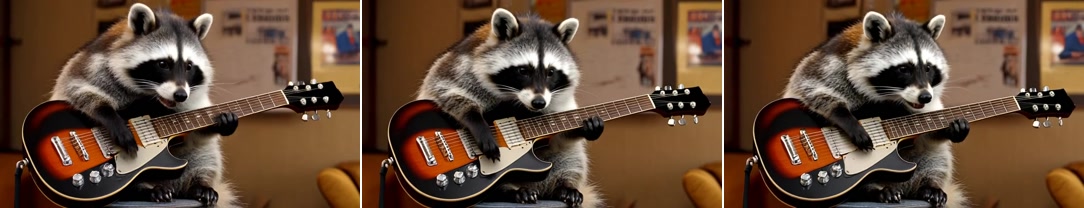}
  {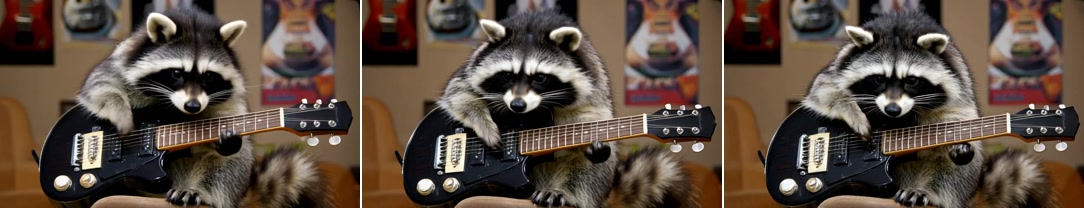}
  {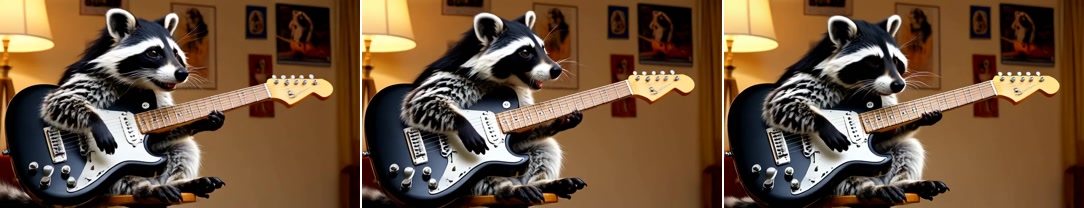}
  {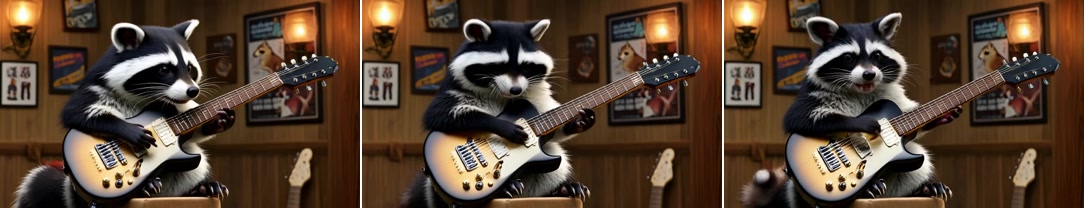}
  {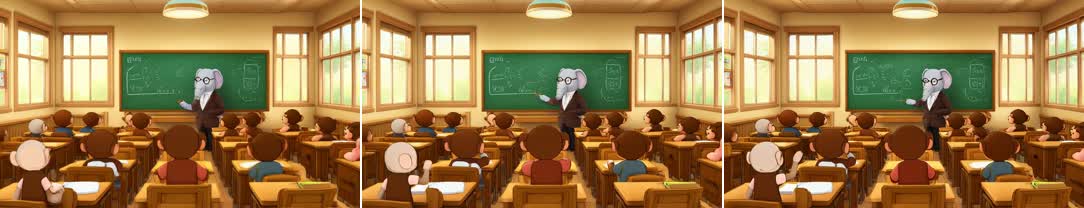}
  {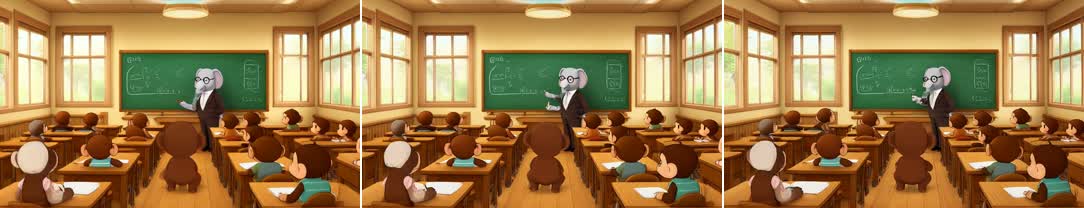}
  {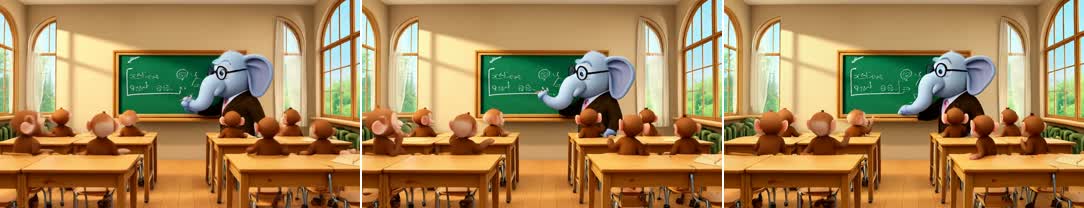}
  {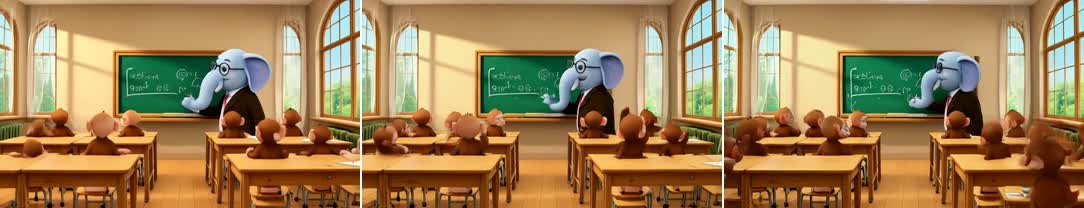}
\caption{Sample videos generated by Self Forcing and Causal Forcing, with and without \textsc{UnStep}, for a raccoon playing guitar (top) and an elephant teaching monkeys (bottom).}
\label{fig:qualitative-raccoon}
\label{fig:qualitative-elephant}
\end{figure}

\subsection{Refining the clean-cache pass}
\label{sec:method:clean}

We next introduce the first inference-only quality recovery mechanism. In flow matching diffusion, the DiT predicts a velocity field. Given the current estimate $\mathbf{x}_i$, a sampler step first places it at the desired noise level, evaluates the DiT at the corresponding timestep label, and converts the predicted velocity into the next estimate as follows:
\begingroup
\setlength{\abovedisplayskip}{3pt}
\setlength{\abovedisplayshortskip}{3pt}
\begin{align}
  \tilde{\mathbf{x}}_i &= (1-\sigma_i)\mathbf{x}_i + \sigma_i \bm{\epsilon}_i,
  \qquad \bm{\epsilon}_i\sim\mathcal{N}(\mathbf{0},\mathbf{I}), \label{eq:clean_renoise}\\
  \mathbf{v}_i &= f_{\bm{\theta}}(\tilde{\mathbf{x}}_i,t_i), \label{eq:clean_forward}\\
  \mathbf{x}_{i+1} &= \tilde{\mathbf{x}}_i - \sigma_i \mathbf{v}_i. \label{eq:clean_emit}
\end{align}
\endgroup
Here $f_{\bm{\theta}}$ is the DiT, $t_i$ is the scalar conditioning label given to the DiT, and $\sigma_i$ is the scalar noise level corresponding to that label by the scheduler. The same $\sigma_i$ controls both the amount of noise injected before the DiT call and the scale of the velocity subtraction after it.

Self Forcing already runs one extra DiT evaluation after each chunk at the clean label $t=0$ (equation \eqref{eq:clean_forward}) to update keys and values for later chunks. In the original sampler, this clean-cache pass uses $\sigma=0$, so equations~\eqref{eq:clean_renoise} and \eqref{eq:clean_emit} are identities: the input is unchanged, the DiT output is discarded, and the pass only updates the KV cache. \textsc{UnStep} keeps the DiT conditioning label at $t=0$, but uses a small positive noise level $\sigma_\star=0.02$ in equations~\eqref{eq:clean_renoise} and \eqref{eq:clean_emit}. This keeps the cache update close to the clean label behavior used by Self Forcing, while allowing the same already run forward pass to refine the current chunk and produce the emitted latent. In C5 of Table~\ref{tab:methods}, this refinement keeps throughput unchanged and recovers VBench total from 83.87 to 84.33. \iftheory{\S\ref{sec:theory:clean} gives a toy model explanation for why a nonzero clean-cache noise level can help after step reduction.}

\subsection{Truncated SVD of the value and output projections}
\label{sec:method:svd}

We finally add a second inference-only quality improving mechanism. Although the clean-cache pass already recovers the original Self Forcing quality, a small spectral truncation of the DiT attention projections further improves quality.  We replace the value and output projections by a truncated SVD of the same matrix. For a projection matrix $\mathbf{W}\in\mathbb{R}^{m\times n}$ with $m\le n$, let
  
  \begin{equation}
      \mathbf{W}=\mathbf{U}\mathbf{S}\mathbf{V}^{\top},
      \qquad
      \mathbf{S}=\operatorname{diag}(\lambda_1,\ldots,\lambda_m),
      \quad \lambda_1\ge\cdots\ge\lambda_m\ge 0 .
  \end{equation}
  We keep the leading $r$ singular directions and use
  \begin{equation}
      \widehat{\mathbf{W}}_r=\mathbf{U}_{r}\mathbf{S}_{r}\mathbf{V}_{r}^{\top}.
      \label{eq:svd}
  \end{equation}

Here $\mathbf{U}_{r}$, $\mathbf{S}_{r}$, and $\mathbf{V}_{r}$ contain only the singular vectors and singular values corresponding to the $r$ largest singular values. We set $r = 0.92m$, i.e., we retain 92\% of the singular directions. We form $\widehat{\mathbf{W}}_r$ once when the wrapper is initialized and store it as a dense matrix with the same shape as $\mathbf{W}$. Therefore, the resulting projection has the same parameter count and multiplication cost as before, only the spectrum of the matrix changes. This differs from post-training compression methods that apply SVD truncation to reduce parameters or FLOPs \citep{lin2025modegpt,chiang2025uniql}. Our motivation is instead closer to LASER: low rank truncation can act as a spectral filter that improves a trained transformer by removing weak directions \citep{sharma2024laser}.

We apply this spectral edit to the value and output projections, leaving the query and key projections unchanged. For fixed attention inputs and cached keys, this changes the layer's output without changing its attention probabilities. In C6 of Table~\ref{tab:methods}, V/O truncated SVD improves VBench total from 84.33 to 84.56 with no throughput loss. \iftheory{\S\ref{sec:theory:svd} gives a toy model explanation for why truncating these projections can remove error from the shortened trajectory.}

%% file: sections/experiments.tex
\section{Experiments}
\label{sec:experiments}

Our main evaluation uses Self Forcing \citep{huang2025selfforcing} on Wan2.1-1.3B \citep{wan2025}: text-to-video generation at resolution $832\times480$ for five-second clips, batch size one, on a single H100. Quality is measured with the VBench 5s T2V suite \citep{huang2024vbench} using all 946 prompts, which is the standard benchmark for evaluating causal video models. Appendix~\ref{app:evaluation} provides further details on VBench. Throughout the paper, the reported \textsc{UnStep} scores, are averages over five runs, with statistics for the default wrapper in Appendix~\ref{app:run-variation}.
%This setting is widely used by causal video distillation work, allowing the direct comparison against models trained with fewer steps, sparse attention, or faster decoders under the same setup (Figure \ref{fig:speed}).

We measure throughput as generated video frames per second (FPS) at batch size one. This includes DiT denoising, KV cache operations, VAE decoding, and GPU postprocessing needed to produce the video tensor. It excludes model loading, compilation, GPU warmup, VBench scoring, and saving videos to disk.
Appendix~\ref{app:evaluation} gives the benchmark, scoring, timing, and reproducibility details.

All \textsc{UnStep} parameters are chosen based on the ablations in Table \ref{tab:methods} with Self Forcing as the base checkpoint, a single H100 as the target GPU, and VBench 5s T2V as the quality benchmark. These parameters include the denoising schedule, attention window and sink sizes, the near clean noise level used in the clean-cache pass, SVD retention, and runtime stack choices. The remaining experiments apply the same wrapper without retuning to other checkpoints, GB200, one-step generation, image-to-video generation, and long video generation.

The experiments show a consistent pattern: \textsc{UnStep} significantly accelerates causal video models built on Wan 2.1 while keeping quality close to, and often above, the unwrapped checkpoint. Figures~\ref{fig:qualitative-otter} and \ref{fig:qualitative-raccoon} show sample outputs with and without the wrapper. Appendix~\ref{app:visualizations} provides further comparisons.

\subsection{Other base models}
\label{sec:exp:othermodels}

We now test \textsc{UnStep} on checkpoints beyond Self Forcing. We consider two prominent causal models built on the same Wan2.1 backbone. Causal Forcing \citep{zhu2026causalforcing} changes the distillation recipe. Instead of initializing the causal student directly from a bidirectional teacher, it first uses an autoregressive teacher for ODE initialization and then applies distribution matching distillation. NVIDIA's LongLive 1.0 \citep{longlive} targets interactive streaming and long video generation. These checkpoints therefore test \textsc{UnStep} on strong causal models that differ from Self Forcing in how they are trained or the setting they target.

Table \ref{tab:othermodels} shows the results. All three wrapped checkpoints reach similar throughput, around 50 FPS, since they all share the same wrapper configuration and backbone Wan 2.1. VBench quality is preserved in all 3 cases, while the total score is slightly higher than the unwrapped checkpoint. This shows that \textsc{UnStep} transfers to other causal checkpoints built on Wan 2.1.

\begin{table}[htbp]
\centering
\small

\begin{tabular}{@{}lrrrr@{}}
\toprule
System & FPS & Total & Quality & Semantic \\
\midrule
Self Forcing \citep{huang2025selfforcing} & 17.0 & 84.31 & 85.28 & 80.44 \\
\quad $+$ \textsc{UnStep}                 & \textbf{49.8} & 84.56 & 85.20 & 82.02 \\
\midrule
Causal Forcing \citep{zhu2026causalforcing} & 17.0 & 84.82 & 85.64 & 81.55 \\
\quad $+$ \textsc{UnStep}                   & \textbf{49.8} & 84.93 & 85.79 & 81.46 \\
\midrule
LongLive 1.0 \citep{longlive}                 & 17.0 & 83.31 & 83.64 & 81.98 \\
\quad $+$ \textsc{UnStep}                 & \textbf{49.8} & 83.42 & 83.82 & 81.81 \\
\bottomrule
\end{tabular}
\caption{\textsc{UnStep} applied unchanged to three causal checkpoints on H100. Performance is measured on VBench 5s T2V with resolution $832\times480$. Appendix~\ref{app:baseline-reproduction} and Appendix Table~\ref{tab:t2v-dimensions} provide more details.}
\label{tab:othermodels}
\end{table}

\subsection{Methods that target speed}
\label{sec:exp:speed}

The left panel of Figure~\ref{fig:speed} compares \textsc{UnStep} to recent causal video methods that explicitly target faster inference and report throughput and VBench scores. We group the methods based on the mechanism they use for the speedup: reducing the denoising schedule or changing the causal sampling procedure \citep{zhao2026causalforcingpp,gu2026anyflow,incontextforcing,dsa,zheng2025rcm,rewardforcing,zheng2026causalrcm}, reducing attention or KV cache cost \citep{lv2026lightforcing,xu2026sparseforcing,mag,msforcing,hybridforcing,guo2026dummyhead}, and replacing or distilling the VAE decoder \citep{diagdistill}. Training explicitly for one-step generation is discussed in \S\ref{sec:exp:onestep}.  We report the results form the original papers. Appendix Table~\ref{tab:speed-h100-main} provides more details on the points in Figure \ref{fig:speed}.

The main observation from Figure~\ref{fig:speed} is that recent methods that target speed cluster around the 20--30 FPS range on H100, whereas \textsc{UnStep} reaches close to 50 FPS at comparable or higher VBench quality. This is an unusually large jump in a regime where many methods already use few denoising steps or restricted attention. The key difference is that \textsc{UnStep} also targets the runtime stack overheads that become visible once the DiT already runs for only a few steps: KV indexing and attention calls, RoPE computation, GPU kernel overheads, and VAE decoding as explained in \S\ref{sec:method:runtime}. By optimizing these operations together with step reduction and limited KV attention, \textsc{UnStep} becomes a substantially faster end-to-end video accelerator without any training.

%Another difference is where the speed-quality trade-off is decided. Existing methods improve throughput by changing what the model is trained to do: they distill a shorter denoising schedule, train a smaller decoder, or finetune with sparse attention or KV-cache policies. They can therefore recover the quality loss during training. \textsc{UnStep} instead wraps an existing checkpoint and chooses the operating point entirely at inference. Any quality loss from acceleration must be recovered without gradients, through the refined clean-cache pass and V/O truncated SVD. This keeps the operating point flexible: the same checkpoint can be run with the default 2-step wrapper or a faster 1-step variant without retraining.

\subsection{Blackwell GPU}
\label{sec:exp:gb200}

So far, we have used H100 GPUs, following most recent causal video generation work. Two recent NVIDIA papers, LongLive 2.0 \citep{longlive2} and Flex-Forcing \citep{flexforcing}, are instead built around NVIDIA's newer Blackwell generation,  GB200. GB200 provides faster tensor core and memory system execution for the matrix multiplications, attention kernels, and convolutions used in video generation. LongLive 2.0 accelerates inference by quantizing weights, activations, and the KV cache to NVFP4, a 4-bit floating-point format designed for Blackwell GPUs. Flex-Forcing searches over handcrafted chunk schedules and sizes and reports the best GB200 configuration for the 5-second VBench benchmark.

We evaluate \textsc{UnStep} unchanged on GB200, using both the 1-step and 2-step variants on the Self Forcing and Causal Forcing checkpoints. Figure~\ref{fig:speed} compares these results to the fast versions of LongLive 2.0 and Flex-Forcing. Without retuning for GB200, the 1-step \textsc{UnStep} variant is more than 30 FPS faster than LongLive 2.0 while also achieving higher VBench quality. The 2-step variant is substantially faster than the best Flex-Forcing configuration and also scores higher. Thus, even though \textsc{UnStep} is designed and tuned on H100, it remains faster than these GB200 optimized models with slightly better VBench quality. Appendix Table~\ref{tab:speed-gb200} lists the exact GB200 values.

\subsection{One denoising step}
\label{sec:exp:onestep}

Several recent video models target the low-latency setting of one denoising step per generated chunk \citep{oneforcing2026,zhao2026causalforcingpp,zheng2025rcm}. These models correspond to the three red points in Figure~\ref{fig:speed}. They are trained specifically for the 1-step regime, so the latency quality tradeoff is determined during distillation. \textsc{UnStep} is tuned for the default 2-step setting, but the number of denoising steps remains an inference time choice. Appendix Table~\ref{tab:speed-h100-one-step} gives the score and throughput sources for all one-step points.

Figure~\ref{fig:speed} shows that the default 2-step \textsc{UnStep} variant is already faster and higher quality than the reported 1-step baselines on H100. We also run a 1-step \textsc{UnStep} variant to show that the same checkpoint can become even faster when desired. This variant reaches 59.6 FPS, trading quality relative to the 2-step wrapper for additional throughput.

A key point is flexibility. With \textsc{UnStep}, the same checkpoint can be run with 2 steps for higher quality, or with 1 step for higher throughput, whereas a model distilled directly for one step fixes this choice in the weights, and must be redistilled for deploying in another setup.

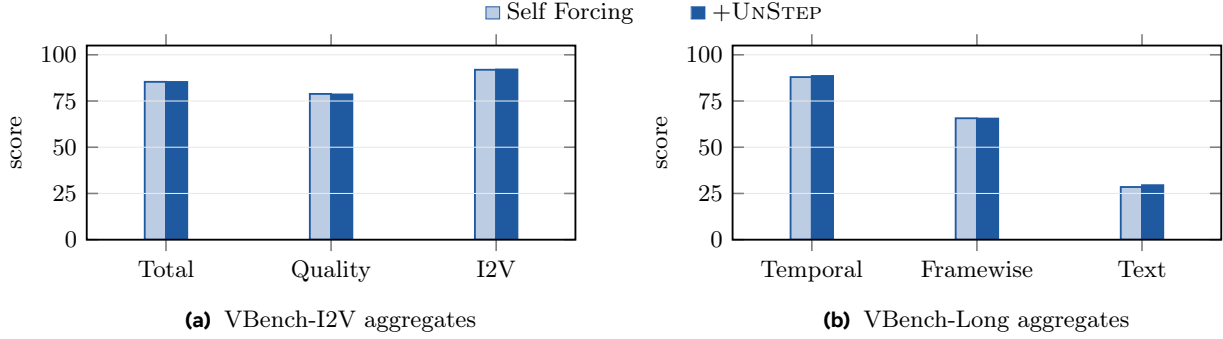
\begin{figure}[t]
\centering
\captionsetup[subfigure]{justification=centering,singlelinecheck=true}
\definecolor{gResult}{RGB}{31,90,160}
{\small \swatch{gResult}{gResult!30}\;Self Forcing \qquad \swatch{gResult}{gResult}\;$+$\textsc{UnStep}}\par\vspace{1mm}
\begin{subfigure}[t]{0.485\linewidth}
\centering
\begin{tikzpicture}[trim axis left,trim axis right]
\begin{axis}[
  width=\linewidth, height=4.15cm, ybar, bar width=8pt,
  ymin=0, ymax=105, ylabel={score}, ylabel style={font=\small},
  xtick={1,2,3}, xticklabels={Total,Quality,I2V},
  xticklabel style={font=\small}, ytick={0,25,50,75,100},
  yticklabel style={font=\small}, ymajorgrids, grid style={gray!18}, axis on top,
  enlarge x limits=0.24,
]
\addplot[fill=gResult!30, draw=gResult, bar shift=-4pt] coordinates {(1,85.37) (2,78.84) (3,91.90)};
\addplot[fill=gResult, draw=gResult, bar shift=4pt] coordinates {(1,85.30) (2,78.55) (3,92.05)};
\end{axis}
\end{tikzpicture}
\caption{VBench-I2V aggregates}
\label{fig:i2v}
\end{subfigure}\hfill
\begin{subfigure}[t]{0.485\linewidth}
\centering
\begin{tikzpicture}[trim axis left,trim axis right]
\begin{axis}[
  width=\linewidth, height=4.15cm, ybar, bar width=8pt,
  ymin=0, ymax=105, ylabel={score}, ylabel style={font=\small},
  xtick={1,2,3}, xticklabels={Temporal,Framewise,Text},
  xticklabel style={font=\small}, ytick={0,25,50,75,100},
  yticklabel style={font=\small}, ymajorgrids, grid style={gray!18}, axis on top,
  enlarge x limits=0.24,
]
\addplot[fill=gResult!30, draw=gResult, bar shift=-4pt] coordinates {(1,87.94) (2,65.68) (3,28.49)};
\addplot[fill=gResult, draw=gResult, bar shift=4pt] coordinates {(1,88.54) (2,65.49) (3,29.47)};
\end{axis}
\end{tikzpicture}
\caption{VBench-Long aggregates}
\label{fig:long}
\end{subfigure}
\caption{Aggregate image-to-video results for 5-second clips (left) and long video results for 30-second clips (right) on a single H100. Pale bars are Self Forcing and solid bars are the same checkpoint wrapped with \textsc{UnStep}. Appendix~\ref{app:extended-i2v-long} reports the full dimension scores.}
\label{fig:i2v-long}
\end{figure}

\subsection{Image-to-video}
\label{sec:exp:i2v}

We next test \textsc{UnStep} on image-to-video (I2V) tasks. In I2V, the model is conditioned on an input image in addition to the text prompt, so acceleration must preserve both generated video quality and fidelity to the conditioning image. We evaluate with VBench-I2V \citep{huang2024vbenchpp}, which reports Total, Quality, and I2V aggregates. Appendix~\ref{app:evaluation:i2v} details the I2V protocol.

We run the original Self Forcing checkpoint with and without \textsc{UnStep}, without I2V-specific retuning. Original Self Forcing runs at 17.0 FPS, while \textsc{UnStep} reaches 49.8 FPS. The quality results are shown in Figure~\ref{fig:i2v}. \textsc{UnStep} preserves video quality and conditioning fidelity while boosting throughput. Appendix~\ref{app:extended-i2v-long} reports all nine component dimensions in Figure~\ref{fig:i2v-components}.

\subsection{Long video}
\label{sec:exp:long}

All previous experiments use five-second clips, matching the standard VBench T2V setting. We next evaluate 30-second streaming generation. We evaluate on VBench-Long \citep{huang2024vbenchpp}, which reports temporal quality, framewise quality, and text alignment aggregates instead of the standard sixteen dimension VBench total. Figure~\ref{fig:long} compares these aggregates. Appendix~\ref{app:evaluation:long} details the protocol, and Appendix~\ref{app:extended-i2v-long} reports all eight component dimensions in Figure~\ref{fig:long-components}.

\textsc{UnStep} is again applied on Self Forcing without retuning. It reaches 51.2 FPS, close to its steady-state streaming speed while performance is maintained. Compared with five-second generation, \textsc{UnStep} is faster because a longer clip contains more later chunks with two denoising steps, so the initial chunk's four-step cost is spread over more generated frames. Self Forcing is slightly slower at 16.6 FPS because a larger fraction of the video is generated with its KV cache at the full five-second capacity. This shows that \textsc{UnStep} transfers to long streaming generation without specific retuning.

%% file: sections/theory.tex
\section{Toy Models for Inference-Only Recovery}
\label{sec:theory}

This section uses toy models to provide intuition why the refined clean-cache pass and reduced rank of the projection matrices can help reduce the error introduced by the shortened schedule. We plot the mean squared error (MSE) between full and shortened schedule latent vectors, using the toy models' mean risks and empirical measurements from the DiT. Figure~\ref{fig:theory_diagnostics} shows both plots as we vary the clean-cache noise level and retained V/O rank fraction. Appendix~\ref{app:theory-proofs} states the model, schedule, and noise assumptions used in the proofs.

\subsection{Clean-cache refinement}
\label{sec:theory:clean}

To analyze the clean-cache refinement, we compare the latent estimate produced by a shortened schedule with the estimate produced by the full schedule. We fix the text prompt and preceding KV cache. Let $\mathbf{x}_{\mathrm{full}},\mathbf{x}_{\mathrm{short}}\in\mathbb{R}^{d}$ be the random latent estimates obtained by recursively applying the flow matching updates in equations~\eqref{eq:clean_renoise}--\eqref{eq:clean_emit} from the same initial noise with the full and shortened schedules, respectively, before the clean-cache pass.

Let $\mathbf{x}_{\mathrm{clean}}(\sigma)$ denote the clean-cache output obtained from $\mathbf{x}_{\mathrm{short}}$ using equations~\eqref{eq:clean_renoise}--\eqref{eq:clean_emit} with $t_i=0$, $\sigma_i=\sigma\in[0,1]$, and Gaussian noise $\bm{\epsilon}_i\sim\mathcal{N}(\mathbf{0},\mathbf{I}_d)$ independent of both schedule estimates. Here $\mathbf{I}_d$ is the $d\times d$ identity matrix. Appendix~\ref{app:theory-proofs} derives $\mathbf{x}_{\mathrm{full}}$ and $\mathbf{x}_{\mathrm{short}}$ in terms of $\mathbf{A}$ and $\bm{\epsilon}$, and Appendix~\ref{app:proof-clean} derives $\mathbf{x}_{\mathrm{clean}}(\sigma)$ in terms of $\mathbf{A}$, $\bm{\epsilon}$, and $\sigma$. We model the DiT at every conditioning label, including the clean label, by the same linear map:
\begin{equation}
    f_{\bm{\theta}}(\mathbf{x},t)=\mathbf{A}\mathbf{x},
    \qquad \mathbf{A}=\mathbf{A}^{\top},
    \qquad \mathbf{0}\preceq\mathbf{A}\prec\mathbf{I}_d.
    \label{eq:theory_clean_linear_model}
\end{equation}
We define the residual as the difference between the estimate after the clean-cache pass and the full schedule estimate, and the risk as the expected squared residual:
\begin{equation}
    \mathbf{r}(\sigma)=\mathbf{x}_{\mathrm{clean}}(\sigma)-\mathbf{x}_{\mathrm{full}},
    \qquad
    R_{\mathrm{clean}}(\sigma)
    =\mathbb{E}
    \left[\|\mathbf{r}(\sigma)\|_2^2\right],
    \label{eq:theory_clean_risk}
\end{equation}
where the expectation is over $\bm{\epsilon}$. Using the linear model in equation~\eqref{eq:theory_clean_linear_model} and expanding the squared residual gives
\begin{equation}
\begin{aligned}
    R_{\mathrm{clean}}(\sigma)
    &=\mathbb{E}\|\mathbf{x}_{\mathrm{short}}-\mathbf{x}_{\mathrm{full}}\|_2^2-2g\sigma+\rho(\sigma),\\
    g
    &=\mathbb{E}\!\left[(\mathbf{x}_{\mathrm{short}}-\mathbf{x}_{\mathrm{full}})^\top(\mathbf{I}_d+\mathbf{A})\mathbf{x}_{\mathrm{short}}\right],
    \qquad |\rho(\sigma)|\leq k\sigma^2,
\end{aligned}
    \label{eq:theory_clean_local_risk}
\end{equation}
where $k$ is a finite positive scalar, and $g>0$ for the shortened schedule (proof in Appendix~\ref{app:proof-clean}).

Our objective is to choose $\sigma\in[0,1]$ that minimizes $R_{\mathrm{clean}}(\sigma)$. For the full schedule, the clean-cache pass at $\sigma=0$ leaves $\mathbf{x}_{\mathrm{full}}$ unchanged and gives zero risk. Every $\sigma>0$ gives positive risk, so $\sigma_{\mathrm{full}}^\star=0$ is the unique minimizer.

For the shortened schedule, let $\sigma_{\mathrm{short}}^\star\in\operatorname*{arg\,min}_{\sigma\in[0,1]}R_{\mathrm{clean}}(\sigma)$ denote a risk minimizer. The shortened schedule minimizer satisfies
\begin{equation}
    \sigma_{\mathrm{short}}^\star
    >\frac{1}{3}\min\!\left\{1,\frac{g}{k}\right\}>0.
    \label{eq:theory_clean_positive_minimizer}
\end{equation}
Appendix~\ref{app:proof-clean} gives the proof. This motivates \textsc{UnStep} to reuse the existing clean-cache pass with a tuned nonzero noise level after shortening the denoising schedule.

\begin{figure}[t]
\centering
\begin{subfigure}[t]{0.485\linewidth}
\centering
\begin{tikzpicture}
\definecolor{gCleanShort}{RGB}{205,102,20}
\definecolor{gCleanActual}{RGB}{0,135,105}
\begin{axis}[
  width=0.95\linewidth,
  height=5.0cm,
  xmin=0,
  xmax=0.05,
  ymin=0.0315,
  ymax=0.039,
  xlabel={clean-cache noise level $\sigma$},
  ylabel={MSE},
  scaled x ticks=false,
  scaled y ticks=false,
  xtick={0,0.01,0.02,0.03,0.04,0.05},
  xticklabels={0,0.01,0.02,0.03,0.04,0.05},
  ytick={0.032,0.034,0.036,0.038},
  yticklabels={0.032,0.034,0.036,0.038},
  tick label style={font=\scriptsize},
  label style={font=\small},
  legend style={draw=none, fill=none, font=\scriptsize, at={(0.5,1.03)}, anchor=south, legend columns=2, /tikz/every even column/.append style={column sep=5pt}},
  ymajorgrids,
  grid style={gray!18},
]
\addplot[thick, color=gCleanShort, domain=0:0.05, samples=201]
  {0.033958208553210988-0.234731262866781*x+6.4985404935318325*x^2-1.195854779603031*x^3+0.055266126345592317*x^4};
\addlegendentry{Toy model}
\addplot[thick, color=gCleanActual, mark=*, mark size=1.25pt]
  coordinates {
    (0.000,0.033996844664)
    (0.005,0.033027219100)
    (0.010,0.032221723793)
    (0.015,0.031989717791)
    (0.020,0.031863245944)
    (0.025,0.032112443015)
    (0.030,0.032723248470)
    (0.040,0.035022420578)
    (0.050,0.038235433508)
  };
\addlegendentry{DiT}
\addplot[only marks, mark=o, mark size=2.1pt, very thick, color=gCleanShort] coordinates {(0.018151145997,0.031831457539)};
\addplot[only marks, mark=o, mark size=2.3pt, very thick, color=gCleanActual]
  coordinates {(0.020,0.031863245944)};
\end{axis}
\end{tikzpicture}
\caption{Clean-cache noise level.}
\label{fig:theory_clean_toy}
\end{subfigure}
\hfill
\begin{subfigure}[t]{0.485\linewidth}
\centering
\begin{tikzpicture}
\definecolor{gShort}{RGB}{205,102,20}
\definecolor{gActual}{RGB}{0,135,105}
\begin{axis}[
  width=0.95\linewidth,
  height=5.0cm,
  xmin=0.86,
  xmax=1,
  ymin=0.036,
  ymax=0.047,
  xlabel={retained rank fraction $p$},
  ylabel={MSE},
  scaled y ticks=false,
  xtick={0.86,0.88,0.90,0.92,0.94,0.96,0.98,1.00},
  ytick={0.037,0.040,0.043,0.046},
  yticklabels={0.037,0.040,0.043,0.046},
  tick label style={font=\scriptsize},
  label style={font=\small},
  legend style={draw=none, fill=none, font=\scriptsize, at={(0.5,1.03)}, anchor=south, legend columns=2, /tikz/every even column/.append style={column sep=5pt}},
  ymajorgrids,
  grid style={gray!18},
]
\addplot[thick, color=gShort] coordinates {
(0.860,0.044788598146494)
(0.862,0.044484470025248)
(0.864,0.044182874879104)
(0.866,0.043883874087322)
(0.868,0.043587533165554)
(0.870,0.043293922209449)
(0.872,0.043003116405063)
(0.874,0.042715196619273)
(0.876,0.042430250086727)
(0.878,0.042148371214116)
(0.880,0.041869662528265)
(0.882,0.041594235802109)
(0.884,0.041322213402881)
(0.886,0.041053729920937)
(0.888,0.040788934157281)
(0.890,0.040527991575741)
(0.892,0.040271087366100)
(0.894,0.040018430324106)
(0.896,0.039770257844672)
(0.898,0.039526842465304)
(0.900,0.039288500623138)
(0.902,0.039055604666504)
(0.904,0.038828599819746)
(0.906,0.038608029006659)
(0.908,0.038394570794633)
(0.910,0.038189100698350)
(0.912,0.037992797699202)
(0.914,0.037807348860834)
(0.916,0.037635405402019)
(0.918,0.037481887628825)
(0.920,0.037360557203303)
(0.922,0.037402365598689)
(0.924,0.037444173994075)
(0.926,0.037485982389461)
(0.928,0.037527790784846)
(0.930,0.037569599180232)
(0.932,0.037611407575618)
(0.934,0.037653215971004)
(0.936,0.037695024366389)
(0.938,0.037736832761775)
(0.940,0.037778641157161)
(0.942,0.037820449552547)
(0.944,0.037862257947932)
(0.946,0.037904066343318)
(0.948,0.037945874738704)
(0.950,0.037987683134090)
(0.952,0.038029491529475)
(0.954,0.038071299924861)
(0.956,0.038113108320247)
(0.958,0.038154916715632)
(0.960,0.038196725111018)
(0.962,0.038238533506404)
(0.964,0.038280341901790)
(0.966,0.038322150297175)
(0.968,0.038363958692561)
(0.970,0.038405767087947)
(0.972,0.038447575483333)
(0.974,0.038489383878718)
(0.976,0.038531192274104)
(0.978,0.038573000669490)
(0.980,0.038614809064876)
(0.982,0.038656617460261)
(0.984,0.038698425855647)
(0.986,0.038740234251033)
(0.988,0.038782042646419)
(0.990,0.038823851041804)
(0.992,0.038865659437190)
(0.994,0.038907467832576)
(0.996,0.038949276227962)
(0.998,0.038991084623347)
(1.000,0.039032893018733)
};
\addlegendentry{Toy model}
\addplot[thick, color=gActual, mark=*, mark size=1.25pt]
  coordinates {
    (0.86,0.046019862600)
    (0.88,0.041425585780)
    (0.90,0.038667289467)
    (0.92,0.037046564267)
    (0.94,0.037714338717)
    (0.96,0.038433201958)
    (0.98,0.038604260884)
    (1.00,0.039045356094)
  };
\addlegendentry{DiT}
\addplot[only marks, mark=o, mark size=2.1pt, very thick, color=gShort]
  coordinates {(0.92,0.037360557203303)};
\addplot[only marks, mark=o, mark size=2.3pt, very thick, color=gActual]
  coordinates {(0.92,0.037046564267)};
\end{axis}
\end{tikzpicture}
\caption{Retained V/O rank fraction.}
\label{fig:theory_svd_toy}
\end{subfigure}
\caption{Mean toy risks (orange) and measured DiT latent MSE (green) as a function of clean-cache noise level $\sigma$ (left) and retained V/O rank fraction $p$ (right). The toy curves evaluate equation~\eqref{eq:theory_clean_risk} conditional on the initial vector and equation~\eqref{eq:theory_svd_risk}, respectively.}
\label{fig:theory_diagnostics}
\end{figure}
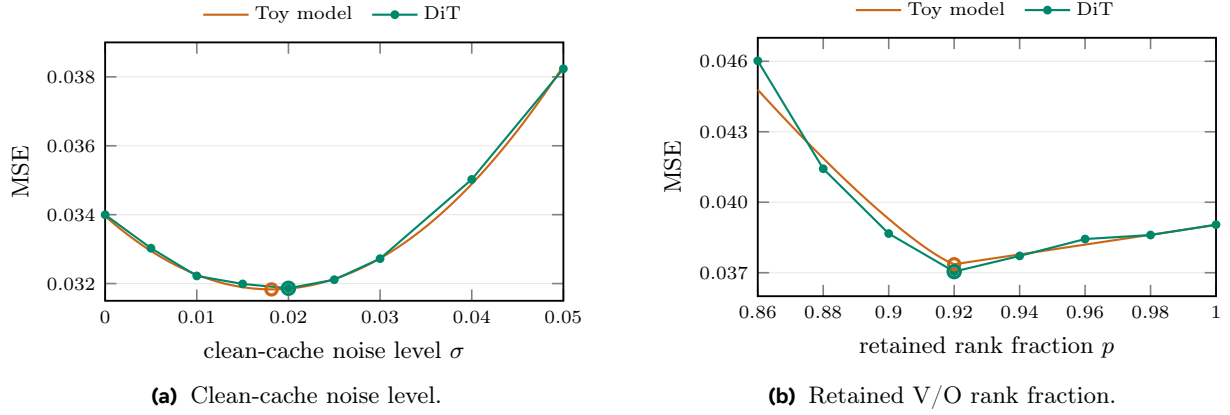

To visualize equation~\eqref{eq:theory_clean_risk}, Figure~\ref{fig:theory_clean_toy} compares the shortened schedule's mean toy risk for a fixed initial vector with the measured DiT MSE, computed between the full and shortened schedule latent vectors. Both the toy model and the DiT measurements have their lowest error at a nonzero noise level. Appendix~\ref{app:theory-proofs} expands the risk into the polynomial in equation~\eqref{eq:appendix_clean_risk_polynomial} and gives the fixed numerical parameters used to plot it.

\subsection{V/O truncated SVD}
\label{sec:theory:svd}

As in \S\ref{sec:theory:clean}, we use the latent estimates $\mathbf{x}_{\mathrm{full}}$ and $\mathbf{x}_{\mathrm{short}}$ and the linear DiT $f_{\bm{\theta}}(\mathbf{x},t)=\mathbf{A}\mathbf{x}$ under the matrix assumptions in equation~\eqref{eq:theory_clean_linear_model}, with the additional assumption that $\mathbf{A}$ is rank deficient, i.e., $\operatorname{rank}(\mathbf{A})<d$.

To model V/O truncation, we represent the part of the DiT being truncated by a fixed symmetric matrix $\mathbf{W}\in\mathbb{R}^{d\times d}$ and the part kept unchanged by $b\mathbf{I}_d$, where $b$ is a small positive number. We replace $\mathbf{W}$ by its rank-$r$ truncated SVD $\widehat{\mathbf{W}}_r$:
\begin{equation}
    \mathbf{A}=b\mathbf{I}_d+\mathbf{W},
    \qquad
    \mathbf{A}_r=b\mathbf{I}_d+\widehat{\mathbf{W}}_r,
    \qquad
    f_r(\mathbf{x},t)=\mathbf{A}_r\mathbf{x}.
    \label{eq:theory_svd_model}
\end{equation}
Let $\mathbf{x}_{\mathrm{full}}(r)$ and $\mathbf{x}_{\mathrm{short}}(r)$ denote the estimates obtained by the same denoising updates with $\mathbf{A}_r$ instead of $\mathbf{A}$. Appendix~\ref{app:proof-svd} applies the shared derivation in Appendix~\ref{app:theory-proofs} to these estimates. At $r=d$, these are the original estimates $\mathbf{x}_{\mathrm{full}}$ and $\mathbf{x}_{\mathrm{short}}$.

We define the risk as the expected squared difference between $\mathbf{x}_{\mathrm{short}}(r)$ and $\mathbf{x}_{\mathrm{full}}$:
\begin{equation}
    R_{\mathrm{svd}}(r)
    =\mathbb{E}\left[\|\mathbf{x}_{\mathrm{short}}(r)-\mathbf{x}_{\mathrm{full}}\|_2^2\right].
    \label{eq:theory_svd_risk}
\end{equation}
The expectation is over $\bm{\epsilon}$. The clean-cache noise level is fixed at zero for this toy rank comparison. Let $q=\operatorname{rank}(\mathbf{A})$. Equation~\eqref{eq:theory_svd_risk} can be written as
\begin{equation}
    R_{\mathrm{svd}}(r)
    =R_{\mathrm{svd}}(q)+
    \begin{cases}
        \displaystyle\sum_{r<\ell\leq q}\eta_{\ell}, & 0\leq r<q,\\[3pt]
        (r-q)\kappa, & q\leq r\leq d.
    \end{cases}
    \label{eq:theory_svd_risk_expansion}
\end{equation}
Here $\eta_{\ell}$ and $\kappa$ are strictly positive scalars derived from the expected squared errors after recursively applying equations~\eqref{eq:clean_renoise}--\eqref{eq:clean_emit} from the same initial noise, using $\mathbf{A}$ for the full schedule and $\mathbf{A}_r$ for the shortened schedule (Proof in Appendix~\ref{app:proof-svd}). 

Relative to $R_{\mathrm{svd}}(q)$, choosing $r<q$ adds $\eta_{\ell}$ for each discarded direction $r<\ell\leq q$, while choosing $r>q$ adds $\kappa$ for each additional retained direction. Since these added terms are positive for every $r\neq q$, equation~\eqref{eq:theory_svd_risk_expansion} is uniquely minimized at $r_{\mathrm{short}}^\star=q<d$. For the full schedule, replacing $\mathbf{x}_{\mathrm{short}}(r)$ by $\mathbf{x}_{\mathrm{full}}(r)$ in equation~\eqref{eq:theory_svd_risk} gives zero risk at $r=d$ and positive risk for every $r<d$, so $r_{\mathrm{full}}^\star=d$ is the unique minimizer. This motivates \textsc{UnStep} to use a tuned reduced rank for selected V/O projections after shortening the denoising schedule.

To visualize equation~\eqref{eq:theory_svd_risk}, Figure~\ref{fig:theory_svd_toy} compares the shortened schedule's mean toy risk, plotted against the retained rank fraction $p=r/d$, with the measured DiT MSE between full and shortened schedule latent vectors. Both the toy model and the DiT measurements have their lowest error below full retention. Appendix~\ref{app:theory-proofs} gives the fixed matrices, schedules, and risk calculation used for the toy curve.

%% file: sections/conclusion.tex
\section{Limitations}
\label{sec:discussion}

\paragraph{Single setup tuning:}
All \textsc{UnStep} parameters are tuned in one setting: Self Forcing as the base checkpoint, text-to-video generation, five-second VBench clips, and a single H100. We then keep the same denoising schedule, attention window and sink sizes, clean-cache noise level, SVD retention, and runtime stack choices fixed for all other experiments. This shows that the wrapper can transfer without retuning, but it also means the reported results are not necessarily optimal for every checkpoint, GPU, task, or clip length. Task-specific tuning could improve speed, quality, or both, and we leave that to future work.

\paragraph{Training-free constraint:}
Throughout the paper we maintain the training-free constraint and avoid applying gradient updates. This constraint makes the wrapper easy to apply to existing checkpoints, but it also limits the extent of acceleration we can reach. More aggressive sparse attention patterns, shorter schedules, quantization, or a smaller decoder could further improve speed, especially if followed by finetuning or distillation to recover quality. We view those training-based extensions as future work complementary to \textsc{UnStep}, but outside the scope of this paper.

\section{Conclusion}
\label{sec:conclusion}

We presented \textsc{UnStep}, a training-free inference wrapper for few-step causal video diffusion models. \textsc{UnStep} accelerates generation by combining runtime stack optimizations for the DiT and VAE decoder with reduced denoising steps and a limited KV attention window. It recovers the resulting quality loss using a refined clean-cache pass and V/O truncated SVD\iftheory{, two mechanisms that we motivate theoretically using toy models}. It enables streaming at 50 FPS on H100 with strong VBench quality, and up to 77 FPS on GB200. The same fixed wrapper transfers across checkpoints and to image-to-video generation and long video streaming, while also supporting a faster one-step configuration. \textsc{UnStep} delivers substantially faster causal video generation while maintaining strong quality, making existing models more practical for real-time applications without retraining.

\subsection*{Reproducibility statement}

Our experiments use publicly available Wan-2.1-1.3B distilled checkpoints such as Self Forcing \citep{huang2025selfforcing} and Causal Forcing \citep{zhu2026causalforcing}, which can be downloaded through their official project repositories. The benchmark data and evaluation code for VBench text-to-video, image-to-video, and long video evaluation are available through the \href{https://github.com/Vchitect/VBench}{VBench GitHub repository}. Appendix~\ref{app:implementation} specifies our model and inference configurations, and Appendix~\ref{app:evaluation} describes the evaluation, timing, and reproducibility protocols. Appendix~\ref{app:theory-proofs} provides the assumptions and proofs for our theoretical results. The \textsc{UnStep} code, together with all reproduction instructions, is available in our \href{https://github.com/facebookresearch/UnStep}{GitHub repository}, and our \href{https://y-mansour.github.io/UnStep/}{project page} contains sample generated  videos.

%% file: sections/appendix/implementation.tex
\section{Implementation Details and Ablations}
\label{app:implementation}

This section describes the model, sampling, and runtime configuration used by \textsc{UnStep}, and provides ablations. It specifies the denoising and clean-cache procedures, the KV attention window, the V/O SVD truncation, and the DiT and VAE runtime optimizations. \textsc{UnStep} performs no retraining or gradient updates. Table~\ref{tab:implementation-summary} summarizes the model and sampling configuration.

\begin{table}[htbp]
\centering
\small
\setlength{\tabcolsep}{5pt}
\begin{tabular}{@{}lllc@{}}
\toprule
Setting & Self Forcing & \textsc{UnStep} & \S \\
\midrule
Output clip & 81 frames, $832\times480$, 16 FPS & 81 frames, $832\times480$, 16 FPS
    & \multirow{7}{*}{\ref{app:implementation:setup}} \\
Initial noise shape & $1\times21\times16\times60\times104$ & $1\times21\times16\times60\times104$ & \\
Causal chunks & seven of three latent frames & seven of three latent frames & \\
First chunk & four denoising steps & four denoising steps & \\
Each later chunk & four denoising steps & two denoising steps & \\
Total DiT forwards (5-s clip) & 35 & 23 & \\
\midrule
Clean-cache DiT label & $t=0$ & $t=0$
    & \multirow{4}{*}{\ref{app:implementation:schedule}} \\
Clean-cache noise scale & $0$ & $\sigma_\star=0.02$ & \\
Clean-cache input & final denoised estimate & near clean renoised estimate & \\
Emitted latent & before clean-cache pass & output of clean-cache pass & \\
\midrule
Temporal self-attention & growing history & 9-frame limit
    & \multirow{2}{*}{\ref{app:implementation:attention}} \\
Protected cache prefix & none & first three latent frames & \\
\midrule
Attention SVD & none & V/O, layers 8--29, 92\% retained & \ref{app:implementation:svd} \\
\bottomrule
\end{tabular}
\caption{Overview of the original Self Forcing and default two-step \textsc{UnStep}
configurations for one 81-frame, 5-second clip. The forward counts include
one clean-cache step for each chunk.}
\label{tab:implementation-summary}
\end{table}

\subsection{Model and generation setup}
\label{app:implementation:setup}

The default setup follows the VBench 5s T2V output format \citep{huang2024vbench}: each clip contains 81 RGB frames at resolution $832\times480$ (width $\times$ height), played at 16 FPS. The backbone is the causal Wan2.1-T2V-1.3B Self Forcing checkpoint, which generates in the latent space of its Wan VAE. The VAE compresses the height and width of the RGB space by a factor of eight, giving a latent spatial size of $60\times104$. It compresses time by a factor of four, except that the first RGB frame is encoded separately, giving $1+(81-1)/4=21$ latent frames.

The VAE architecture uses 16 latent channels, so the initial Gaussian noise has shape $1\times21\times16\times60\times104$. Self Forcing divides this latent sequence into 7 chunks, each of size $1\times3\times16\times60\times104$, ordered in batch size, temporal frames, channels, height, and width. Each DiT forward pass emits one chunk with 3 latent frames.

For each of the seven chunks, Self Forcing runs four denoising steps followed by one clean-cache step, giving $7\times(4+1)=35$ DiT forwards for the 5-second clip. \textsc{UnStep} keeps four denoising steps for the first chunk, but uses two for each of the six later chunks, and also performs one clean-cache step for all chunks. It therefore uses $(4+1)+6\times(2+1)=23$ DiT forwards.

\subsection{Denoising schedules and clean-cache procedure}
\label{app:implementation:schedule}

Using the notation in \S~\ref{sec:method:clean}, $t_i$ is the DiT conditioning label and $\sigma_i$ is its noise level. Self Forcing begins with four equally spaced raw schedule coordinates $u_i\in\{1000,750,500,250\}$, corresponding to noise levels $\{1,0.75,0.5,0.25\}$. It then uses a shift five scheduler to map these values to $\{1,0.9375,0.8333,0.625\}$, respectively, concentrating the schedule at higher noise levels:

\begin{equation}
    \sigma_i=\frac{5u_i}{1000+4u_i},
    \qquad t_i=1000\sigma_i.
    \label{eq:appendix-shift}
\end{equation}

The shift remaps the timesteps leaving the endpoints $\{1000, 0\}$ unchanged. Table~\ref{tab:sf-schedule} gives the resulting schedule and the separate clean-cache settings.

\begin{table}[htbp]
\centering
\small
\begin{tabular}{@{}crcc@{}}
\toprule
Schedule index $i$ & Raw $u_i$ & Noise level $\sigma_i$ & DiT label $t_i$ \\
\midrule
0 & 1000 & $1$ & $1000$ \\
1 & 750  & $15/16=0.9375$ & $937.5$ \\
2 & 500  & $5/6\approx0.8333$ & $833.3$ \\
3 & 250  & $5/8=0.625$ & $625$ \\
\midrule
Clean SF & -- & $0$ & $0$ \\
Clean \textsc{UnStep} & -- & $0.02$ & $0$ \\
\bottomrule
\end{tabular}
\caption{Self Forcing's four denoising steps after shift five mapping and the clean-cache settings
used by Self Forcing and \textsc{UnStep}.}
\label{tab:sf-schedule}
\end{table}

Self Forcing uses schedule indices 0, 1, 2, and 3 for every chunk. \textsc{UnStep} uses the same four indices for the first chunk. For each later chunk, it uses index 0 followed by either index 2 or index 3, alternating between them because this performed better than always using index 2 or 3. The one-step variant of \textsc{UnStep} keeps the first chunk unchanged and uses only index 0 for later chunks, with no change to the clean-cache step. The input to every denoising step after the first is formed by renoising the previous step's clean prediction with fresh Gaussian noise at the new step's noise level, as in Equation~\eqref{eq:clean_renoise}.

After each chunk's last denoising step, both methods run the clean-cache DiT call in Equation~\eqref{eq:clean_forward} at $t=0$. Self Forcing uses $\sigma=0$, so Equations~\eqref{eq:clean_renoise} and~\eqref{eq:clean_emit} leave the latent unchanged, it discards the prediction and only updates the KV cache. \textsc{UnStep} instead uses $\sigma_\star=0.02$, and emits the updated latent from Equation~\eqref{eq:clean_emit}. Its KV cache is updated from the same near clean input.

Figure~\ref{fig:recovery-sweep-sigma} shows VBench total as we vary the clean-cache noise level $\sigma_\star$, keeping all other wrapper settings at their defaults.

\subsection{KV attention window}
\label{app:implementation:attention}

Original Self Forcing retains the keys and values of every generated latent frame, so its self-attention context grows by three latent frames after each chunk. For an 81-frame, 5-second clip, the final chunk can attend to all 21 latent frames. At $832\times480$, each $60\times104$ latent frame becomes $30\times52=1{,}560$ tokens after the DiT's $1\times2\times2$ patch embedding, giving Self Forcing a maximum context of $21\times1{,}560=32{,}760$ visual tokens.

\textsc{UnStep} instead caps the cache at nine latent frames, or 14,040 tokens. Once the cache is full, it contains three sink frames (the first three latent frames) and the six most recent frames, including the current three-frame chunk.

We found that further reducing the attention window from nine to eight frames in all layers increases speed but lowers quality. As a tradeoff, the default wrapper uses eight frames in some selected layers and nine in the remaining layers, preserving quality while slightly improving speed. The selected layers attend to two sink frames plus the same six recent frames.

\subsection{Layerwise V/O truncated SVD}
\label{app:implementation:svd}

The Wan2.1-T2V-1.3B DiT contains 30 transformer layers, indexed 0--29. \textsc{UnStep} leaves the self-attention query and key projections unchanged in every layer, and leaves all self-attention projections in layers 0--7 unchanged. In layers 8--29, each $1{,}536\times1{,}536$ value and output
weight $\mathbf{W}$ is replaced by the dense truncated SVD reconstruction in
Equation~\eqref{eq:svd} at rank

\begin{equation}
    r=\operatorname{round}(0.92\times1536)=1413.
\end{equation}

This layer selective truncation is motivated by LASER, which found that truncating early transformer layers can degrade performance, whereas improvements are typically concentrated later
\citep{sharma2024laser}.
%Intuitively, an early-layer edit is propagated through every subsequent layer, while restricting the edit to later layers perturbs a shorter portion of the network.

The decomposition and reconstruction are computed once in \texttt{float32} during wrapper initialization. Biases are copied unchanged, and the modified transformer is then cast to \texttt{bfloat16}. Each reconstructed weight matrix retains its original dense shape, so the projection's parameter count and FLOPs are unchanged.

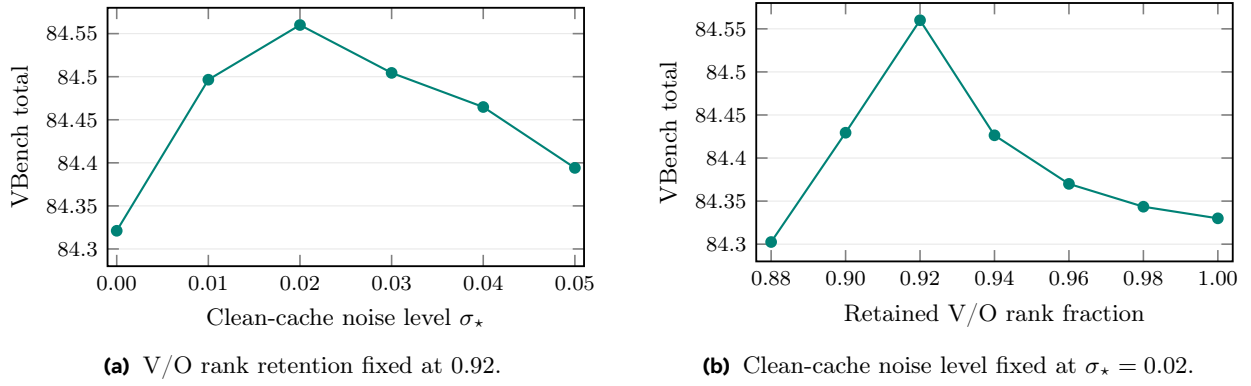
\begin{figure}[!t]
\centering
\pgfplotsset{recovery sweep/.style={
  width=0.98\linewidth,
  height=5.0cm,
  ymin=84.28,
  ymax=84.58,
  ylabel={VBench total},
  ytick={84.30,84.35,84.40,84.45,84.50,84.55},
  scaled ticks=false,
  tick label style={font=\footnotesize},
  label style={font=\small},
  ymajorgrids,
  grid style={gray!18},
}}
\begin{subfigure}[t]{0.485\linewidth}
\centering
\begin{tikzpicture}
\begin{axis}[
  recovery sweep,
  xmin=-0.001,
  xmax=0.051,
  xlabel={Clean-cache noise level $\sigma_\star$},
  xtick={0,0.01,0.02,0.03,0.04,0.05},
  xticklabels={0.00,0.01,0.02,0.03,0.04,0.05},
]
\addplot[thick, color=cFlex, mark=*, mark size=1.8pt] coordinates {
  (0.00,84.3211)
  (0.01,84.4965)
  (0.02,84.5600)
  (0.03,84.5044)
  (0.04,84.4648)
  (0.05,84.3941)
};
\end{axis}
\end{tikzpicture}
\caption{V/O rank retention fixed at 0.92.}
\label{fig:recovery-sweep-sigma}
\end{subfigure}
\hfill
\begin{subfigure}[t]{0.485\linewidth}
\centering
\begin{tikzpicture}
\begin{axis}[
  recovery sweep,
  xmin=0.876,
  xmax=1.004,
  xlabel={Retained V/O rank fraction},
  xtick={0.88,0.90,0.92,0.94,0.96,0.98,1.00},
  xticklabels={0.88,0.90,0.92,0.94,0.96,0.98,1.00},
]
\addplot[thick, color=cFlex, mark=*, mark size=1.8pt] coordinates {
  (0.88,84.3025)
  (0.90,84.4294)
  (0.92,84.5600)
  (0.94,84.4264)
  (0.96,84.3700)
  (0.98,84.3435)
  (1.00,84.3300)
};
\end{axis}
\end{tikzpicture}
\caption{Clean-cache noise level fixed at $\sigma_\star=0.02$.}
\label{fig:recovery-sweep-svd}
\end{subfigure}
\caption{VBench 5s T2V total for clean-cache noise level and V/O rank retention sweeps. Points show measured scores, joined by lines.}
\label{fig:recovery-sweeps}
\end{figure}

Figure~\ref{fig:recovery-sweep-svd} shows VBench total as we vary V/O rank retention, keeping all other wrapper settings at their defaults.

\subsection{Diffusion transformer runtime}
\label{app:implementation:dit-runtime}

As described in \S~\ref{sec:method:runtime}, Self Forcing stores the global and local KV cache end positions as GPU scalar tensors. The global position is the token index immediately after the most recently written chunk in the full video. The local position is the index immediately after its keys and values in the cache buffer. Repeated denoising calls overwrite the same chunk's cache entries without advancing these positions. Python reads these values to select where to write keys and values and which cached range to use. \textsc{UnStep} stores these indices on CPU and reuses each read within the attention call, avoiding GPU synchronization for these scalar reads. The keys and values remain on GPU. Self Forcing uses variable-length FlashAttention, which supports different sequence lengths within a batch without padding every sequence to the longest one. Its wrapper constructs extra GPU tensors containing the query and key sequence lengths and their start/end indices on every call, including at batch size one. For a single video, each sequence starts at index zero and its end boundary is the token count, one past the last token's index. \textsc{UnStep} uses fixed-length FlashAttention for this setting. The query and key tensors' sequence dimensions already specify their token counts, so this requires no separate GPU tensors for lengths or boundaries. This avoids their repeated allocation and the GPU operations used to construct them. The query and key lengths may differ and can change between calls as the KV cache grows.

\textsc{UnStep} also converts the scheduler's fixed timestep and noise level lookup tables to \texttt{float64}, copies them to the GPU, and caches them there. Each denoising step uses the cached scheduler tables to obtain the noise level $\sigma_i$ corresponding to its conditioning label $t_i$ and applies Equation~\eqref{eq:clean_emit} to convert the predicted flow into the clean latent estimate. Original Self Forcing also retains its native scheduler tables on GPU after the first renoising step, but converts them to \texttt{float64} again for each clean estimate update. \textsc{UnStep} additionally retains the \texttt{float64} copies, avoiding repeated dtype conversions.

The fused Triton RoPE kernel described in \S~\ref{sec:method:runtime} uses cached \texttt{float32} sine and cosine tables. Self Forcing already stores a base table of rotation coefficients for temporal and spatial positions. On every RoPE call, it selects the coefficients for the chunk's frame, row, and column positions and combines them into one coefficient vector per token. \textsc{UnStep} builds this  once and retains the resulting sine and cosine tables on GPU for each starting latent frame index and grid size. The starting index is the position of the chunk's first latent frame in the full video. For three-frame chunks, these indices are $0,3,6,\ldots$, so the chunk starting at $6$ uses temporal positions $6,7,8$ for RoPE. The grid size specifies the number of latent frames, spatial token rows, and spatial token columns in the chunk. These cached tables provide the coefficients for the current chunk's queries and newly computed keys in all 30 self-attention layers and denoising calls.

Wan2.1-1.3B reshapes each query and newly computed key, after linear projection and RMS normalization, into 12 attention heads of width 128. For one token and one head, we can write this vector as $\mathbf{x}=(x_0,\ldots,x_{127})$. The index $j\in\{0,\ldots,63\}$ selects the coordinate pair $(x_{2j},x_{2j+1})$. In the snippet below, \texttt{x\_even} and \texttt{x\_odd} are these two coordinates, while \texttt{cos} and \texttt{sin} are the cached RoPE coefficients for that token and pair, shared across heads:

\begin{lstlisting}[language=Python,keywordstyle=\ttfamily\mdseries]
x_even = tl.load(x_ptr + x_base, mask=mask, other=0.0).to(tl.float32)
x_odd = tl.load(x_ptr + x_base + x_stride_d, mask=mask, other=0.0).to(tl.float32)
cos = tl.load(cos_ptr + freq_base, mask=mask, other=1.0)
sin = tl.load(sin_ptr + freq_base, mask=mask, other=0.0)

y_even = x_even * cos - x_odd * sin
y_odd = x_even * sin + x_odd * cos
tl.store(out_ptr + out_base, y_even, mask=mask)
tl.store(out_ptr + out_base + out_stride_d, y_odd, mask=mask)
\end{lstlisting}

The snippet shows the complete arithmetic done by the Triton kernel used by \textsc{UnStep} for one coordinate pair. Original Self Forcing converts the input to \texttt{float64}, treats coordinate pairs as complex numbers, multiplies by the complex RoPE coefficients, and concatenates and stacks the results before converting back to \texttt{bfloat16}. These stages allocate and write intermediate tensors. For each query or key RoPE call, Triton loads the \texttt{bfloat16} input, rotates it in \texttt{float32}, and stores the \texttt{bfloat16} result in one kernel, avoiding those intermediate copies and launches.

The 30-second setting uses the same scheduler cache, Triton kernel, three-latent-frame chunks, and nine-frame KV limit as the 5-second setting. It executes more chunks at later latent frame positions and caches the corresponding sine and cosine tables for those chunks.

\subsection{VAE decoder runtime}
\label{app:implementation:vae-runtime}

The original Self Forcing baseline runs both the DiT and VAE in \texttt{bfloat16}. \textsc{UnStep} keeps the DiT in \texttt{bfloat16}, but casts the VAE weights to \texttt{float16} before compilation and the generated latents to \texttt{float16} immediately before decoding. Both methods rescale the normalized DiT latents to the scale expected by the VAE decoder using Wan's fixed per channel mean and standard deviation. Each channel uses the same constants across all temporal and spatial positions. The rescaling tensors are constructed once per video on the GPU, in \texttt{bfloat16} for Self Forcing and \texttt{float16} for \textsc{UnStep}, matching their decoder precision. Both methods convert the decoded output to \texttt{float32} before rescaling to $[0,1]$.

Self Forcing leaves the VAE's 3D convolution weights in PyTorch's default layout and passes decode inputs as a permuted $[B,C,T,H,W]$ view of the $[B,T,C,H,W]$ latent tensor, where $B,C,T,H,W$ denote batch size, channels, temporal length, height, and width. This permutation changes the axis order seen by the decoder without rearranging the stored values. \textsc{UnStep} stores both the convolution weights and decode inputs in the \texttt{channels\_last\_3d} memory format. For a decode input with logical axes $[B,C,T,H,W]$, this format uses $[B,T,H,W,C]$ storage, with channels as the innermost dimension, enabling faster H100 convolution kernels. \textsc{UnStep} also enables cuDNN benchmarking with a limit of 32.

Self Forcing appends each decoded temporal slice by allocating a larger output tensor and copying both the previously decoded output and the new slice into it. \textsc{UnStep} instead preallocates the output tensor once and writes each decoded slice into its assigned temporal range, avoiding repeated allocation and copying of earlier slices. The maximum temporal block size is selected to fit decoder memory, and is set to five latent frames in our setup. Both methods cache intermediate results from earlier frames during decoding. For 5-second clips, \textsc{UnStep} compiles the full VAE decoding code, including latent rescaling, the input convolution, decoder calls, and output writes. For long clips of 30 seconds or more, compiling all these operations for the full clip can exhaust GPU memory. \textsc{UnStep} instead compiles the decoder alone and calls it on successive temporal blocks, writing each block's decoded frames into the same preallocated output tensor.

\subsection{Recovery mechanisms with the full denoising schedule}
\label{app:implementation:full-schedule}

In this section we provide an ablation for the recovery mechanisms with the full denoising schedule. We keep Self Forcing's full four-step denoising schedule for every chunk and evaluate VBench 5s T2V with full attention as in default Self Forcing, or limited attention as in default \textsc{UnStep}, using its nine-latent-frame KV window (Appendix~\ref{app:implementation:attention}). Table~\ref{tab:full-schedule-recovery} compares clean-cache refinement and V/O SVD truncation, applied separately and together, with all other settings fixed within each attention setting.

\begin{table}[htbp]
\centering
\small
\setlength{\tabcolsep}{9pt}
\begin{tabular}{@{}lcccc@{}}
\toprule
Attention & Baseline & SVD truncation & \shortstack{Clean-cache\\refinement} & Both \\
\midrule
Full & \textbf{84.31} & 84.18 & 83.88 & 83.83 \\
Limited & 84.03 & \textbf{84.10} & 83.59 & 83.53 \\
\bottomrule
\end{tabular}
\caption{VBench total with the full denoising schedule. SVD retains 92\% of V/O rank, and clean-cache refinement uses $\sigma_\star=0.02$. The baseline uses $\sigma_\star=0$ and full rank. Bold marks the highest score per row.}
\label{tab:full-schedule-recovery}
\end{table}

Under the full schedule, these mechanisms provide little benefit and can reduce quality, with only a slight gain from SVD under limited attention. This contrasts with their combined benefit under the shortened schedule (Table~\ref{tab:methods}).

%% file: sections/appendix/evaluation.tex
\section{Evaluation and Reproducibility Details}
\label{app:evaluation}

\subsection{VBench 5s text-to-video evaluation}
\label{app:evaluation:t2v}

We evaluate our 5-second T2V outputs on the standard VBench prompt suite \citep{huang2024vbench}, which contains 946 text prompt entries assigned across 16 dimensions. Seven quality dimensions score the generated video without using prompt text, while nine semantic dimensions evaluate whether the video satisfies the semantic content of the original VBench prompt. We generate one 81-frame, $832\times480$ video at 16 FPS for each entry.

Following Self Forcing \citep{huang2025selfforcing}, we use its extended prompts for generation but use the corresponding original VBench prompts for semantic evaluation. Self Forcing produced the extended prompts with Qwen2.5-7B-Instruct using Wan2.1's English prompt extension instruction. For example, the original prompt ``a suitcase and a vase'' becomes ``A well-lit, detailed close-up of a vintage leather suitcase and a delicate glass vase placed side by side on a wooden table\ldots''

Each VBench dimension is evaluated on the prompts assigned to that dimension. The seven quality dimensions are subject consistency, background consistency, temporal flickering, motion smoothness, dynamic degree, aesthetic quality, and imaging quality. Their evaluators use DINO ViT-B/16, CLIP ViT-B/32, consecutive frame pixel differences, AMT-S frame interpolation, RAFT optical flow, the LAION aesthetic predictor with CLIP ViT-L/14 features, and MUSIQ-SPAQ, respectively. The nine semantic dimensions are object class, multiple objects, human action, color, spatial relationship, scene, appearance style, temporal style, and overall consistency. GRiT evaluates object class, multiple objects, color, and spatial relationships. UMT evaluates human actions. Tag2Text evaluates scenes. CLIP evaluates appearance style. ViCLIP evaluates temporal style and overall consistency.

We use the official \texttt{vbench\_standard} metric implementations and aggregation. VBench first normalizes each dimension. After that quality and semantic scores are calculated as weighted means of their seven and nine dimensions, respectively. Dynamic degree has weight $0.5$ and every other dimension has weight $1$. The total is $(4\,\mathrm{Quality}+\mathrm{Semantic})/5$, and all reported scores are multiplied by 100.

\subsection{Image-to-video evaluation}
\label{app:evaluation:i2v}

VBench-I2V \citep{huang2024vbenchpp} contains 1,118 image--prompt entries derived from 355
reference images: 246 subject entries, 109 background entries, and seven camera motion instructions
for each background image, giving 763 camera motion entries. We generate one 5-second clip per prompt using the English prompt and the official 16:9 reference image crop, with the same $832\times480$ resolution and 16 FPS as T2V. The \textsc{UnStep} denoising schedule, KV cache policy, clean-cache setting, SVD, and runtime optimizations are applied without I2V-specific retuning.

VBench-I2V reports three main aggregates: Total, Quality, and I2V. Quality aggregates six video quality dimensions: subject consistency, background consistency, motion smoothness, dynamic degree, aesthetic quality, and imaging quality. I2V aggregates three conditioning fidelity dimensions: video--image subject fidelity, video--image background fidelity, and video--text camera motion adherence. Subject and background fidelity use DINO ViT-B/16 and DreamSim features, respectively. Each score combines the maximum similarity between the reference and a generated frame with the mean and minimum similarities between consecutive video frames, using weights $0.4,0.3,0.3$. CoTracker2 checks whether the generated motion matches the prompt's camera instruction: pan left or right, tilt up or down, zoom in or out, or static.

After normalization, the I2V score weights subject and background fidelity by $1$ and camera motion by $0.1$. The Quality score weights dynamic degree by $0.5$ and its other five dimensions by $1$. The Total score is the equal average of I2V and Quality. Figure~\ref{fig:i2v} reports the aggregates: Total, Quality, and I2V, and Figure~\ref{fig:i2v-components} reports their nine component dimensions on a 0--100 scale.

For measuring I2V throughput, we follow the same per clip timing region as in Appendix~\ref{app:evaluation:reproducibility}. The reference image is preprocessed and VAE-encoded before the timer begins. The timed region covers generation conditioned on that encoded latent and VAE decoding of the complete video. As before, the first clip speed is not considered. Scoring on the other hand uses the standard VBench-I2V evaluator explained above.

\subsection{Long video evaluation}
\label{app:evaluation:long}

We use the 40-prompt long video competition track from VBench++
\citep{huang2024vbenchpp}. We generate one video per prompt with 123 latent frames in chunks of
three, yielding 489 RGB frames, or 30.56 seconds at 16 FPS. 
\textsc{UnStep} retains the T2V denoising schedule, KV cache limit, sink size, and SVD configuration
without retuning. Its long video VAE compilation path is described in
Appendix~\ref{app:implementation:vae-runtime}. \textsc{UnStep}'s throughput timer uses the same
GPU generation region as T2V. The reported long video FPS excludes the first clip, which includes compilation and autotuning, and averages the remaining
measurements.

VBench-Long reports three main aggregates instead of a combined total: Temporal Quality, Framewise Quality, and Text Alignment. These aggregates are formed from eight component metrics. The scoring pipeline uses PySceneDetect to detect scene cuts and divides each resulting scene into 2-second clips. A final partial clip is extended with preceding frames, while a scene shorter than two seconds is retained as one clip. Each metric is first averaged across clips from the same video and then across the 40 videos. Temporal Quality combines DINO subject consistency, CLIP background consistency, AMT-S motion smoothness, and RAFT dynamic degree with weights $1,1,1,0.5$. Our configuration uses the within clip consistency scores and gives the optional cross clip term zero weight. Framewise Quality is the equal average of the LAION aesthetic and MUSIQ-SPAQ imaging scores. Text Alignment is the equal average of ViCLIP overall consistency and CLIP ViT-B/32 frame--text similarity. Figure~\ref{fig:long} reports the aggregates: Temporal Quality, Framewise Quality, and Text Alignment, and Figure~\ref{fig:long-components} reports their eight component metrics on a 0--100 scale.

\subsection{Reproducibility and timing}
\label{app:evaluation:reproducibility}

We divide the 946 benchmark entries into eight fixed groups (shards) of consecutive entries. At the start of each
shard, the process resets the global random number generator (RNG), which produces the Gaussian
noise used during sampling, to seed zero. For every entry preceding the shard in benchmark order, \textsc{UnStep} consumes
its initial noise draw, while the upstream Self Forcing baseline advances the random draws made by
its original inference path. This defines a deterministic random sequence under the fixed
logical shard protocol. The eight shards can run concurrently on eight GPUs as one per GPU, or in
waves on fewer GPUs. Preserving the logical shard count and boundaries preserves prompt assignment
and RNG consumption. Each video is generated on one GPU, so the reported FPS remains single GPU
throughput.

For \textsc{UnStep}, each FPS measurement uses CUDA synchronization at both boundaries and
includes all DiT calls, KV cache updates, VAE decoding, and final GPU postprocessing.
Model and wrapper loading, compilation, and text encoding occur before the timer starts.

The first generated clip on each GPU completes compilation and autotuning and serves as
warmup. We report the average FPS over all remaining clips, starting from the second clip on each GPU. The timings therefore exclude the first prompt, but the quality scoring includes all 946 prompts.

Our H100 runs use PyTorch \texttt{2.15.0a0+git65cd5f0}, CUDA runtime 13.3 (toolkit 13.3.1), cuDNN 9.25.1, Triton 3.8.0, and FlashAttention-3. For \textsc{UnStep}, the DiT and the Wan VAE's \texttt{decode} and \texttt{cached\_decode} functions use \texttt{torch.compile} in \texttt{max-autotune-no-cudagraphs} mode, and cuDNN benchmarking is enabled with \texttt{torch.backends.cudnn.benchmark\_limit = 32}. Our GB200 runs follow the same setup, except that we use FlashAttention-2 instead of FlashAttention-3.

%% file: sections/appendix/extended_results.tex
\section{Extended Results}
\label{app:extended-results}

\subsection{Run-to-run variation}
\label{app:run-variation}

As described in Appendix~\ref{app:evaluation:reproducibility}, repeated runs use the same generation seed and sharding protocol, with the checkpoint and sampling configuration also fixed. Despite these fixed settings, scores vary between executions. Therefore all VBench quality scores reported in the paper are the mean of five runs. For the \textsc{UnStep} default two-step version on the Self Forcing checkpoint, Table~\ref{tab:run-variation} reports the mean and standard deviation of the Total, Quality, and Semantic scores over the full 946-prompt VBench benchmark.

% Six supplied run-level scores, ordered as Total, Quality, Semantic.
% 84.3939, 84.9984, 81.9760
% 84.6649, 85.3034, 82.1109
% 84.5994, 85.2645, 81.9389
% 84.5217, 85.1732, 81.9159
% 84.6197, 85.2889, 81.9430
% 84.5803, 85.1708, 82.2181
% Sample standard deviations use denominator n - 1, with n = 6.
\begin{center}
\begin{minipage}{\linewidth}
\centering
\small
\begin{tabular}{@{}lrr@{}}
\toprule
Metric & Mean & Standard deviation \\
\midrule
Total    & 84.56 & 0.10 \\
Quality  & 85.20 & 0.11 \\
Semantic & 82.02 & 0.12 \\
\bottomrule
\end{tabular}
\captionsetup{hypcap=false}
\captionof{table}{Run-to-run variation for Self Forcing $+$ \textsc{UnStep} with two denoising steps on H100.}
\label{tab:run-variation}
\end{minipage}
\end{center}

\subsection{Figure~\ref{fig:speed} comparison details}
\label{app:speed-sources}

\noindent\begin{minipage}{\linewidth}
This section provides more details on the configurations in Figure~\ref{fig:speed}. Table~\ref{tab:speed-h100-main} lists the H100 comparison without the one-step configurations. We consider methods that target fast generation and report VBench scores as well as end-to-end GPU throughput, including both the DiT and VAE. All FPS and VBench scores are taken from the cited papers.

\vspace{8pt}
\begin{center}
\footnotesize
\begin{tabular*}{\linewidth}{@{\extracolsep{\fill}}>{\raggedright\arraybackslash}p{3.45cm}>{\raggedright\arraybackslash}p{1.55cm}rrrr>{\raggedright\arraybackslash}p{3.5cm}@{}}
\toprule
Method & Venue & FPS & Total & Quality & Semantic & Dominant speed gain \\
\midrule
Causal Forcing $+$ \textsc{UnStep}, 2 steps
  & This work & 49.8 & 84.93 & 85.79 & 81.46 & Runtime $+$ steps/KV \\
Self Forcing $+$ \textsc{UnStep}, 2 steps
  & This work & 49.8 & 84.56 & 85.20 & 82.02 & Runtime $+$ steps/KV \\
\midrule
DiagDistill \citep{diagdistill}
  & ICLR'26 & 31.0 & 84.48 & 85.26 & 81.73 & Tiny decoder \\
Hybrid Forcing \citep{hybridforcing}
  & arXiv & 29.5 & 83.60 & 84.45 & 80.19 & Sparse/linear attention \\
Causal-rCM \citep{zheng2026causalrcm}
  & arXiv & 25.6 & 84.24 & 84.96 & 81.36 & Fewer denoising steps \\
Dummy Forcing \citep{guo2026dummyhead}
  & arXiv & 24.3 & 83.90 & 84.63 & 80.98 & Reduced KV attention \\
Reward Forcing \citep{rewardforcing}
  & CVPR'26 & 23.1 & 84.13 & 84.84 & 81.32 & Reduced KV attention \\
rCM, 2 steps \citep{zheng2025rcm}
  & ICLR'26 & 23.0 & 84.09 & 84.90 & 80.86 & Fewer denoising steps \\
DSA \citep{dsa}
  & CVPR'26 & 22.6 & 83.73 & 84.75 & 79.67 & Adaptive denoising steps \\
Ms. Forcing \citep{msforcing}
  & arXiv & 21.9 & 83.89 & 84.61 & 81.01 & Multiple patch sizes \\
MAG \citep{mag}
  & arXiv & 21.7 & 83.52 & 84.11 & 81.14 & KV compression \\
Sparse Forcing \citep{xu2026sparseforcing}
  & arXiv & 19.9 & 83.99 & 84.65 & 81.36 & Sparse attention \\
Light Forcing \citep{lv2026lightforcing}
  & ICML'26 & 18.9 & 84.50 & 85.40 & 80.90 & Sparse attention \\
In-Context Forcing \citep{incontextforcing}
  & arXiv & 18.4 & 84.34 & 84.99 & 81.77 & Parallel causal sampling \\
AnyFlow \citep{gu2026anyflow}
  & arXiv & 17.4 & 84.31 & 85.15 & 80.94 & Fewer denoising steps \\
\bottomrule
\end{tabular*}
\captionsetup{hypcap=false}
\captionof{table}{Details on the entries in the H100 panel of Figure~\ref{fig:speed} (without one-step models).}
\label{tab:speed-h100-main}
\end{center}
\end{minipage}
\par\medskip

\noindent\begin{minipage}{\linewidth}
Table~\ref{tab:speed-h100-one-step} lists the one-step H100 configurations. We use the VBench scores reported in the original papers for all methods. We measured end-to-end H100 throughput for both One-Forcing variants \citep{oneforcing2026} and Causal Forcing++ \citep{zhao2026causalforcingpp}, because One-Forcing reports no throughput and Causal Forcing++ reports only DiT throughput on A800. The rCM throughput is taken from its paper.

\vspace{8pt}
\begin{center}
\footnotesize
\begin{tabular*}{\linewidth}{@{\extracolsep{\fill}}llrrrr@{}}
\toprule
Method & Venue & FPS & Total & Quality & Semantic \\
\midrule
Self Forcing $+$ \textsc{UnStep}, 1 step
  & This work & 59.6 & 83.34 & 83.84 & 81.34 \\
Causal Forcing $+$ \textsc{UnStep}, 1 step
  & This work & 59.6 & 83.10 & 83.81 & 80.27 \\
\midrule
rCM, 1 step \citep{zheng2025rcm}
  & ICLR'26 & 32.3 & 82.65 & 83.60 & 78.82 \\
One-Forcing, 1 step (chunkwise) \citep{oneforcing2026}
  & arXiv & 25.2 & 81.60 & 83.65 & 73.41 \\
One-Forcing, 1 step (framewise) \citep{oneforcing2026}
  & arXiv & 20.7 & 83.76 & 85.22 & 77.91 \\
Causal Forcing++, 1 step \citep{zhao2026causalforcingpp}
  & arXiv & 20.4 & 83.35 & 84.50 & 78.75 \\
\bottomrule
\end{tabular*}
\captionsetup{hypcap=false}
\captionof{table}{One-step models in the H100 panel of Figure~\ref{fig:speed}.}
\label{tab:speed-h100-one-step}
\end{center}
\end{minipage}
\par\medskip

\noindent\begin{minipage}{\linewidth}
Table~\ref{tab:speed-gb200} lists the GB200 configurations. Flex-Forcing uses larger chunks, which denoises more latent frames per emitted chunk and therefore reduces the number of DiT forward passes. Attention is bidirectional within each chunk and causal between chunks. Since the chunk sizes affect speed and quality, the paper searches different configurations of the 21 latent frames in a 5-second video and selects the best based on VBench scores and throughput. LongLive 2.0 quantizes weights, activations, and the KV cache to NVFP4. Its GB200 configuration uses an additional GPU for VAE decoding, which decodes one chunk while the DiT denoises the next. Both methods remain slower than \textsc{UnStep}.

\vspace{8pt}
\begin{center}
\footnotesize
\begin{tabular*}{\linewidth}{@{\extracolsep{\fill}}>{\raggedright\arraybackslash}p{3.45cm}>{\raggedright\arraybackslash}p{1.55cm}rrrr>{\raggedright\arraybackslash}p{3.5cm}@{}}
\toprule
Method & Venue & FPS & Total & Quality & Semantic & Dominant speed gain \\
\midrule
Self Forcing $+$ \textsc{UnStep}, 2 steps
  & This work & 63.7 & 84.36 & 85.06 & 81.56 & Runtime $+$ steps/KV \\
Causal Forcing $+$ \textsc{UnStep}, 2 steps
  & This work & 63.7 & 84.59 & 85.50 & 80.97 & Runtime $+$ steps/KV \\
Self Forcing $+$ \textsc{UnStep}, 1 step
  & This work & 77.4 & 83.49 & 84.09 & 81.08 & Runtime $+$ steps/KV \\
Causal Forcing $+$ \textsc{UnStep}, 1 step
  & This work & 77.4 & 83.16 & 84.15 & 79.22 & Runtime $+$ steps/KV \\
\midrule
Flex-Forcing (12--6--3) \citep{flexforcing}
  & ICML'26 & 45.6 & 84.29 & 85.33 & 80.12 & Chunk schedule search \\
Flex-Forcing (7--7--7) \citep{flexforcing}
  & ICML'26 & 48.9 & 83.92 & 84.96 & 79.78 & Chunk schedule search \\
Flex-Forcing (3--3--3--3--3--3--3) \citep{flexforcing}
  & ICML'26 & 41.5 & 83.91 & 84.85 & 80.16 & Chunk schedule search \\
LongLive 2.0 \citep{longlive2}
  & arXiv & 45.7 & 83.14 & 85.40 & 74.12 & NVFP4 quantization / additional VAE GPU \\
\bottomrule
\end{tabular*}
\captionsetup{hypcap=false}
\captionof{table}{Points in the GB200 panel of Figure~\ref{fig:speed}.}
\label{tab:speed-gb200}
\end{center}
\end{minipage}
\par\medskip

\subsection{Reproduced Baselines}
\label{app:baseline-reproduction}

To evaluate \textsc{UnStep} on Self Forcing, Causal Forcing and LongLive 1.0 as shown in Table \ref{tab:othermodels}, we evaluate the VBench score of each checkpoint before and after applying the wrapper. We  evaluate all checkpoints under the same 946-prompt protocol in Appendix~\ref{app:evaluation}. 

We reproduce Self Forcing's published total, with slightly different quality and semantic scores. Our reproduced Causal Forcing total is higher than the original paper's report, while our LongLive 1.0 total is lower. LongLive's official repository does not provide a VBench evaluation command or the prompt preprocessing used for that result, therefore we could not reproduce the reported score. However, our LongLive 1.0 total remains in the 83-point range reported by other recent work \citep{lv2026lightforcing,zheng2026causalrcm,rewardforcing,ji2026forcingkv,videossm}.

\subsection{Full text-to-video dimension scores}
\label{app:t2v}

Table~\ref{tab:t2v-dimensions} reports all 16 normalized VBench T2V dimensions for Self Forcing and Causal
Forcing before and after applying \textsc{UnStep}. Each delta compares runs generated and scored
with the same protocol, using the aggregation in Appendix~\ref{app:evaluation:t2v}. For Self
Forcing, \textsc{UnStep} changes the quality aggregate by
$-0.08$ and the semantic aggregate by $+1.58$. For Causal Forcing, the corresponding changes are
$+0.16$ and $-0.09$. The total scores increase by $0.25$ and $0.11$, respectively, but individual dimensions move in both directions.

\begin{table}[htbp]
\centering
\footnotesize
\setlength{\tabcolsep}{3pt}
\begin{tabular}{@{}lccc@{\hspace{10pt}\vrule width 0.4pt\hspace{10pt}}ccc@{}}
\toprule
& SF & SF $+$ \textsc{UnStep} & $\Delta_{\mathrm{SF}}$ & CF & CF $+$ \textsc{UnStep} & $\Delta_{\mathrm{CF}}$ \\
\midrule
\multicolumn{7}{@{}l}{\emph{Quality dimensions}} \\
Subject consistency    & 94.94 & 95.10 & $+0.16$ & 94.99 & 94.67 & $-0.32$ \\
Background consistency & 95.37 & 95.16 & $-0.21$ & 94.73 & 95.04 & $+0.31$ \\
Temporal flickering    & 97.64 & 97.64 & $0.00$ & 95.13 & 96.79 & $+1.66$ \\
Motion smoothness      & 95.09 & 95.53 & $+0.44$ & 92.07 & 93.19 & $+1.12$ \\
Dynamic degree         & 71.31 & 69.79 & $-1.52$ & 86.11 & 84.72 & $-1.39$ \\
Aesthetic quality      & 66.11 & 66.19 & $+0.08$ & 66.16 & 66.20 & $+0.04$ \\
Imaging quality        & 69.52 & 69.29 & $-0.23$ & 70.50 & 69.40 & $-1.10$ \\
\midrule
\multicolumn{7}{@{}l}{\emph{Semantic dimensions}} \\
Object class           & 94.17 & 96.00 & $+1.83$ & 95.97 & 94.09 & $-1.88$ \\
Multiple objects       & 85.35 & 87.52 & $+2.17$ & 89.71 & 93.79 & $+4.08$ \\
Human action           & 94.54 & 96.00 & $+1.46$ & 96.00 & 95.25 & $-0.75$ \\
Color                  & 89.33 & 91.00 & $+1.67$ & 89.24 & 90.13 & $+0.89$ \\
Spatial relationship   & 81.24 & 84.11 & $+2.87$ & 79.26 & 76.88 & $-2.38$ \\
Scene                  & 70.26 & 70.05 & $-0.21$ & 70.54 & 70.07 & $-0.46$ \\
Appearance style       & 70.41 & 71.57 & $+1.16$ & 71.93 & 71.21 & $-0.73$ \\
Temporal style         & 65.47 & 67.10 & $+1.63$ & 68.36 & 68.45 & $+0.09$ \\
Overall consistency    & 73.19 & 74.82 & $+1.63$ & 72.95 & 73.26 & $+0.32$ \\
\midrule
Quality aggregate      & 85.28 & 85.20 & $-0.08$ & 85.64 & 85.79 & $+0.16$ \\
Semantic aggregate     & 80.44 & 82.02 & $+1.58$ & 81.55 & 81.46 & $-0.09$ \\
Total                  & 84.31 & 84.56 & $+0.25$ & 84.82 & 84.93 & $+0.11$ \\
\bottomrule
\end{tabular}
\caption{Normalized VBench dimension scores from runs over all 946 prompts. SF and CF denote Self Forcing
and Causal Forcing. Each $\Delta$ is the wrapped score minus its corresponding unwrapped score.}
\label{tab:t2v-dimensions}
\end{table}

\subsection{Full image-to-video and long video dimension scores}
\label{app:extended-i2v-long}

Figures~\ref{fig:i2v-components} and~\ref{fig:long-components} expand the aggregate results in Figure~\ref{fig:i2v-long} into the component dimensions defined in Appendix~\ref{app:evaluation:i2v} and~\ref{app:evaluation:long}. Pale bars are Self Forcing and solid bars are the same checkpoint wrapped with \textsc{UnStep}.

\begin{center}
\begin{minipage}{\linewidth}
\centering
\begin{tikzpicture}
\definecolor{gAgg}{RGB}{31,90,160}\definecolor{gCond}{RGB}{150,60,150}\definecolor{gVid}{RGB}{205,102,20}
\begin{axis}[
  width=\linewidth, height=4.6cm, ybar, bar width=6pt,
  ymin=0, ymax=112, clip=false, ylabel={VBench-I2V score}, ylabel style={font=\normalsize},
  xmin=0.4, xmax=12.6, xtick={1,...,12},
  xticklabels={\emph{total},\emph{quality},\emph{I2V},I2V subj.,I2V bkg.,Camera,Subject,Background,Aesthetic,Imaging,Motion,Dynamic},
  xticklabel style={rotate=42, anchor=east, font=\footnotesize},
  ytick={0,25,50,75,100}, yticklabel style={font=\small},
  ymajorgrids, grid style={gray!18}, axis on top,
]
\addplot[fill=gAgg!30, draw=gAgg, bar shift=-3pt, forget plot] coordinates {(1,85.37) (2,78.84) (3,91.90)};
\addplot[fill=gAgg, draw=gAgg, bar shift=3pt, forget plot] coordinates {(1,85.30) (2,78.55) (3,92.05)};
\addplot[fill=gCond!30, draw=gCond, bar shift=-3pt, forget plot] coordinates {(4,94.50) (5,95.34) (6,31.59)};
\addplot[fill=gCond, draw=gCond, bar shift=3pt, forget plot] coordinates {(4,94.38) (5,95.42) (6,34.99)};
\addplot[fill=gVid!30, draw=gVid, bar shift=-3pt, forget plot] coordinates {(7,89.70) (8,92.65) (9,61.79) (10,70.97) (11,97.17) (12,42.68)};
\addplot[fill=gVid, draw=gVid, bar shift=3pt, forget plot] coordinates {(7,89.39) (8,92.76) (9,61.55) (10,70.70) (11,97.28) (12,40.65)};
\node[anchor=south, font=\normalsize, inner sep=1.5pt, yshift=3pt] at (rel axis cs:0.5,1.0)
  {\swatch{gAgg}{gAgg!30}\swatch{gCond}{gCond!30}\swatch{gVid}{gVid!30}\;Self Forcing,
   $17.0$~FPS \qquad
   \swatch{gAgg}{gAgg}\swatch{gCond}{gCond}\swatch{gVid}{gVid}\;$+$\textsc{UnStep},
   $49.8$~FPS};
\end{axis}
\end{tikzpicture}
\captionsetup{hypcap=false}

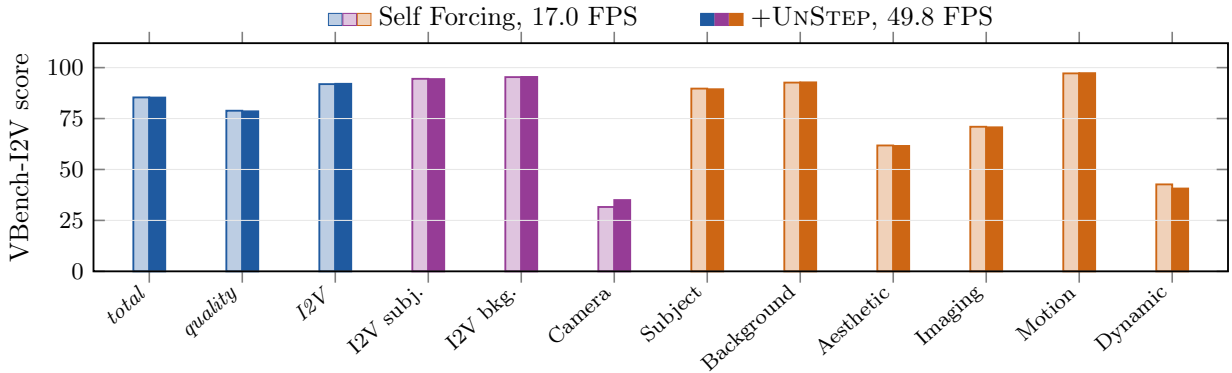
\captionof{figure}{Full VBench-I2V results on 1,118 image--prompt entries with 16:9 crops on one H100. The first three bars are the Total, Quality, and I2V aggregates. The remaining bars are their component dimensions. }
\label{fig:i2v-components}
\end{minipage}
\end{center}

\begin{center}
\begin{minipage}{\linewidth}
\centering
\begin{tikzpicture}
\definecolor{gTmp}{RGB}{31,90,160}\definecolor{gFrm}{RGB}{205,102,20}\definecolor{gTxt}{RGB}{130,50,150}
\begin{axis}[
  width=\linewidth, height=4.6cm, ybar, bar width=6pt,
  ymin=0, ymax=112, clip=false, ylabel={VBench-Long score}, ylabel style={font=\normalsize},
  xmin=0.4, xmax=11.6, xtick={1,...,11},
  xticklabels={\emph{temporal},Subject,Background,Motion,Dynamic,\emph{framewise},Aesthetic,Imaging,\emph{text},Overall,CLIP},
  xticklabel style={rotate=42, anchor=east, font=\footnotesize},
  ytick={0,25,50,75,100}, yticklabel style={font=\small},
  ymajorgrids, grid style={gray!18}, axis on top,
]
\addplot[fill=gTmp!30, draw=gTmp, bar shift=-3pt, forget plot] coordinates {(1,87.94) (2,98.13) (3,96.59) (4,99.09) (5,27.97)};
\addplot[fill=gTmp, draw=gTmp, bar shift=3pt, forget plot] coordinates {(1,88.54) (2,97.04) (3,95.76) (4,98.63) (5,36.92)};
\addplot[fill=gFrm!30, draw=gFrm, bar shift=-3pt, forget plot] coordinates {(6,65.68) (7,59.43) (8,71.93)};
\addplot[fill=gFrm, draw=gFrm, bar shift=3pt, forget plot] coordinates {(6,65.49) (7,60.02) (8,70.96)};
\addplot[fill=gTxt!30, draw=gTxt, bar shift=-3pt, forget plot] coordinates {(9,28.49) (10,24.67) (11,32.31)};
\addplot[fill=gTxt, draw=gTxt, bar shift=3pt, forget plot] coordinates {(9,29.47) (10,25.94) (11,33.01)};
\node[anchor=south, font=\normalsize, inner sep=1.5pt, yshift=3pt] at (rel axis cs:0.5,1.0)
  {\swatch{gTmp}{gTmp!30}\swatch{gFrm}{gFrm!30}\swatch{gTxt}{gTxt!30}\;Self Forcing,
   $16.6$~FPS \qquad
   \swatch{gTmp}{gTmp}\swatch{gFrm}{gFrm}\swatch{gTxt}{gTxt}\;$+$\textsc{UnStep},
   $51.2$~FPS};
\end{axis}
\end{tikzpicture}
\captionsetup{hypcap=false}

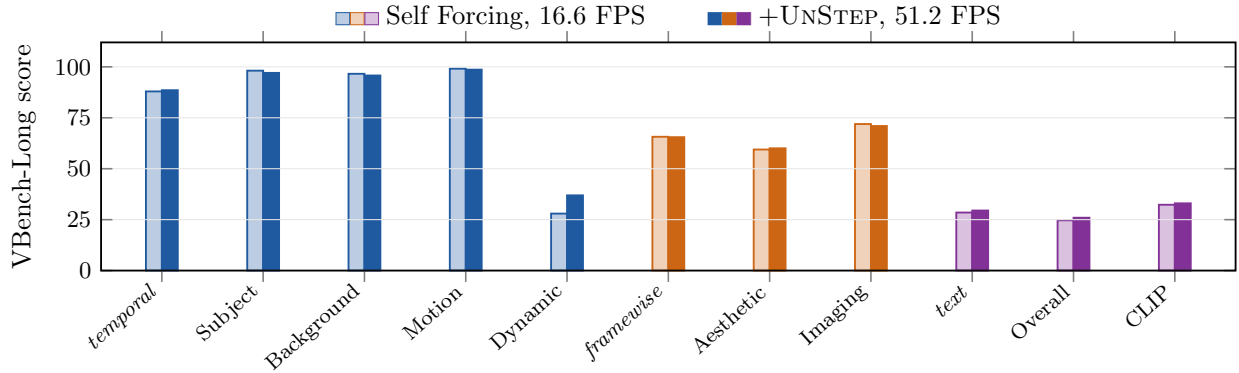
\captionof{figure}{Full VBench-Long results on 40 prompts at 30 seconds per clip on one H100. The first bar in each color group is the aggregate, followed by its component dimensions. }
\label{fig:long-components}
\end{minipage}
\end{center}

%% file: sections/appendix/additional_related_work.tex
\section{Additional Related Work}
\label{app:related}

\paragraph{Broader few-step generation.}
Diffusion Forcing introduced per token noise levels sampled independently for sequence generation \citep{chen2024diffusionforcing}. Closest to \textsc{UnStep} are the causal few-step video models distilled from multi-step bidirectional teachers. The broader
literature includes progressive distillation, consistency models, and distribution matching
distillation \citep{salimans2022progressive,song2023consistency,yin2024dmd,yin2024dmd2}, as well
as adversarial distillation, shortcut models, and continuous time consistency models
\citep{sauer2023add,lin2024sdxllightning,frans2024shortcut,lu2024scm}. Video-specific extensions
include adversarial post-training and one-step or transition matching objectives
\citep{lin2025seaweedapt,aad1,aapt,asd,tmd}. Other work reduces the number of denoising evaluations through few-step consistency or distribution matching objectives \citep{zheng2025rcm,zhao2026causalforcingpp,zheng2026causalrcm,rewardforcing}, reduced steps or one-step training \citep{gu2026anyflow,oneforcing2026}, and noisy causal context with parallel denoising \citep{incontextforcing}. Salt combines self-consistent distribution matching with cache-aware mixed step training to improve low step autoregressive generation \citep{ge2026salt}. DSA learns to allocate steps across frames, while Flex-Forcing trains a unified causal and bidirectional model and evaluates different chunk configurations \citep{dsa,flexforcing}. These methods provide broader context
for trained low step generation, but require optimization of the model rather than changing only
its inference procedure.

\paragraph{Caching and sparse attention for diffusion models.}
Feature reuse in conventional diffusion models includes DeepCache, tokenwise feature caching,
and Pyramid Attention Broadcast \citep{ma2024deepcache,zou2025toca,zhao2025pab}. Feature caching methods such as TeaCache and FasterCache reuse computations across nearby denoising evaluations in video diffusion \citep{liu2025teacache,lv2025fastercache}. Other methods
reduce diffusion transformer attention or token computation through compressed attention,
spatiotemporal sparsity, or adaptive sparsity
\citep{yuan2024ditfastattn,xi2025sparsevideogen,yang2025sparsevideogen2,xia2025adaspa}.
Jenga selects which token pairs interact in attention and progressively increases latent resolution
\citep{zhang2025jenga}. Residual, sparsity, and second-order caching methods provide further acceleration for multi-step
diffusion trajectories \citep{resca,rapid,d2cache}. The caching approaches are less direct comparisons
for a few-step autoregressive model, where there are fewer adjacent denoising evaluations to
reuse.

\paragraph{Efficient video generation and decoding.}
FlashAttention, FlashAttention-3, and \texttt{torch.compile} reduce standard transformer execution overhead \citep{dao2022flashattention,shah2024flashattention3,ansel2024pytorch2}, while video-specific systems optimize positional encoding, attention execution, or low-precision hardware paths \citep{helios,seqparrope,longlive2}.
StreamDiT targets real-time streaming text-to-video generation \citep{streamdit}, while
PyramidalWan finetunes pretrained video diffusion models to denoise at progressively higher resolutions
\citep{pyramidalwan}. Turbo-VAED distills compact video VAE decoders for mobile devices
\citep{turbovaed}. Neodragon combines distillation and denoiser pruning for mobile video generation
\citep{neodragon}, while DeltaQuant quantizes video diffusion models to four bits \citep{deltaquant}.
\textsc{UnStep} instead reduces denoising steps, limits KV attention, and optimizes execution of the pretrained causal DiT and VAE decoder, without retraining or four-bit quantization.

\paragraph{Low rank compression and slimming.}
SVD-LLM uses matrix decomposition for language model compression \citep{wang2025svdllm}.
SVDQuant uses low rank components to absorb outliers during low-bit diffusion model quantization
\citep{li2025svdquant}, while SlimDiff uses activation statistics for training-free
diffusion model slimming \citep{slimdiff}. Increment-calibrated caching uses channel-aware SVD to cheaply update cached outputs based on how the input changes between denoising steps \citep{chen2025increment}. These methods use low rank decomposition or structured slimming to reduce storage
or execution cost. \textsc{UnStep} instead reconstructs the selected V/O matrices at their
original dense shape and uses truncation only as a spectral edit for quality recovery.

%% file: sections/appendix/proofs.tex
\section{Proofs for the Toy Models}
\label{app:theory-proofs}

This appendix gives the proofs and plotting details for the clean-cache and V/O truncation models in \S\ref{sec:theory}.

\paragraph{Shared model and schedule assumptions.}
We assume the linear DiT and matrix conditions in equation~\eqref{eq:theory_clean_linear_model} at every conditioning label, with the text prompt and preceding KV cache fixed. Write the full schedule's noise levels as $(s_0,\ldots,s_n)$, and let $F=\{0,\ldots,n\}$ be its indices. We assume $s_0=1$ and $0<s_j<1$ for $j>0$. Let $J$ denote the indices retained by the shortened schedule. We assume $J\subsetneq F$ and $0\in J$. These conditions include the full and shortened schedules in Appendix~\ref{app:implementation:schedule}.

\paragraph{Shared noise assumptions.}
We assume independent $\bm{\epsilon}_j\sim\mathcal{N}(\mathbf{0},\mathbf{I}_d)$ for $j=0,\ldots,n$. Here $\bm{\epsilon}_0$ is the initial noise, and $\bm{\epsilon}_j$ is the renoising vector at level $s_j$ for $j>0$. We assume that all compared trajectories use the same $\bm{\epsilon}_j$ at each retained level, including initialization. In both proofs, the expectation over $\bm{\epsilon}$ includes initialization and the subsequent noises.

\paragraph{Derivation of the shared denoising update from equations~\eqref{eq:clean_renoise}--\eqref{eq:clean_emit}.}
For a linear DiT $f_{\bm{\theta}}(\mathbf{x},t)=\mathbf{A}\mathbf{x}$ and an incoming latent estimate $\mathbf{x}\in\mathbb{R}^{d}$, equations~\eqref{eq:clean_renoise}--\eqref{eq:clean_emit} give
\begin{equation}
    \mathbf{x}_{\mathrm{next}}
    =(\mathbf{I}_d-s_j\mathbf{A})\left[(1-s_j)\mathbf{x}+s_j\bm{\epsilon}_j\right].
    \label{eq:appendix_clean_sampling_update}
\end{equation}
At $s_0=1$, this gives $(\mathbf{I}_d-\mathbf{A})\bm{\epsilon}_0$, independently of the incoming estimate. The full schedule applies this update at every index in $F$. The shortened schedule applies it only at indices in $J$. At each later step, $\mathbf{x}$ is the output of the preceding step in that schedule.

\paragraph{Shared derivation of $\mathbf{x}_{\mathrm{full}}$ and $\mathbf{x}_{\mathrm{short}}$.}
The matrix assumptions imply that $\mathbf{A}$ has orthonormal eigenvectors $\mathbf{u}_{\ell}$ with eigenvalues $a_{\ell}\in[0,1)$. Write $\epsilon_{j,\ell}=\mathbf{u}_{\ell}^{\top}\bm{\epsilon}_j$. For a scalar eigenvalue $a\in[0,1)$, one step of equation~\eqref{eq:appendix_clean_sampling_update} multiplies the previous coordinate by $q_j(a)$ and adds $v_j(a)\epsilon_{j,\ell}$, where
\begin{equation}
    q_j(a)=(1-s_j)(1-s_j a),
    \qquad
    v_j(a)=s_j(1-s_j a).
\end{equation}
The schedule assumptions and these eigenvalue bounds give $q_0(a)=0$ and $v_0(a)=1-a>0$. For $j>0$, they give $0<q_j(a)<1$ and $v_j(a)>0$. For a schedule with index set $S\in\{F,J\}$, define the coefficient of the noise at index $j$ by
\begin{equation}
    c_{S,j}(a)=
    \begin{cases}
        v_j(a)\displaystyle\prod_{\substack{k\in S\\k>j}}q_k(a), & j\in S,\\
        0, & j\notin S.
    \end{cases}
    \label{eq:appendix_shared_noise_coefficients}
\end{equation}
An empty product equals one. Recursively applying equation~\eqref{eq:appendix_clean_sampling_update} gives
\begin{equation}
\begin{aligned}
    \mathbf{u}_{\ell}^{\top}\mathbf{x}_{\mathrm{full}}
    &=\sum_{j=0}^{n}c_{F,j}(a_{\ell})\epsilon_{j,\ell},\\
    \mathbf{u}_{\ell}^{\top}\mathbf{x}_{\mathrm{short}}
    &=\sum_{j=0}^{n}c_{J,j}(a_{\ell})\epsilon_{j,\ell}.
\end{aligned}
    \label{eq:appendix_shared_schedule_estimates}
\end{equation}

For $j\in J$, the full schedule's coefficient contains the additional factors from the omitted steps. Thus
\begin{equation}
    c_{F,j}(a)=c_{J,j}(a)P_j(a),
    \qquad
    P_j(a)=\prod_{\substack{k\in F\setminus J\\k>j}}q_k(a).
    \label{eq:appendix_shared_coefficient_relation}
\end{equation}

\subsection{Clean-cache refinement}
\label{app:proof-clean}

\paragraph{Derivation of $\mathbf{x}_{\mathrm{clean}}(\sigma)$ in equation~\eqref{eq:theory_clean_risk}.}
For the schedule estimates in equation~\eqref{eq:appendix_shared_schedule_estimates}, write
\begin{equation}
    \mathbf{e}=\mathbf{x}_{\mathrm{short}}-\mathbf{x}_{\mathrm{full}}.
    \label{eq:theory_clean_residual}
\end{equation}
For the clean-cache step, we assume an additional standard Gaussian vector $\bm{\epsilon}_i$, with $i=n+1$, independent of the denoising noises. We fix the DiT conditioning label at $t_i=0$, treat the noise level as a variable $\sigma\in[0,1]$, and write the updates in equations~\eqref{eq:clean_renoise}--\eqref{eq:clean_emit} as
\begin{equation}
    \tilde{\mathbf{x}}_{\mathrm{short}}(\sigma)=(1-\sigma)\mathbf{x}_{\mathrm{short}}+\sigma\bm{\epsilon}_i,
    \qquad
    \mathbf{x}_{\mathrm{clean}}(\sigma)=\tilde{\mathbf{x}}_{\mathrm{short}}(\sigma)-\sigma f_{\bm{\theta}}(\tilde{\mathbf{x}}_{\mathrm{short}}(\sigma),0),
    \label{eq:theory_clean_update}
\end{equation}
\paragraph{Proof of the risk expansion in equation~\eqref{eq:theory_clean_local_risk}.}
Define $\mathbf{M}(\sigma)=(1-\sigma)(\mathbf{I}_d-\sigma\mathbf{A})$. Substituting the linear predictor into equation~\eqref{eq:theory_clean_update} gives
\begin{equation}
    \mathbf{r}(\sigma)
    =\mathbf{M}(\sigma)\mathbf{x}_{\mathrm{short}}-\mathbf{x}_{\mathrm{full}}
      +\sigma(\mathbf{I}_d-\sigma\mathbf{A})\bm{\epsilon}_i.
\end{equation}
The independence and standard Gaussian law of $\bm{\epsilon}_i$ imply that, conditional on the two schedule outputs, the last term has zero mean and expected squared norm $\sigma^2\|\mathbf{I}_d-\sigma\mathbf{A}\|_F^2$, where $\|\cdot\|_F$ is the Frobenius norm. Hence
\begin{equation}
    R_{\mathrm{clean}}(\sigma)
    =\mathbb{E}\left\|
      \mathbf{M}(\sigma)\mathbf{x}_{\mathrm{short}}-\mathbf{x}_{\mathrm{full}}
      \right\|_2^2
      +\sigma^2\|\mathbf{I}_d-\sigma\mathbf{A}\|_F^2.
    \label{eq:theory_clean_linear_risk}
\end{equation}
Since $\mathbf{M}(\sigma)=\mathbf{I}_d-\sigma(\mathbf{I}_d+\mathbf{A})+\sigma^2\mathbf{A}$, expanding equation~\eqref{eq:theory_clean_linear_risk} gives
\begin{equation}
    R_{\mathrm{clean}}(\sigma)
    =\mathbb{E}\|\mathbf{e}\|_2^2
      -2g\sigma+\beta_2\sigma^2+\beta_3\sigma^3+\beta_4\sigma^4,
    \label{eq:appendix_clean_risk_polynomial}
\end{equation}
where
\begin{equation}
\begin{aligned}
    \beta_2
    &=\mathbb{E}\|(\mathbf{I}_d+\mathbf{A})\mathbf{x}_{\mathrm{short}}\|_2^2
      +2\mathbb{E}[\mathbf{e}^{\top}\mathbf{A}\mathbf{x}_{\mathrm{short}}]+d,\\
    \beta_3
    &=-2\mathbb{E}\!\left[
      ((\mathbf{I}_d+\mathbf{A})\mathbf{x}_{\mathrm{short}})^{\top}\mathbf{A}\mathbf{x}_{\mathrm{short}}
      \right]-2\operatorname{tr}(\mathbf{A}),\\
    \beta_4
    &=\mathbb{E}\|\mathbf{A}\mathbf{x}_{\mathrm{short}}\|_2^2+\|\mathbf{A}\|_F^2.
\end{aligned}
\end{equation}
Thus equation~\eqref{eq:theory_clean_local_risk} holds with
\begin{equation}
    \rho(\sigma)=\beta_2\sigma^2+\beta_3\sigma^3+\beta_4\sigma^4,
    \qquad
    k=|\beta_2|+|\beta_3|+\beta_4.
    \label{eq:appendix_clean_remainder_constant}
\end{equation}
All moments are finite because the outputs are finite linear combinations of Gaussian noise. If $\mathbf{A}\neq\mathbf{0}$, then $\beta_4>0$. If $\mathbf{A}=\mathbf{0}$, then $\beta_2=\mathbb{E}\|\mathbf{x}_{\mathrm{short}}\|_2^2+d>0$. In either case, $k>0$, and $|\rho(\sigma)|\leq k\sigma^2$ for $\sigma\in[0,1]$.

\paragraph{Proof that $g>0$ for equation~\eqref{eq:theory_clean_positive_minimizer}.}
The shared noise assumptions and orthonormality of the eigenvectors imply that the $\epsilon_{j,\ell}$ have zero mean and unit variance and are independent across levels. Substituting equation~\eqref{eq:appendix_shared_schedule_estimates} into $g$ and taking the expectation eliminates cross terms between distinct noise indices. Since $c_{J,j}=0$ for $j\notin J$, equation~\eqref{eq:appendix_shared_coefficient_relation} gives
\begin{equation}
\begin{aligned}
    g&=\mathbb{E}\!\left[\mathbf{e}^{\top}
      (\mathbf{I}_d+\mathbf{A})\mathbf{x}_{\mathrm{short}}\right]\\
    &=\sum_{\ell=1}^{d}(1+a_{\ell})
      \sum_{j\in J}c_{J,j}(a_{\ell})^2[1-P_j(a_{\ell})]>0.
\end{aligned}
    \label{eq:appendix_clean_shortened_schedule}
\end{equation}
Every summand is nonnegative. For every $\ell$, the initial noise term is strictly positive because $c_{J,0}(a_{\ell})>0$ and $P_0(a_{\ell})<1$, as the full schedule contains at least one additional level.

\paragraph{Proof of the unique full schedule minimizer.}
For the full schedule, replace $\mathbf{x}_{\mathrm{short}}$ by $\mathbf{x}_{\mathrm{full}}$ in equation~\eqref{eq:theory_clean_linear_risk}. The risk is zero at $\sigma=0$. For $\sigma\in(0,1]$, the noise term $\sigma^2\|\mathbf{I}_d-\sigma\mathbf{A}\|_F^2$ is strictly positive because every eigenvalue of $\mathbf{I}_d-\sigma\mathbf{A}$ is positive. The other term is nonnegative, so $\sigma_{\mathrm{full}}^\star=0$ is the unique minimizer.

\paragraph{Proof of the minimizer bound in equation~\eqref{eq:theory_clean_positive_minimizer}.}
The risk is a polynomial, so it attains a minimum on $[0,1]$. For the shortened schedule, equation~\eqref{eq:appendix_clean_shortened_schedule} gives $g>0$. Set $\delta=\min\{1,g/k\}$. Because $k \delta\leq g$, the remainder bound gives
\begin{equation}
    R_{\mathrm{clean}}(\delta)
    \leq R_{\mathrm{clean}}(0)-2g \delta+k \delta^2
    \leq R_{\mathrm{clean}}(0)-g \delta.
\end{equation}
For every $0\leq\sigma\leq \delta/3$, the lower remainder bound $\rho(\sigma)\geq-k\sigma^2$ gives
\begin{equation}
\begin{aligned}
    R_{\mathrm{clean}}(\sigma)
    &\geq R_{\mathrm{clean}}(0)-2g\sigma-k\sigma^2\\
    &\geq R_{\mathrm{clean}}(0)-\frac{7}{9}g \delta
    >R_{\mathrm{clean}}(\delta).
\end{aligned}
\end{equation}
No minimizer lies in $[0,\delta/3]$. This proves equation~\eqref{eq:theory_clean_positive_minimizer} and shows that every shortened schedule minimizer has strictly lower risk than $\sigma=0$.

\paragraph{Derivation of the toy curve in Figure~\ref{fig:theory_clean_toy}.}
For the plotted instance, choose $\mathbf{A}=a\mathbf{I}_d$, a full schedule $(1,s_1,s_2,s_3)$, and a shortened schedule $(1,s_1)$. We evaluate the risk conditional on the initial vector $\bm{\epsilon}_0=\mathbf{z}$. Both schedules use equation~\eqref{eq:appendix_clean_sampling_update}. The full schedule uses independent standard Gaussian vectors $\bm{\epsilon}_1,\bm{\epsilon}_2,\bm{\epsilon}_3$ after initialization. For this plot, we instead assume that both schedules share $\bm{\epsilon}_3$ at their final denoising step. The shortened schedule's clean-cache pass uses fresh standard Gaussian noise independent of the denoising noises. Appendix~\ref{app:theory-plotting} gives the numerical parameters.

For this conditional risk, we compute $g$ from the two schedule outputs. Write $q(s)=(1-s)(1-as)$ and $b(s)=s(1-as)$ for the scalar coefficients of equation~\eqref{eq:appendix_clean_sampling_update}, and define
\begin{equation}
    c_{\mathrm{short}}=(1-a)q(s_1),
    \qquad
    c_{\mathrm{full}}=(1-a)q(s_1)q(s_2)q(s_3).
\end{equation}
Recursively applying equation~\eqref{eq:appendix_clean_sampling_update} at each denoising step gives
\begin{equation}
\begin{aligned}
    \mathbf{x}_{\mathrm{short}}&=c_{\mathrm{short}}\mathbf{z}+b(s_1)\bm{\epsilon}_3,\\
    \mathbf{x}_{\mathrm{full}}&=c_{\mathrm{full}}\mathbf{z}
      +b(s_1)q(s_2)q(s_3)\bm{\epsilon}_1
      +b(s_2)q(s_3)\bm{\epsilon}_2+b(s_3)\bm{\epsilon}_3.
\end{aligned}
\end{equation}
Independence of $\bm{\epsilon}_1,\bm{\epsilon}_2,\bm{\epsilon}_3$ and their standard Gaussian moments give
\begin{equation}
    \frac{g}{d}=(1+a)\left[
      c_{\mathrm{short}}(c_{\mathrm{short}}-c_{\mathrm{full}})\frac{\|\mathbf{z}\|_2^2}{d}
      +b(s_1)\bigl(b(s_1)-b(s_3)\bigr)\right].
\end{equation}
The parameters in Appendix~\ref{app:theory-plotting} give $c_{\mathrm{short}}>c_{\mathrm{full}}>0$ and $b(s_1)>b(s_3)>0$, so $g>0$. The same risk expansion and minimizer bound argument apply to this conditional risk.

To obtain the plotted function, set $m(\sigma)=(1-\sigma)(1-a\sigma)$. Substituting the two outputs into equation~\eqref{eq:theory_clean_linear_risk} and taking the expectation gives
\begin{equation}
\begin{aligned}
    \overline{R}_{\mathrm{clean}}(\sigma)
    &=\frac{1}{d}\mathbb{E}\!\left[\|\mathbf{r}(\sigma)\|_2^2\mid\bm{\epsilon}_0=\mathbf{z}\right]\\
    &=\frac{\|\mathbf{z}\|_2^2}{d}
      \bigl(m(\sigma)c_{\mathrm{short}}-c_{\mathrm{full}}\bigr)^2
      +\bigl(m(\sigma)b(s_1)-b(s_3)\bigr)^2\\
    &\quad+\bigl(b(s_1)q(s_2)q(s_3)\bigr)^2
      +\bigl(b(s_2)q(s_3)\bigr)^2
      +\sigma^2(1-a\sigma)^2.
\end{aligned}
    \label{eq:appendix_clean_scalar_curve}
\end{equation}

\subsection{V/O truncated SVD}
\label{app:proof-svd}

\paragraph{Additional matrix assumptions for equations~\eqref{eq:theory_svd_risk}--\eqref{eq:theory_svd_risk_expansion}.}
We additionally assume $0<\operatorname{rank}(\mathbf{A})<d$. Define $q=\operatorname{rank}(\mathbf{A})$, and let $a_{\min}^{+}$ be the smallest positive eigenvalue of $\mathbf{A}$. We assume $0<b<\min\{a_{\min}^{+}/2,b_{\mathrm{schedule}}\}$. The existence of $b_{\mathrm{schedule}}>0$ is proved below when deriving $\kappa>0$. We use the reference $\mathbf{x}_{\mathrm{full}}=\mathbf{x}_{\mathrm{full}}(d)$ and set the clean-cache noise level to zero in this calculation.

\paragraph{Derivation of the truncated model in equation~\eqref{eq:theory_svd_model}.}
Set $\mathbf{W}=\mathbf{A}-b\mathbf{I}_d$ and $\mathbf{A}_r=b\mathbf{I}_d+\widehat{\mathbf{W}}_r$ as in equation~\eqref{eq:theory_svd_model}. Since $\mathbf{A}$ is positive semidefinite and has rank $q$, it has $q$ positive eigenvalues $a_{\ell}$ and $d-q$ zero eigenvalues. The assumed bound $b<a_{\min}^{+}/2$ then gives $q$ eigenvalues $w_{\ell}=a_{\ell}-b>b$ of $\mathbf{W}$ and $d-q$ eigenvalues equal to $-b$. Its singular values are $|w_{\ell}|$, so the positive eigenvalues come first when ordered by nonincreasing magnitude. Using the same ordered orthonormal eigenvectors $\mathbf{u}_{\ell}$ for every rank, the truncated SVD gives
\begin{equation}
    \mathbf{W}=\sum_{\ell=1}^{d}w_{\ell}\mathbf{u}_{\ell}\mathbf{u}_{\ell}^{\top},
    \qquad
    \widehat{\mathbf{W}}_r=\sum_{\ell=1}^{r}w_{\ell}\mathbf{u}_{\ell}\mathbf{u}_{\ell}^{\top}.
    \label{eq:appendix_svd_weights}
\end{equation}
In this eigenbasis, $\mathbf{A}_r$ has eigenvalue $a_{\ell}=b+w_{\ell}$ for $\ell\leq r$ and $b$ otherwise. Since $0\leq a_{\ell}<1$ and $0<b<1$, this proves $\mathbf{0}\preceq\mathbf{A}_r\prec\mathbf{I}_d$ for every rank.

\paragraph{Applying the shared derivation to $\mathbf{A}_r$.}
The truncated matrix has eigenvalue $a_{r,\ell}=a_{\ell}$ for $\ell\leq r$ and $a_{r,\ell}=b$ otherwise. Replacing $a_{\ell}$ by $a_{r,\ell}$ in equation~\eqref{eq:appendix_shared_schedule_estimates} gives $\mathbf{x}_{\mathrm{full}}(r)$ and $\mathbf{x}_{\mathrm{short}}(r)$. The reference $\mathbf{x}_{\mathrm{full}}$ retains the original eigenvalues $a_{\ell}$.
\paragraph{Derivation of the risk in equation~\eqref{eq:theory_svd_risk}.}
The shared noise assumptions imply that the noise coordinates have zero mean and unit variance and are independent across indices. Squaring the difference from the fixed reference and taking its expectation gives
\begin{equation}
    R_{\mathrm{svd}}(r)
    =\sum_{\ell=1}^{d}\sum_{j=0}^{n}
      \left[c_{J,j}(a_{r,\ell})-c_{F,j}(a_{\ell})\right]^2.
    \label{eq:appendix_svd_exact_risk}
\end{equation}

\paragraph{Proof of the risk expansion in equation~\eqref{eq:theory_svd_risk_expansion}.}
In direction $\mathbf{u}_{\ell}$, retaining the corresponding component of $\mathbf{W}$ leaves the model eigenvalue at $a_{\ell}$, while discarding it replaces that eigenvalue by $b$. Denote the expected squared errors for these two choices by $e_{\ell}^{\mathrm{keep}}$ and $e_{\ell}^{\mathrm{drop}}$. Equation~\eqref{eq:appendix_svd_exact_risk} gives
\begin{equation}
\begin{aligned}
    e_{\ell}^{\mathrm{keep}}
    &=\sum_{j=0}^{n}[c_{J,j}(a_{\ell})-c_{F,j}(a_{\ell})]^2,\\
    e_{\ell}^{\mathrm{drop}}
    &=\sum_{j=0}^{n}[c_{J,j}(b)-c_{F,j}(a_{\ell})]^2.
\end{aligned}
    \label{eq:appendix_svd_direction_errors}
\end{equation}
The total risk is therefore
\begin{equation}
    R_{\mathrm{svd}}(r)
    =\sum_{\ell=1}^{r}e_{\ell}^{\mathrm{keep}}
      +\sum_{\ell=r+1}^{d}e_{\ell}^{\mathrm{drop}}.
    \label{eq:appendix_svd_direction_sum}
\end{equation}
For $\ell>q$, the original eigenvalue is zero and the noise law is the same in every direction. Both $e_{\ell}^{\mathrm{keep}}$ and $e_{\ell}^{\mathrm{drop}}$ are therefore constant across these directions. Define
\begin{equation}
\begin{aligned}
    \eta_{\ell}
    &=e_{\ell}^{\mathrm{drop}}-e_{\ell}^{\mathrm{keep}}, && 1\leq\ell\leq q,\\
    \kappa
    &=e_{\ell}^{\mathrm{keep}}-e_{\ell}^{\mathrm{drop}}, && \ell>q.
\end{aligned}
    \label{eq:appendix_svd_error_increases}
\end{equation}
At rank $q$, the first $q$ directions are retained and the rest are discarded. For $r<q$, discarding the directions $r<\ell\leq q$ adds $\sum_{r<\ell\leq q}\eta_{\ell}$ to this risk. For $r\geq q$, retaining the additional $r-q$ directions adds $(r-q)\kappa$. This proves equation~\eqref{eq:theory_svd_risk_expansion}.

\paragraph{Proof that $\eta_{\ell}>0$ and $\kappa>0$ in equation~\eqref{eq:theory_svd_risk_expansion}.}
For $j\in J$, write $C_j(a)=c_{J,j}(a)$, so equation~\eqref{eq:appendix_shared_coefficient_relation} gives $c_{F,j}(a)=C_j(a)P_j(a)$. For $a\in[0,1)$, $C_j(a)>0$, $0<P_j(a)\leq1$, and $C_j(a)$ is strictly decreasing because each factor $1-s_k a$ is positive and strictly decreasing.

For $\ell\leq q$, $a_{\ell}>b$ gives
\begin{equation}
    0\leq C_j(a_{\ell})[1-P_j(a_{\ell})]
    <C_j(b)-C_j(a_{\ell})P_j(a_{\ell}).
\end{equation}
Squaring shows that each term with $j\in J$ in $e_{\ell}^{\mathrm{keep}}$ is smaller than its counterpart in $e_{\ell}^{\mathrm{drop}}$. Terms with $j\notin J$ are equal because $c_{J,j}=0$ and the reference is unchanged. Thus $e_{\ell}^{\mathrm{keep}}<e_{\ell}^{\mathrm{drop}}$, proving $\eta_{\ell}>0$.

For $\ell>q$, $a_{\ell}=0$. Writing $\kappa$ as a function of $b$ and substituting equation~\eqref{eq:appendix_svd_direction_errors} into equation~\eqref{eq:appendix_svd_error_increases} gives
\begin{equation}
\begin{aligned}
    \kappa(b)=\sum_{j\in J}\bigl\{
    C_j(0)^2[1-P_j(0)]^2
    -[C_j(b)-C_j(0)P_j(0)]^2\bigr\}.
\end{aligned}
    \label{eq:appendix_svd_tail_increase}
\end{equation}
We have $\kappa(0)=0$, and differentiating gives
\begin{equation}
    \kappa'(0)=-2\sum_{j\in J}C_j'(0)C_j(0)[1-P_j(0)]>0.
    \label{eq:appendix_svd_tail_positive}
\end{equation}
Indeed, $C_j'(0)=-C_j(0)\sum_{k\in J,\,k\geq j}s_k<0$. Every contribution to $\kappa'(0)$ is nonnegative, and the term at $j=0$ is positive because the shortened schedule omits at least one later step, giving $P_0(0)<1$. Since $\kappa(b)$ is a polynomial, it is positive throughout some interval $0<b<b_{\mathrm{schedule}}$. Choosing $b$ in this interval and below half the smallest positive eigenvalue of $\mathbf{A}$ ensures both the required singular value ordering and $\kappa>0$.

\paragraph{Proof of the unique shortened schedule minimizer.}
Equation~\eqref{eq:theory_svd_risk_expansion} and the strict positivity of $\eta_{\ell}$ and $\kappa$ give the unique minimizer $r_{\mathrm{short}}^\star=q<d$.

\paragraph{Proof of the unique full schedule minimizer.}
Using the full schedule coefficients gives
\begin{equation}
    \mathbb{E}\|\mathbf{x}_{\mathrm{full}}(r)-\mathbf{x}_{\mathrm{full}}\|_2^2
    =\sum_{\ell=r+1}^{d}\sum_{j=0}^{n}
      [c_{F,j}(b)-c_{F,j}(b+w_{\ell})]^2.
\end{equation}
For each removed direction, the final noise coefficient differs by $v_n(b)-v_n(b+w_{\ell})=s_n^2w_{\ell}\neq0$. Hence every $r<d$ gives positive risk, while $r=d$ gives zero risk. Thus the unique full schedule minimizer is $d$.

\paragraph{Verification of the toy curve in Figure~\ref{fig:theory_svd_toy}.}
Use the full schedule $(1,s_1,s_2,s_3)$ and shortened schedule $(1,s_3)$, with the shared indexed standard Gaussian noises defined above. For integers $0<k<q<d$ and $b<h<1-b$, define the fixed matrix $\mathbf{W}=\operatorname{diag}(w_1,\ldots,w_d)$ by
\begin{equation}
    w_{\ell}=\begin{cases}
        h, & 1\leq\ell\leq k,\\
        b+(h-b)\left(\dfrac{q-\ell+1/2}{q-k}\right)^{1/3}, & k<\ell\leq q,\\
        -b, & q<\ell\leq d.
    \end{cases}
    \label{eq:appendix_svd_numeric_weights}
\end{equation}
The magnitudes are nonincreasing, and every positive eigenvalue exceeds the magnitude $b$ of the negative eigenvalues. We retain the first $r$ coordinate directions in the SVD of $\mathbf{W}$, including when equal singular values occur. All eigenvalues of $\mathbf{A}_r$ lie in $[0,b+h]\subset[0,1)$ for every $r$.

This spectrum gives $\operatorname{rank}(\mathbf{A})=q$. The positive eigenvalues of $\mathbf{W}$ give $\eta_{\ell}>0$ for $\ell\leq q$, and Appendix~\ref{app:theory-plotting} verifies $\kappa>0$ for the plotted parameters.

\subsection{Plotting parameters}
\label{app:theory-plotting}

\paragraph{Clean-cache curve.}
Figure~\ref{fig:theory_clean_toy} plots equation~\eqref{eq:appendix_clean_scalar_curve} using $a=0.1$, $\|\mathbf{z}\|_2^2/d=7.57$, full denoising levels $(1,0.131,0.015,0.0033)$, and shortened levels $(1,0.131)$. The plotted interval is $\sigma\in[0,0.05]$.

\paragraph{SVD curve.}
Figure~\ref{fig:theory_svd_toy} plots $\overline{R}_{\mathrm{svd}}(r)=R_{\mathrm{svd}}(r)/d$ from equation~\eqref{eq:appendix_svd_exact_risk}, with the weights in equation~\eqref{eq:appendix_svd_numeric_weights}. We use $d=500$, $b=0.05$, $h=0.534$, $k=430$, and $q=460$. The full denoising levels are $(1,15/16,5/6,5/8)$ and the shortened levels are $(1,5/8)$, one of the two-step schedules in Appendix~\ref{app:implementation:schedule}. Substituting these parameters into equation~\eqref{eq:appendix_svd_tail_increase} gives $\kappa\approx0.0209>0$. We plot every integer rank from $k$ to $d$ against $p=r/d$.

\paragraph{DiT curves.}
For each sweep value, we average each schedule's latent tensors over five paired runs, giving vectors $\overline{\mathbf{x}}_{\mathrm{full}}$ and $\overline{\mathbf{x}}_{\mathrm{short}}$ with $d_{\mathrm{lat}}$ coordinates. The plotted MSE is
\begin{equation}
    \frac{\|\overline{\mathbf{x}}_{\mathrm{short}}-\overline{\mathbf{x}}_{\mathrm{full}}\|_2^2}{d_{\mathrm{lat}}}.
\end{equation}
In both sweeps, the full schedule reference uses $\sigma=0$ and full V/O retention ($p=1$). For the shortened schedule, we sweep $\sigma$ with $p=1$, then sweep $p$ with $\sigma=0.02$, following the clean-cache then SVD order in Table~\ref{tab:methods}. Both use the denoising noise levels specified in Appendix~\ref{app:implementation:schedule} and are shown over the same horizontal ranges as their corresponding toy curves.

%% file: sections/appendix/visualizations.tex
\section{Visualizations}
\label{app:visualizations}

For the text-to-video examples below, we compare Self Forcing and Causal Forcing with the same checkpoints wrapped by the default two-step \textsc{UnStep} configuration. Every comparison uses the same VBench entry and seed across all four rows. We show the original models at 17.0 FPS and their \textsc{UnStep} variants at 49.8 FPS. Each row contains frames 0, 20, 40, 60, and 80 from an 81-frame clip, corresponding to 0, 1.25, 2.5, 3.75, and 5 seconds. Every clip is generated from the extended prompt. The first example shows its full extended prompt, while the remaining examples show only the original VBench prompts.

\newlength{\visframewidth}
\newlength{\visframeheight}
\newlength{\visstripwidth}
\newlength{\vislabelwidth}
\setlength{\visframewidth}{0.173\linewidth}
\setlength{\visframeheight}{0.09981\linewidth}
\setlength{\visstripwidth}{0.873\linewidth}
\setlength{\vislabelwidth}{0.12\linewidth}

\newcommand{\vistimes}{%
  \noindent\makebox[\linewidth][c]{%
    \hspace*{\vislabelwidth}\hspace{0.004\linewidth}%
    \makebox[\visframewidth][c]{\scriptsize 0 s}\hspace{0.002\linewidth}%
    \makebox[\visframewidth][c]{\scriptsize 1.25 s}\hspace{0.002\linewidth}%
    \makebox[\visframewidth][c]{\scriptsize 2.5 s}\hspace{0.002\linewidth}%
    \makebox[\visframewidth][c]{\scriptsize 3.75 s}\hspace{0.002\linewidth}%
    \makebox[\visframewidth][c]{\scriptsize 5 s}%
  }\par\vspace{0.4pt}%
}

\newcommand{\visrow}[2]{%
  \setlength{\lineskip}{0pt}%
  \noindent\makebox[\linewidth][c]{%
    \begin{minipage}[b][\visframeheight][c]{\vislabelwidth}\raggedleft\scriptsize #1\end{minipage}\hspace{0.004\linewidth}%
    \def\visstripfile{#2}\includegraphics[width=\visstripwidth]\visstripfile%
  }\par\vspace{0.4pt}%
}

\newcommand{\visprompt}[1]{%
  \vspace{2pt}%
  {\small\textit{#1}}\par%
}

\newcommand{\sfvislabel}{\shortstack[r]{Self Forcing\\17.0 FPS}}
\newcommand{\sfunstepvislabel}{\shortstack[r]{SF $+$ \textsc{UnStep}\\49.8 FPS}}
\newcommand{\cfvislabel}{\shortstack[r]{Causal Forcing\\17.0 FPS}}
\newcommand{\cfunstepvislabel}{\shortstack[r]{CF $+$ \textsc{UnStep}\\49.8 FPS}}

\newcommand{\vispage}[3]{%
  \begin{tikzpicture}
    \path[use as bounding box] (0,0) rectangle (\linewidth,\textheight);
    \node[anchor=north west,inner sep=0pt,outer sep=0pt] at (0,\textheight) {#1};
    \node[anchor=center,inner sep=0pt,outer sep=0pt] at ($(0,0)!0.5!(\linewidth,\textheight)$) {#2};
    \node[anchor=south west,inner sep=0pt,outer sep=0pt] at (0,0) {#3};
  \end{tikzpicture}%
}

\par\medskip
\noindent
\begin{minipage}{\linewidth}
\centering
\vistimes
\visrow{\sfvislabel}{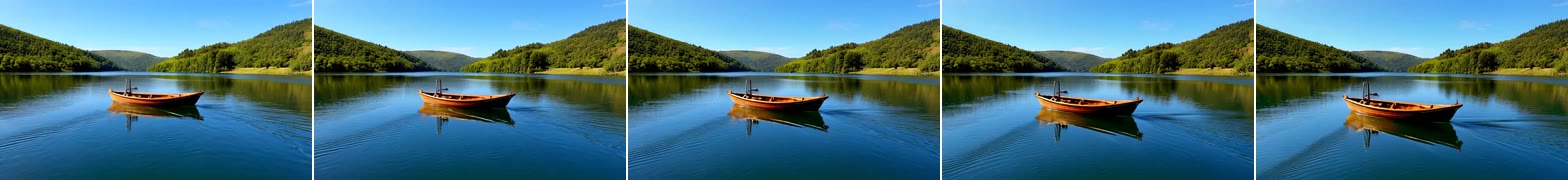}
\visrow{\sfunstepvislabel}{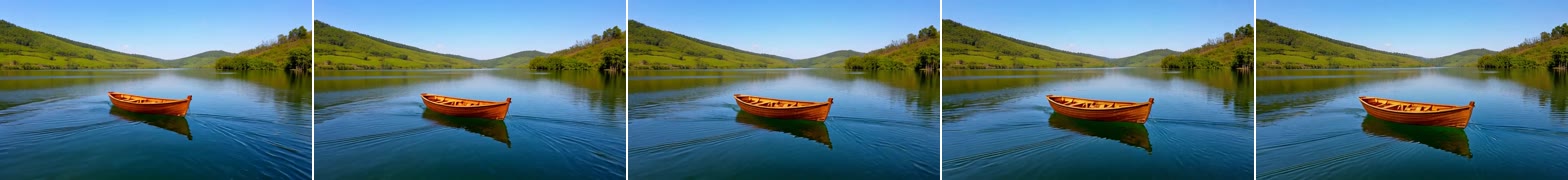}
\visrow{\cfvislabel}{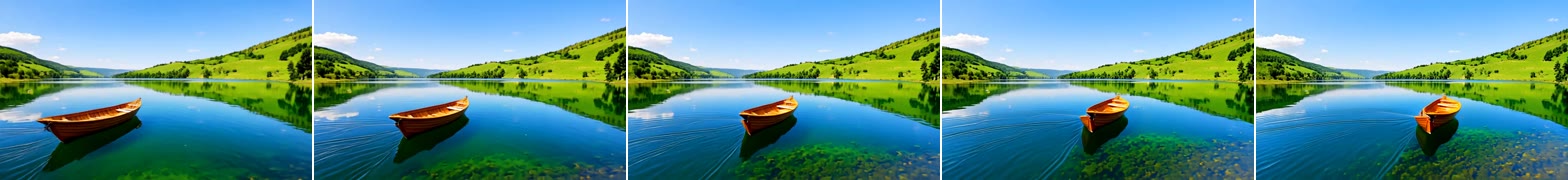}
\visrow{\cfunstepvislabel}{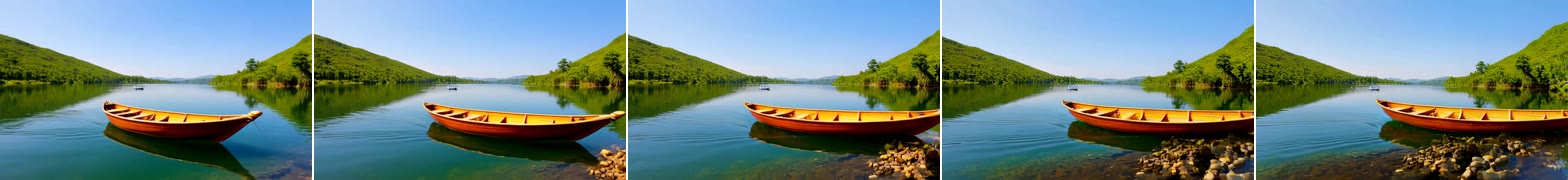}
\visprompt{A serene, picturesque scene of a small wooden boat gently sailing across a calm, tranquil lake under a clear blue sky. The boat moves smoothly with the gentle breeze, creating only minimal ripples on the water's surface. The background showcases lush green hills and a few scattered trees reflecting in the crystal-clear water. The camera focuses on the boat from a medium distance, capturing the peaceful atmosphere of the lake. The scene remains static, emphasizing the calmness and tranquility of the environment.}
\end{minipage}

\begin{figure}[p]
\centering
\vispage{
\begin{minipage}{\linewidth}
\centering
\vistimes
\visrow{\sfvislabel}{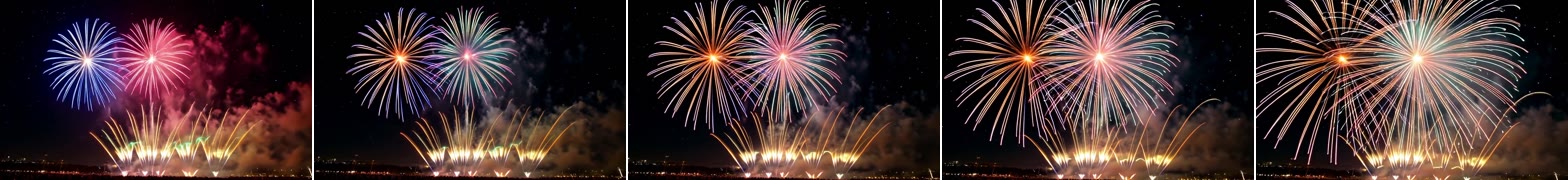}
\visrow{\sfunstepvislabel}{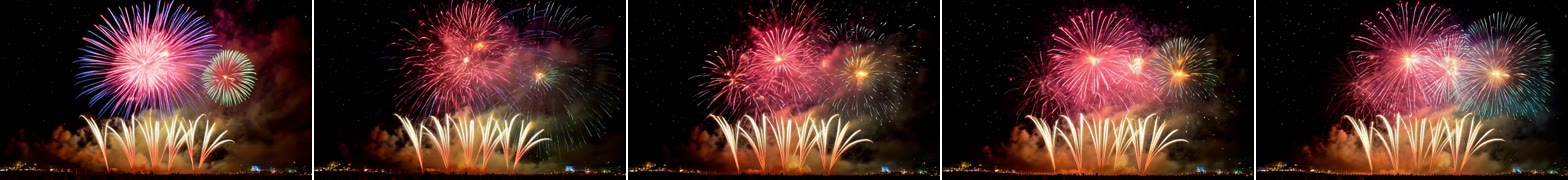}
\visrow{\cfvislabel}{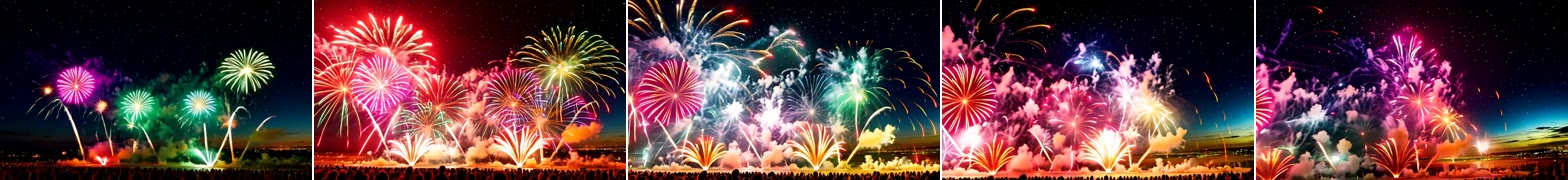}
\visrow{\cfunstepvislabel}{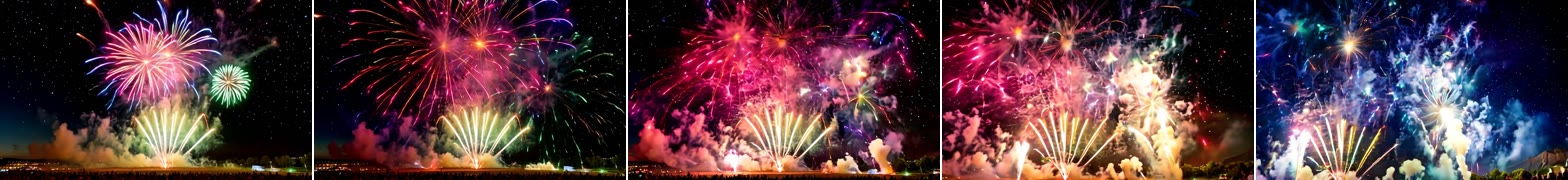}
\visprompt{Fireworks.}
\end{minipage}
}{
\begin{minipage}{\linewidth}
\centering
\visrow{\sfvislabel}{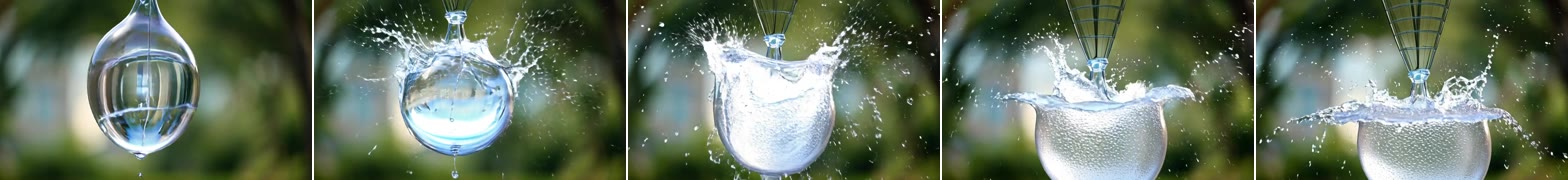}
\visrow{\sfunstepvislabel}{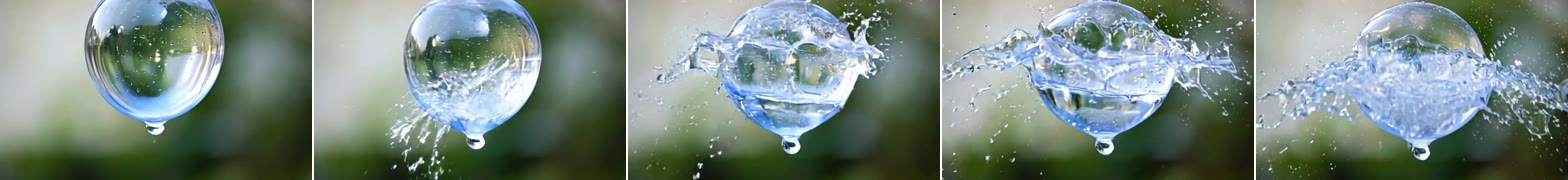}
\visrow{\cfvislabel}{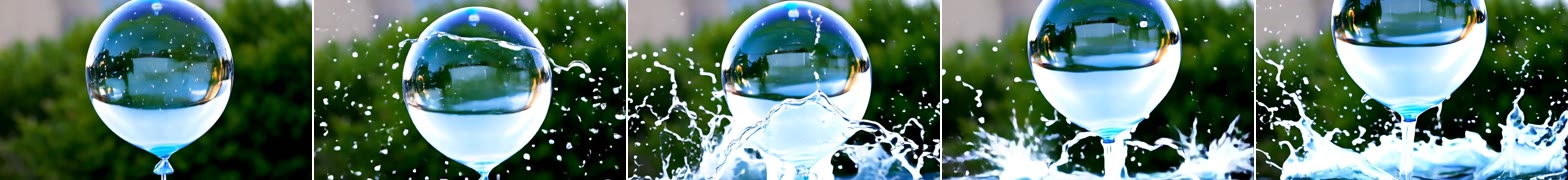}
\visrow{\cfunstepvislabel}{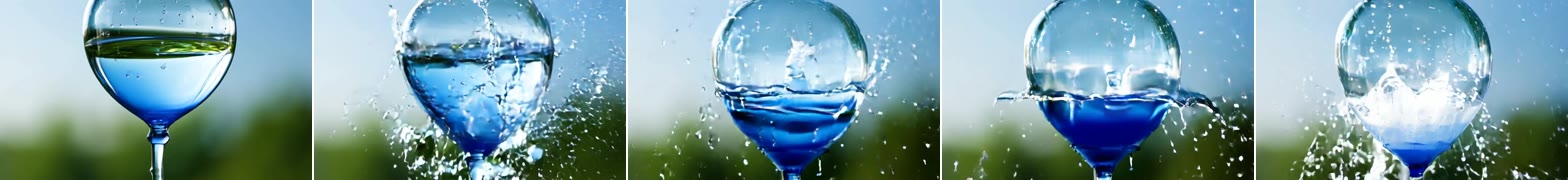}
\visprompt{Balloon full of water exploding in extreme slow motion.}
\end{minipage}
}{
\begin{minipage}{\linewidth}
\centering
\visrow{\sfvislabel}{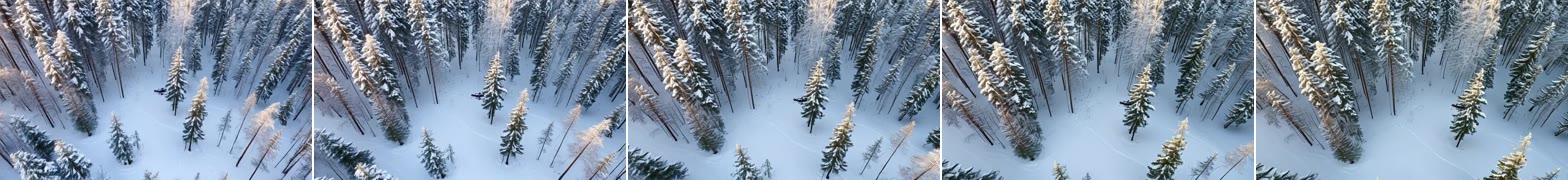}
\visrow{\sfunstepvislabel}{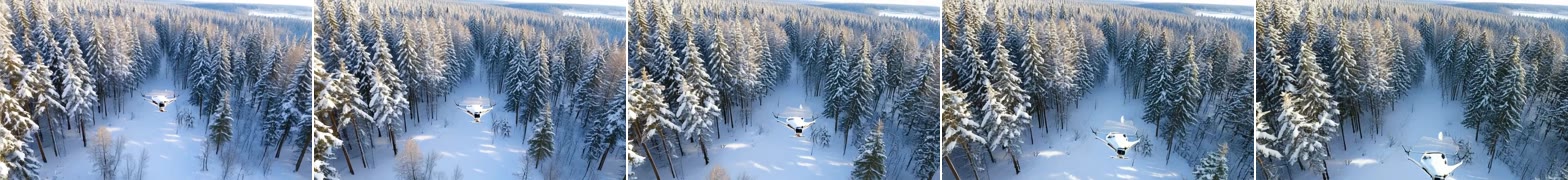}
\visrow{\cfvislabel}{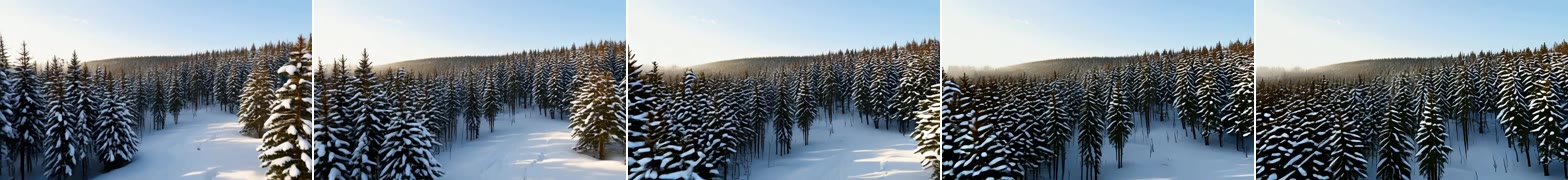}
\visrow{\cfunstepvislabel}{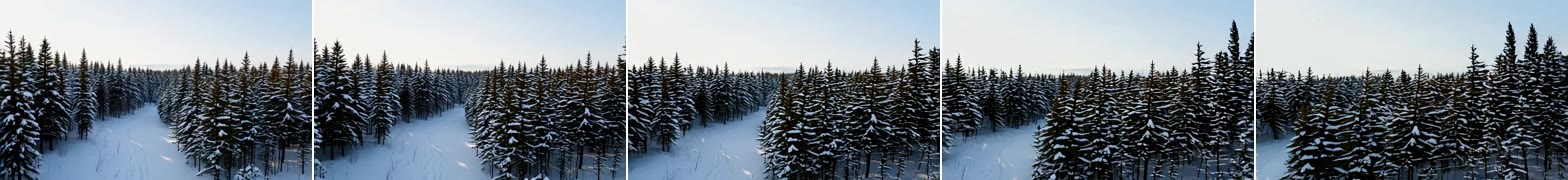}
\visprompt{a drone flying over a snowy forest.}
\end{minipage}
}
\end{figure}

\begin{figure}[p]
\centering
\vispage{
\begin{minipage}{\linewidth}
\centering
\vistimes
\visrow{\sfvislabel}{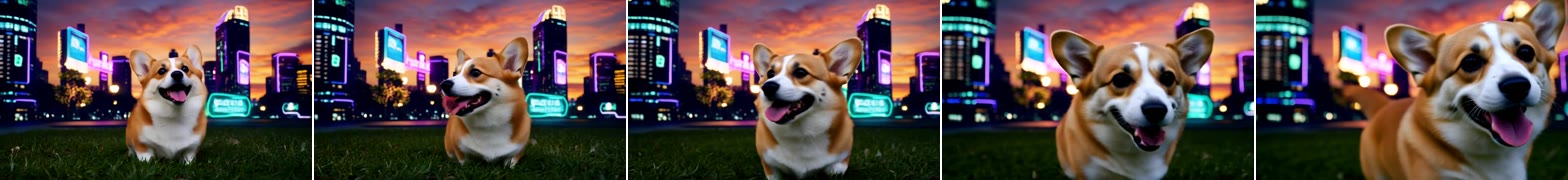}
\visrow{\sfunstepvislabel}{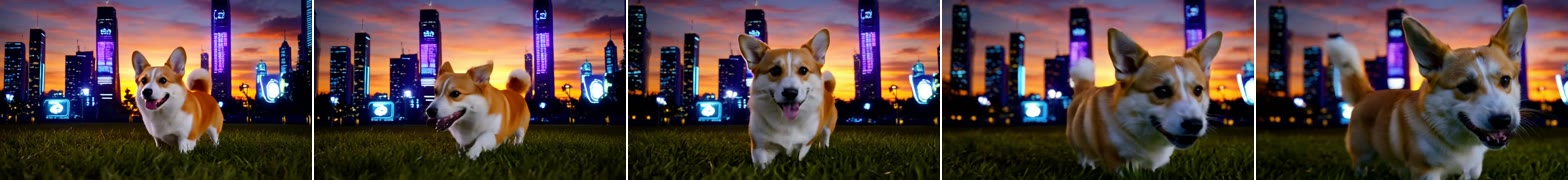}
\visrow{\cfvislabel}{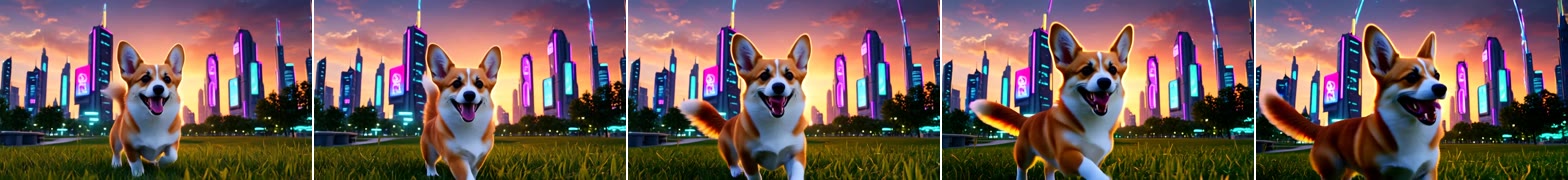}
\visrow{\cfunstepvislabel}{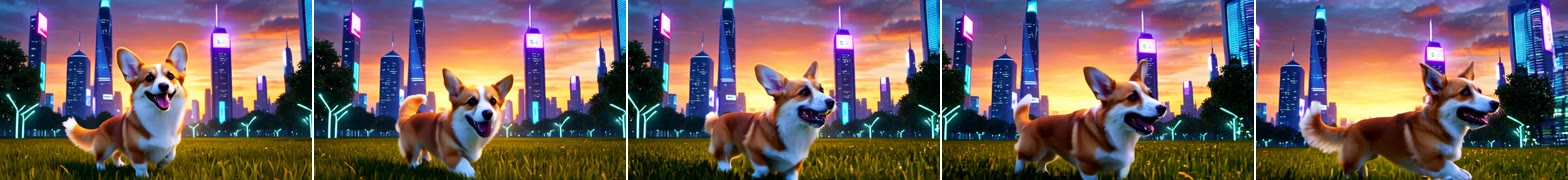}
\visprompt{A cute happy Corgi playing in park, sunset, in cyberpunk style}
\end{minipage}
}{
\begin{minipage}{\linewidth}
\centering
\visrow{\sfvislabel}{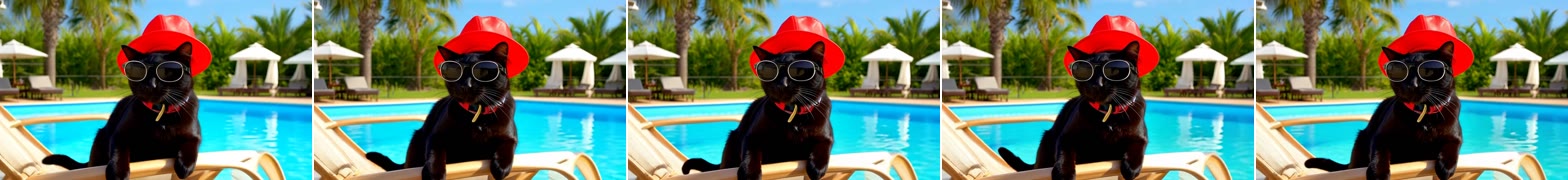}
\visrow{\sfunstepvislabel}{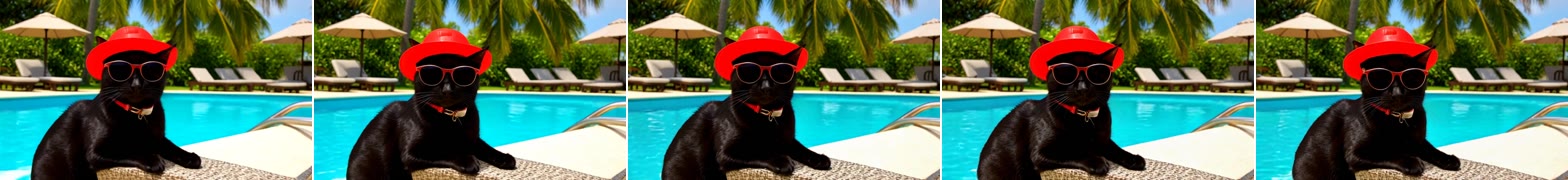}
\visrow{\cfvislabel}{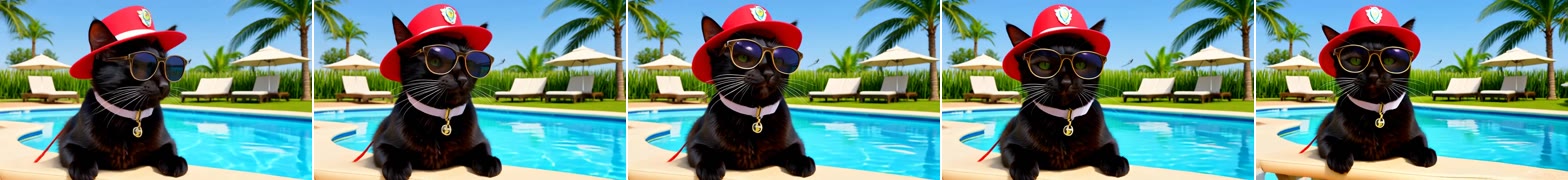}
\visrow{\cfunstepvislabel}{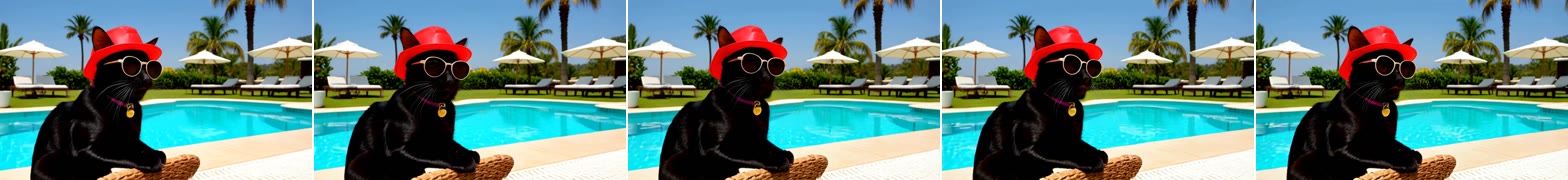}
\visprompt{A cat wearing sunglasses and working as a lifeguard at a pool.}
\end{minipage}
}{
\begin{minipage}{\linewidth}
\centering
\visrow{\sfvislabel}{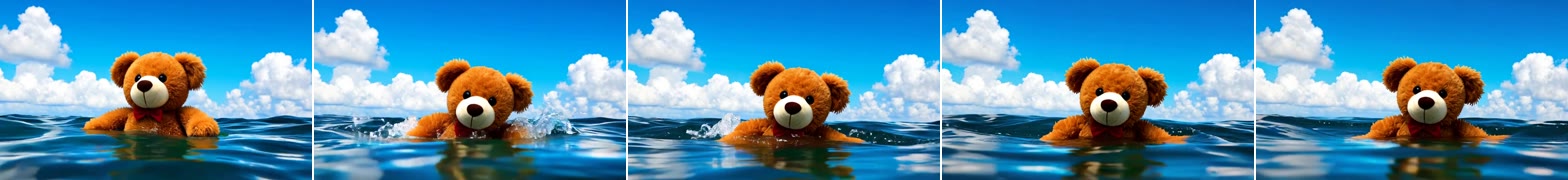}
\visrow{\sfunstepvislabel}{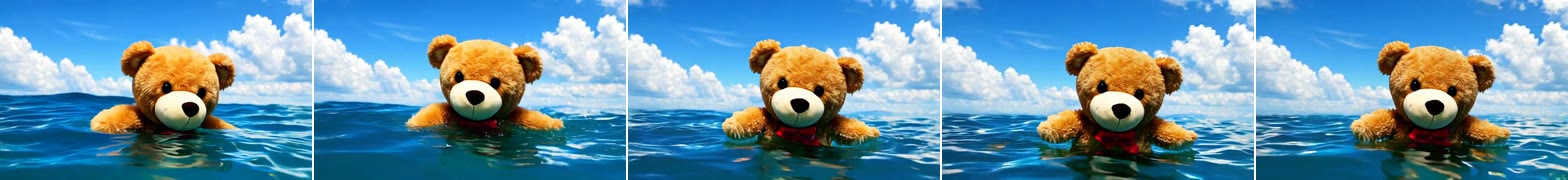}
\visrow{\cfvislabel}{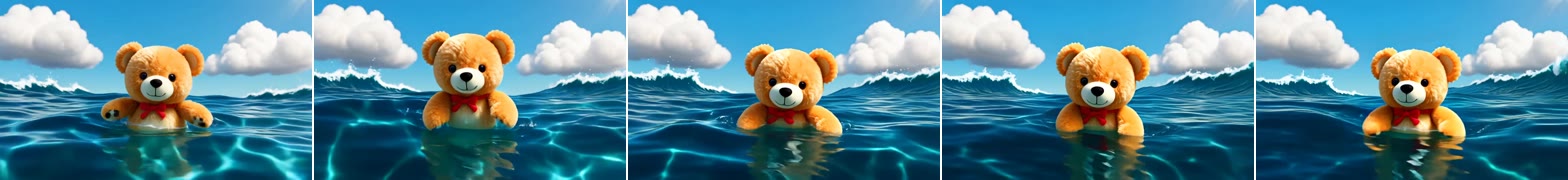}
\visrow{\cfunstepvislabel}{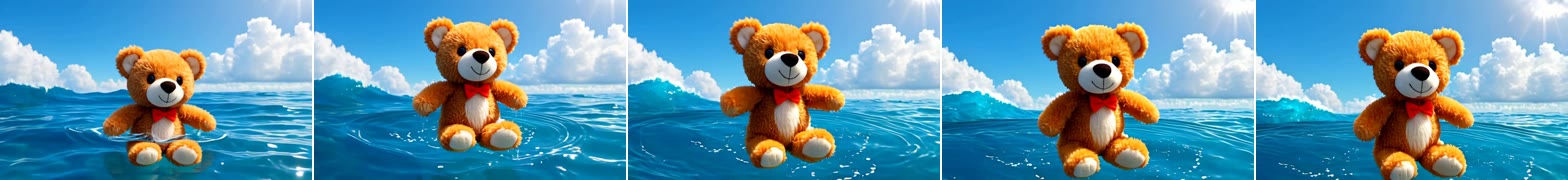}
\visprompt{a teddy bear is swimming in the ocean.}
\end{minipage}
}
\end{figure}

\begin{figure}[p]
\centering
\vispage{
\begin{minipage}{\linewidth}
\centering
\vistimes
\visrow{\sfvislabel}{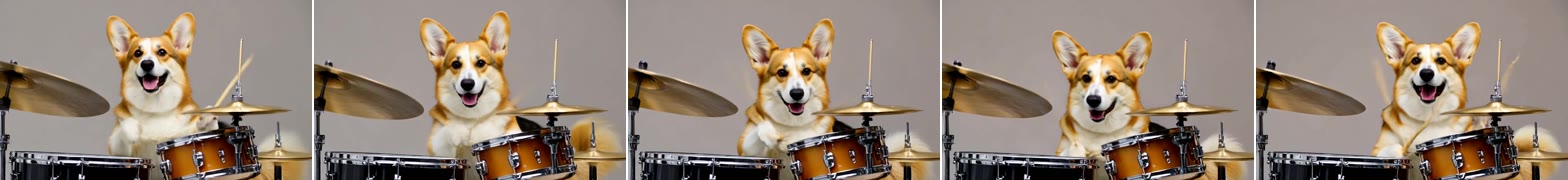}
\visrow{\sfunstepvislabel}{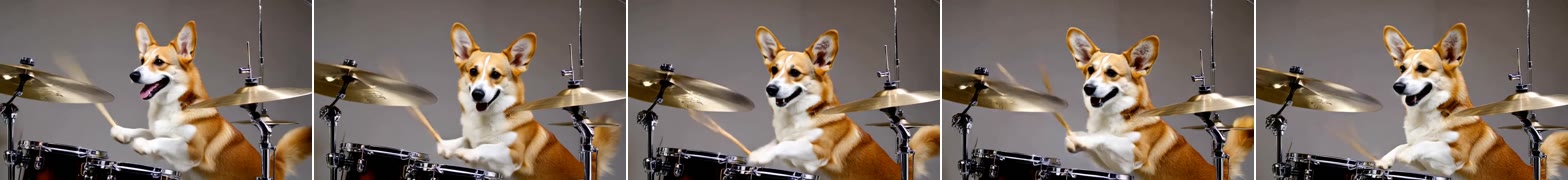}
\visrow{\cfvislabel}{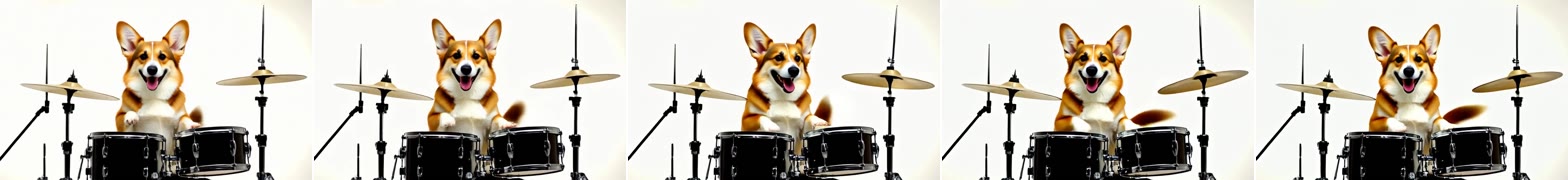}
\visrow{\cfunstepvislabel}{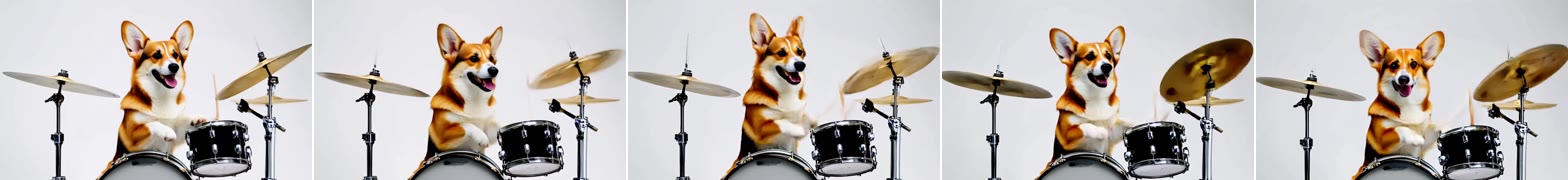}
\visprompt{A corgi is playing drum kit.}
\end{minipage}
}{
\begin{minipage}{\linewidth}
\centering
\visrow{\sfvislabel}{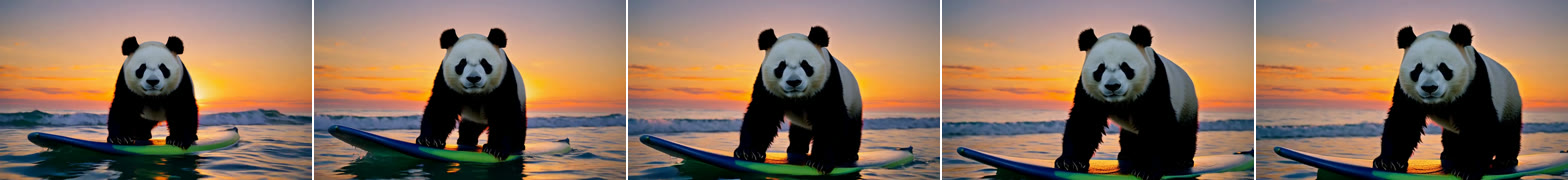}
\visrow{\sfunstepvislabel}{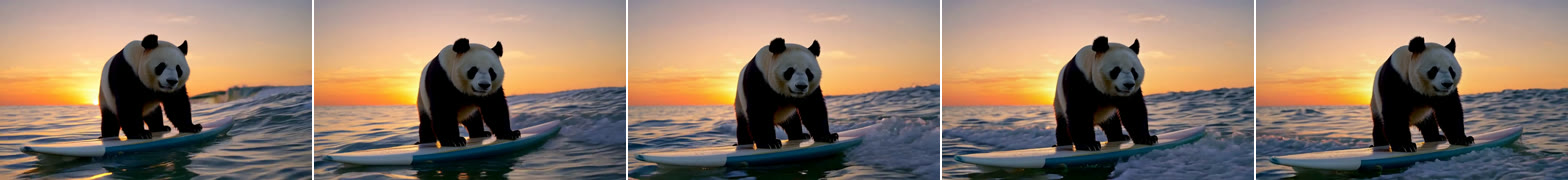}
\visrow{\cfvislabel}{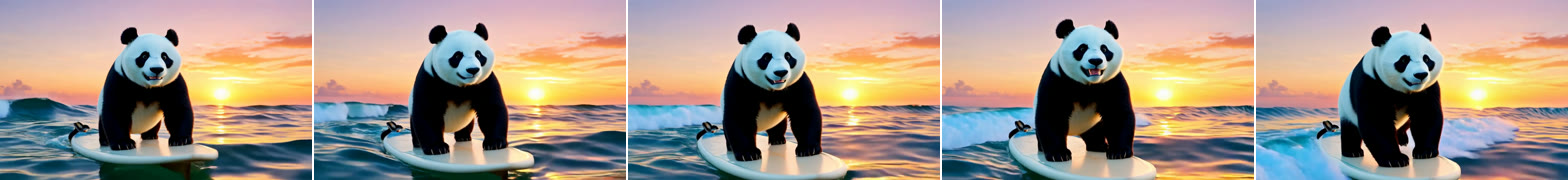}
\visrow{\cfunstepvislabel}{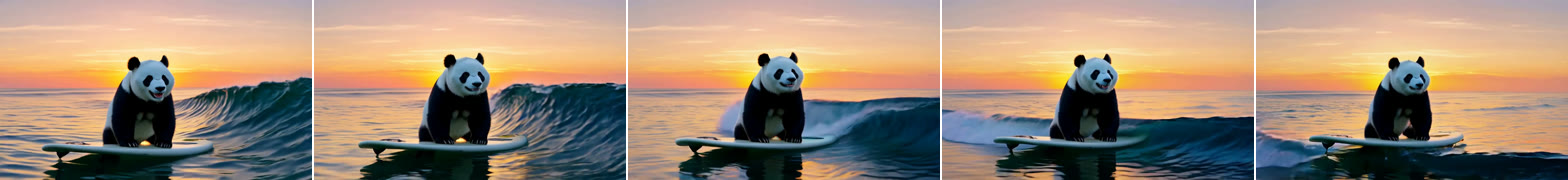}
\visprompt{A panda standing on a surfboard in the ocean in sunset.}
\end{minipage}
}{
\begin{minipage}{\linewidth}
\centering
\visrow{\sfvislabel}{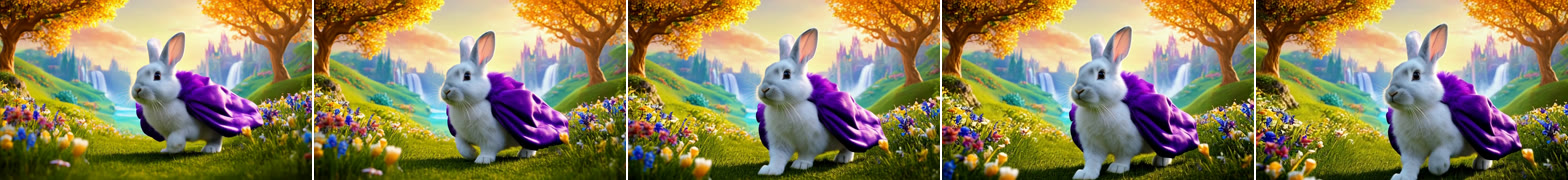}
\visrow{\sfunstepvislabel}{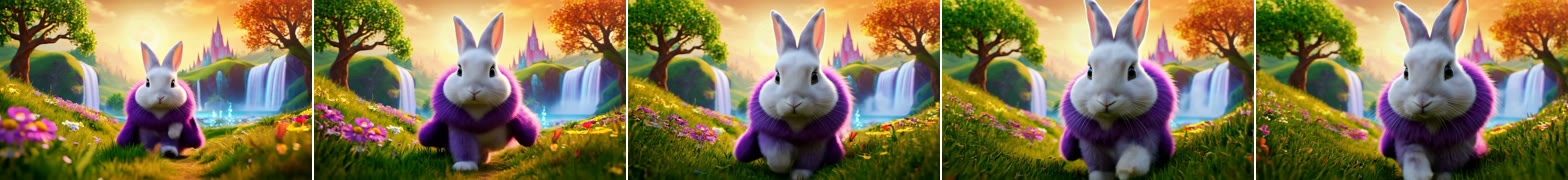}
\visrow{\cfvislabel}{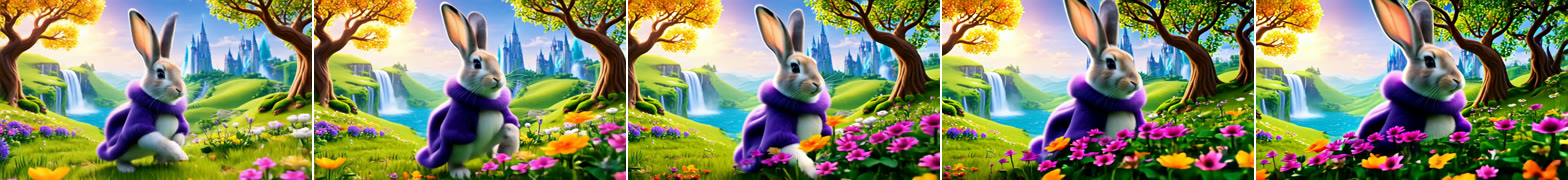}
\visrow{\cfunstepvislabel}{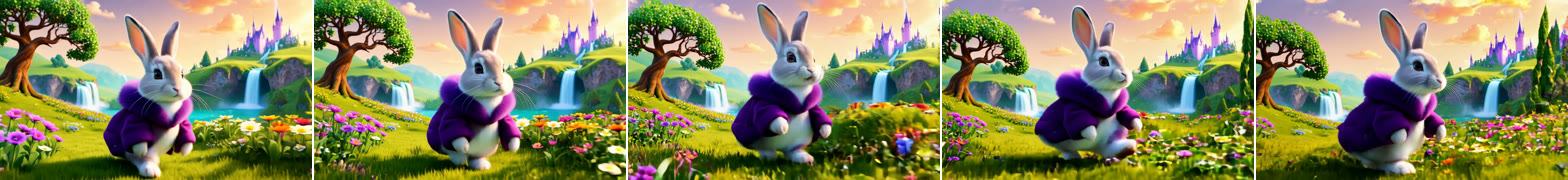}
\visprompt{A fat rabbit wearing a purple robe walking through a fantasy landscape.}
\end{minipage}
}
\end{figure}

\begin{figure}[p]
\centering
\vispage{
\begin{minipage}{\linewidth}
\centering
\vistimes
\visrow{\sfvislabel}{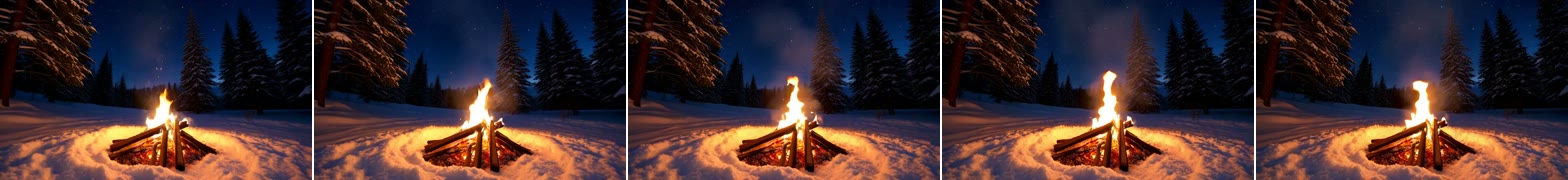}
\visrow{\sfunstepvislabel}{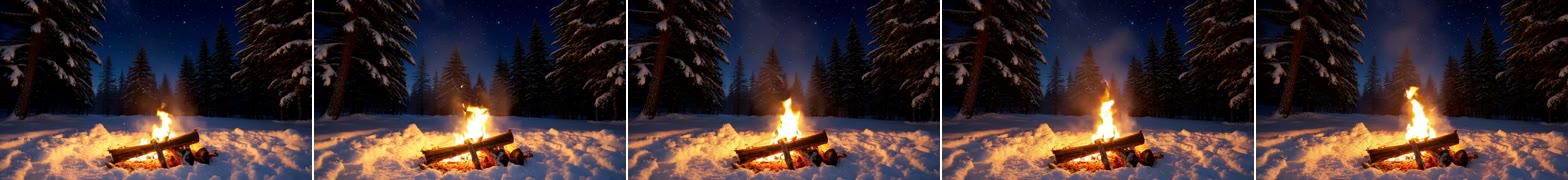}
\visrow{\cfvislabel}{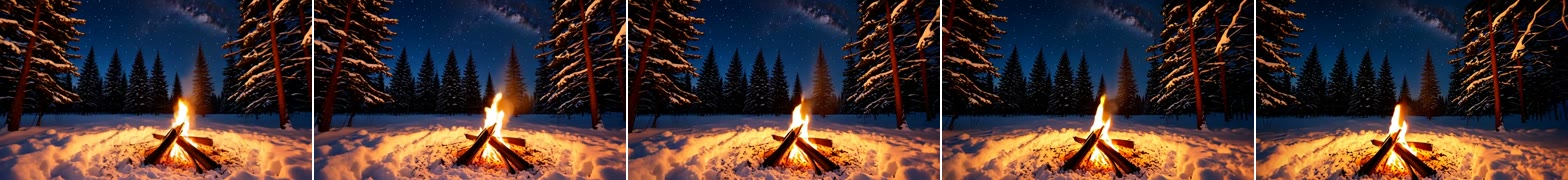}
\visrow{\cfunstepvislabel}{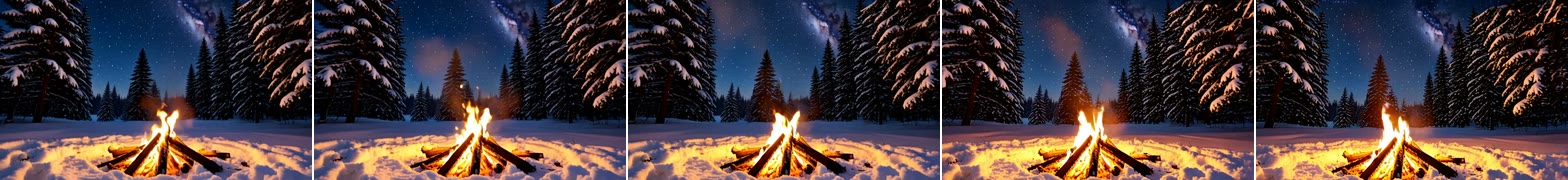}
\visprompt{Campfire at night in a snowy forest with starry sky in the background.}
\end{minipage}
}{
\begin{minipage}{\linewidth}
\centering
\visrow{\sfvislabel}{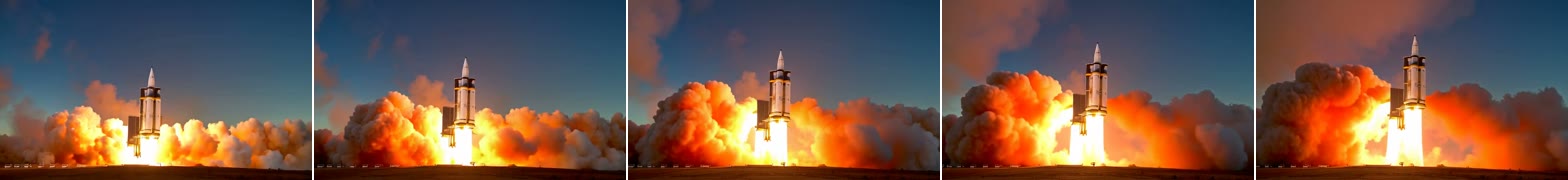}
\visrow{\sfunstepvislabel}{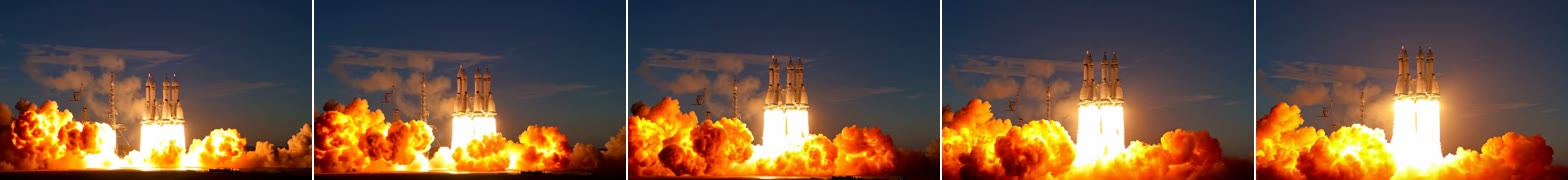}
\visrow{\cfvislabel}{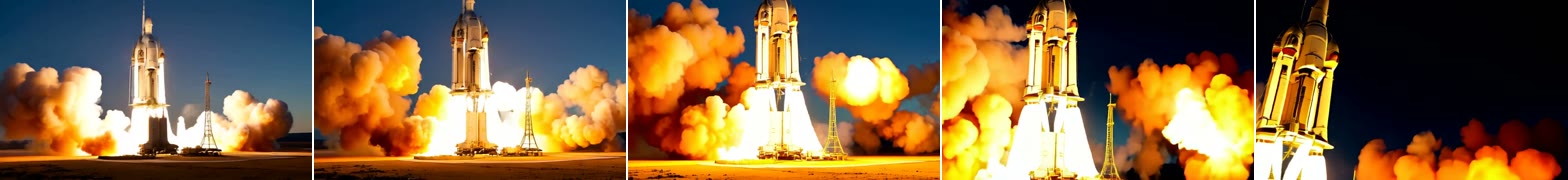}
\visrow{\cfunstepvislabel}{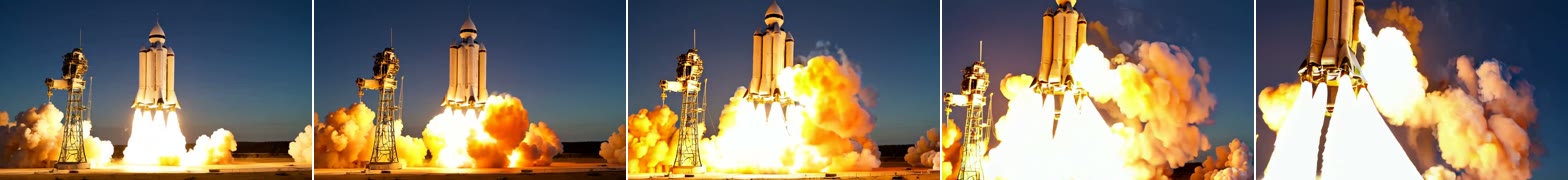}
\visprompt{A space shuttle launching into orbit, with flames and smoke billowing out from the engines}
\end{minipage}
}{
\begin{minipage}{\linewidth}
\centering
\visrow{\sfvislabel}{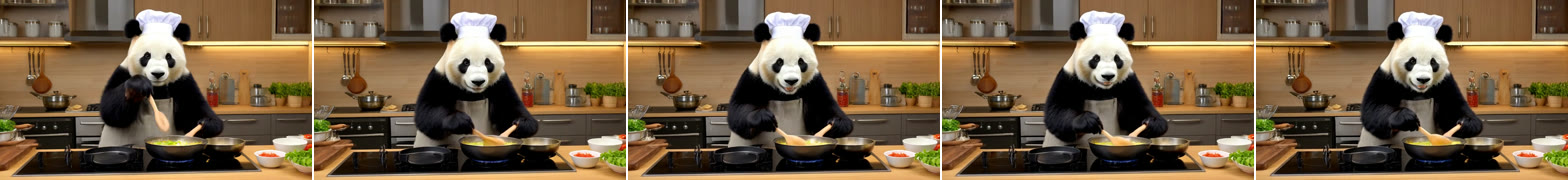}
\visrow{\sfunstepvislabel}{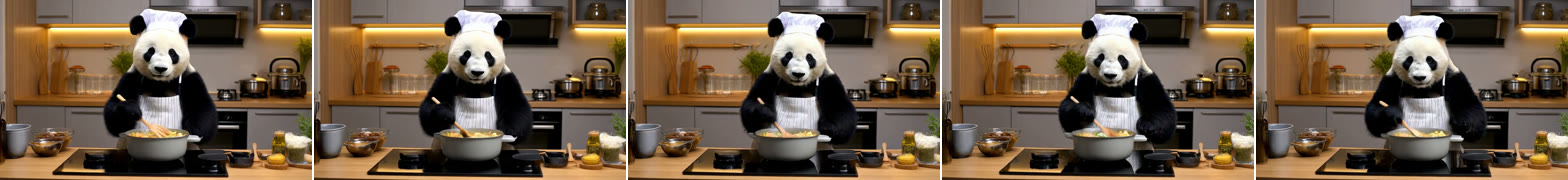}
\visrow{\cfvislabel}{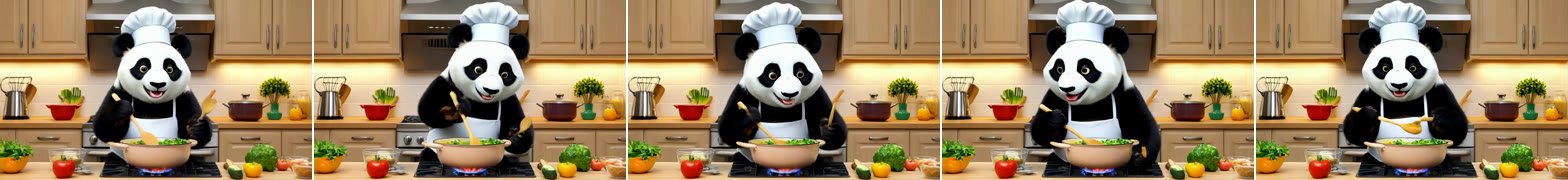}
\visrow{\cfunstepvislabel}{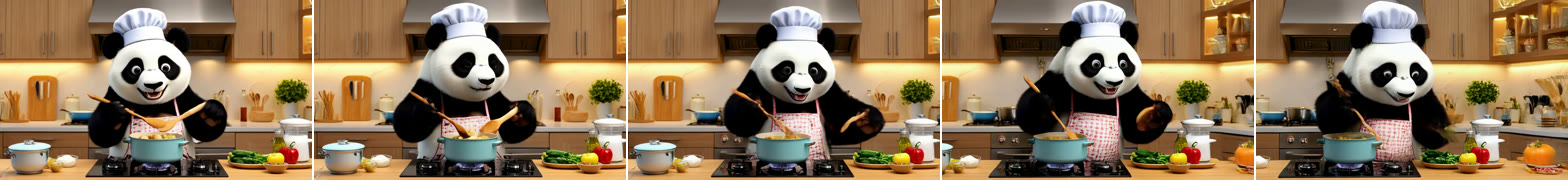}
\visprompt{A panda cooking in the kitchen}
\end{minipage}
}
\end{figure}

\begin{figure}[p]
\centering
\vispage{
\begin{minipage}{\linewidth}
\centering
\vistimes
\visrow{\sfvislabel}{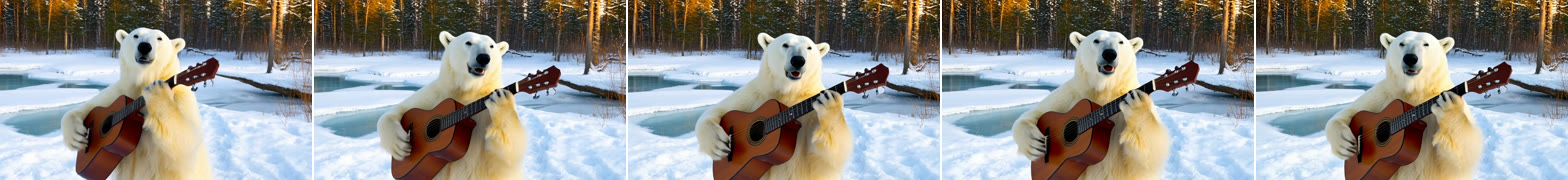}
\visrow{\sfunstepvislabel}{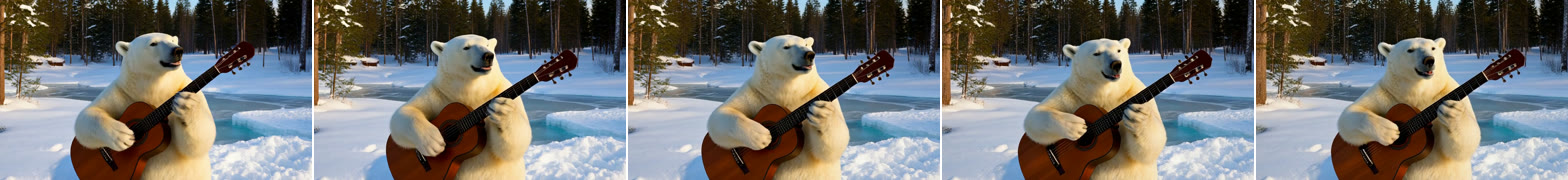}
\visrow{\cfvislabel}{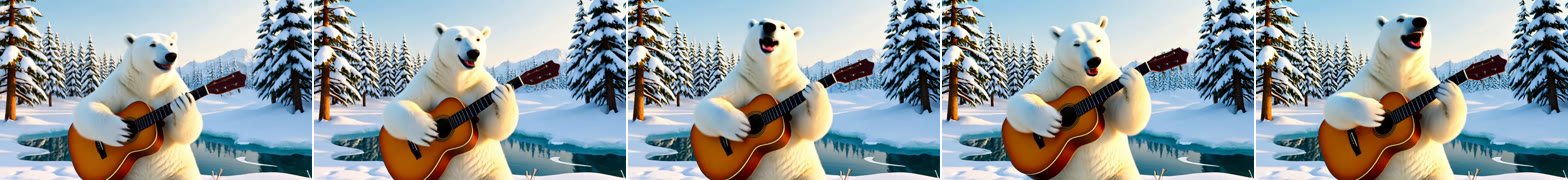}
\visrow{\cfunstepvislabel}{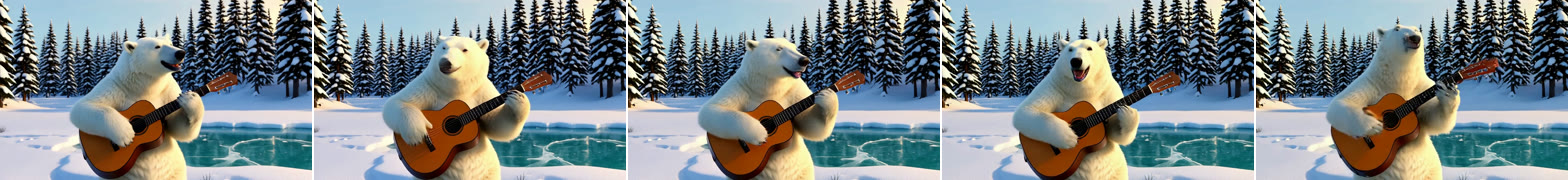}
\visprompt{A polar bear is playing guitar}
\end{minipage}
}{
\begin{minipage}{\linewidth}
\centering
\visrow{\sfvislabel}{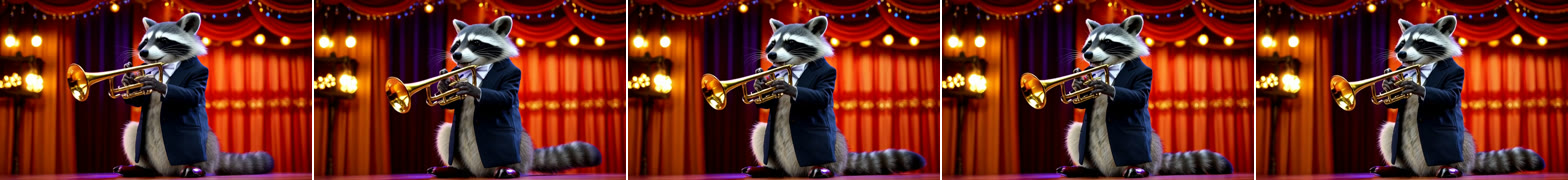}
\visrow{\sfunstepvislabel}{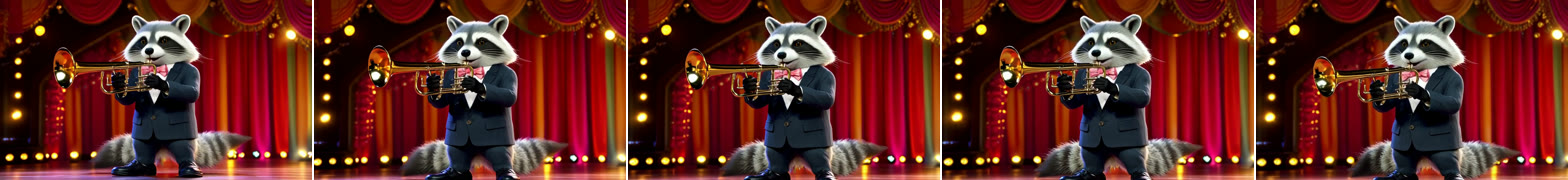}
\visrow{\cfvislabel}{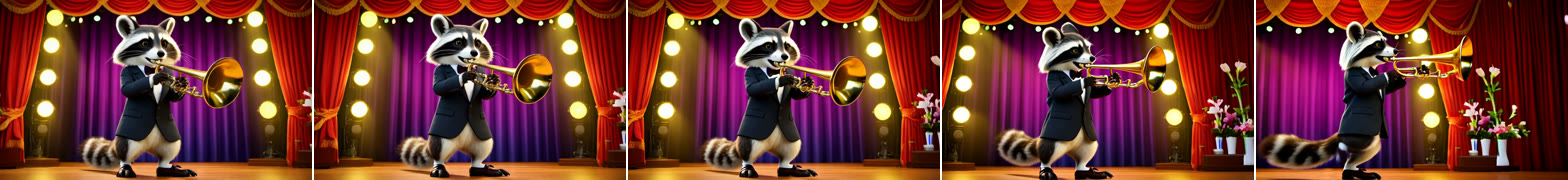}
\visrow{\cfunstepvislabel}{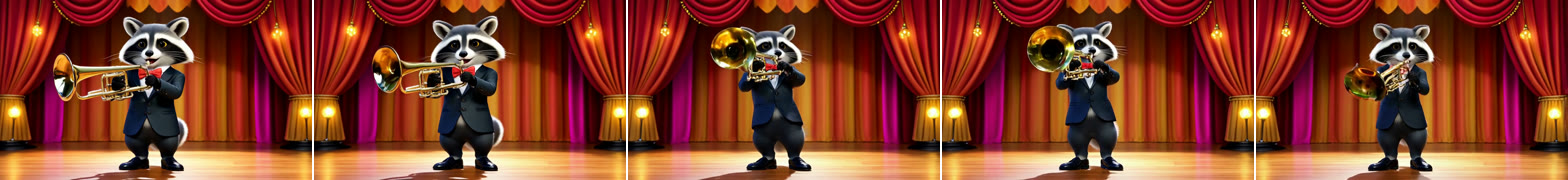}
\visprompt{A raccoon dressed in suit playing the trumpet, stage background}
\end{minipage}
}{
\begin{minipage}{\linewidth}
\centering
\visrow{\sfvislabel}{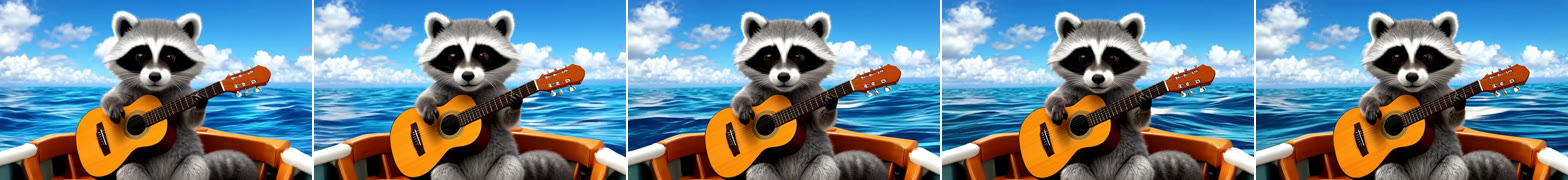}
\visrow{\sfunstepvislabel}{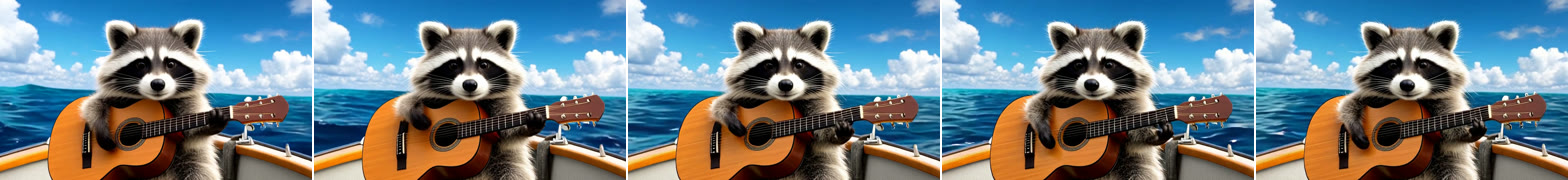}
\visrow{\cfvislabel}{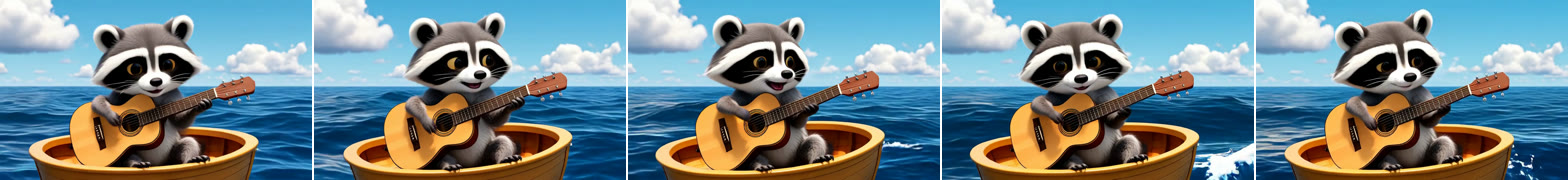}
\visrow{\cfunstepvislabel}{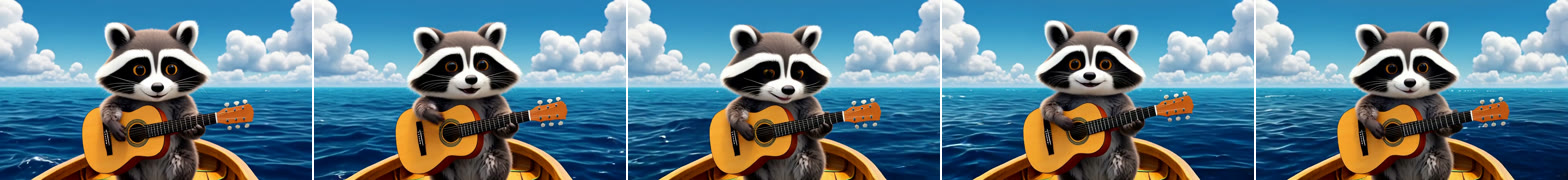}
\visprompt{A cute raccoon playing guitar in a boat on the ocean}
\end{minipage}
}
\end{figure}

\begin{figure}[p]
\centering
\vispage{
\begin{minipage}{\linewidth}
\centering
\vistimes
\visrow{\sfvislabel}{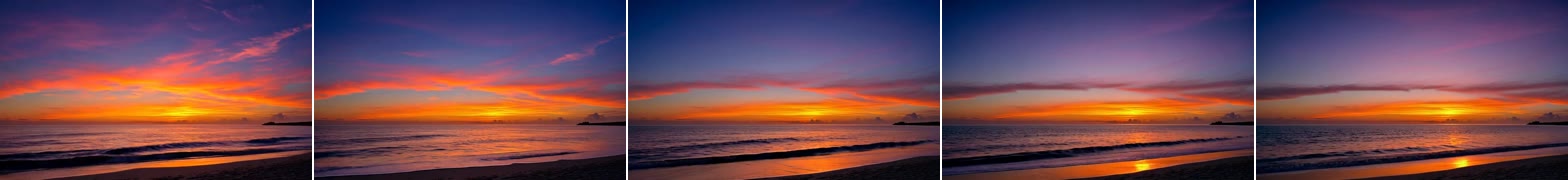}
\visrow{\sfunstepvislabel}{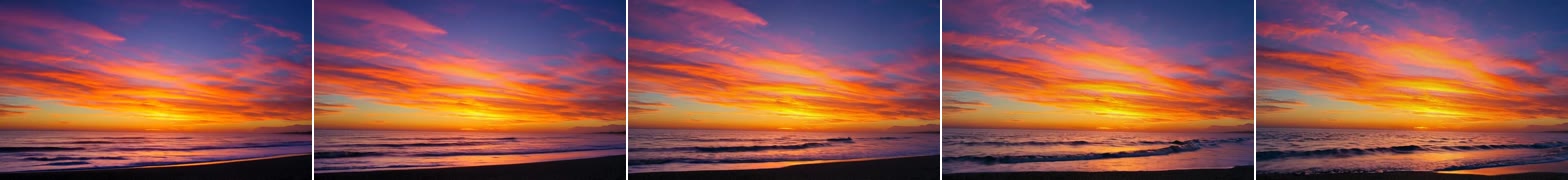}
\visrow{\cfvislabel}{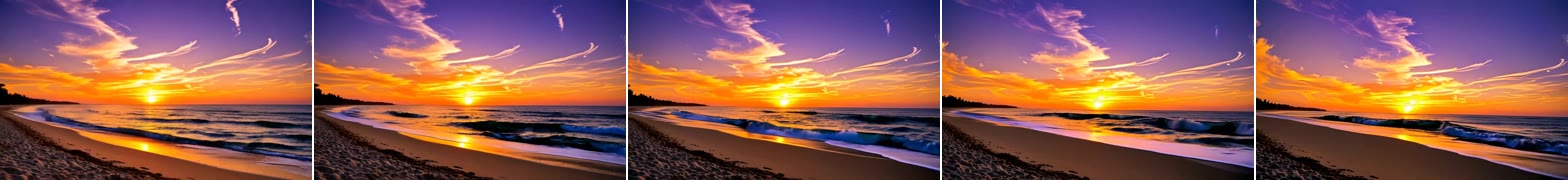}
\visrow{\cfunstepvislabel}{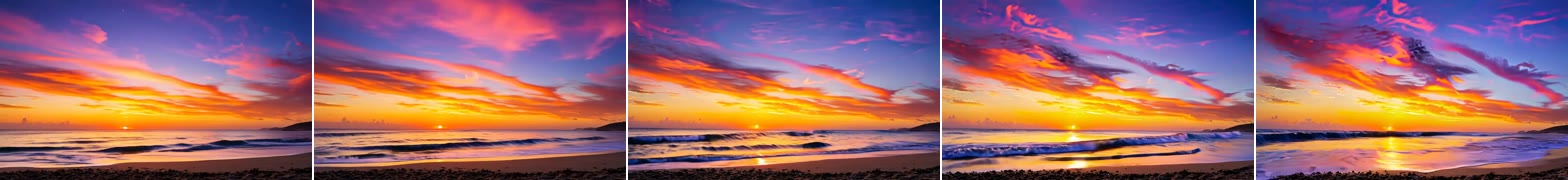}
\visprompt{Sunset time lapse at the beach with moving clouds and colors in the sky.}
\end{minipage}
}{
\begin{minipage}{\linewidth}
\centering
\visrow{\sfvislabel}{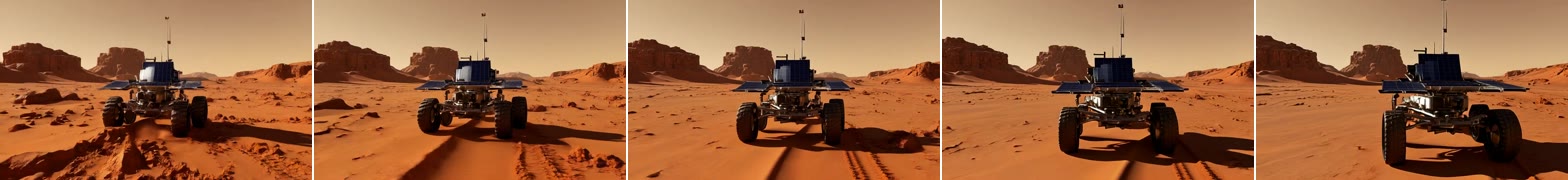}
\visrow{\sfunstepvislabel}{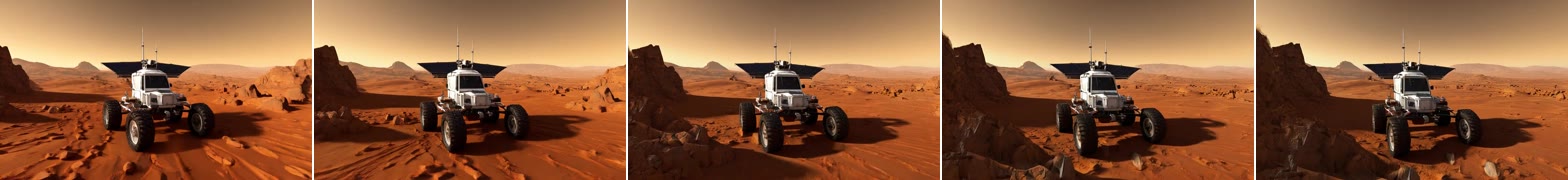}
\visrow{\cfvislabel}{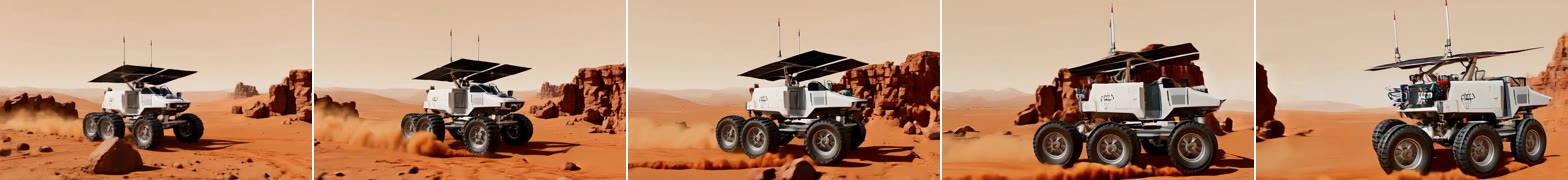}
\visrow{\cfunstepvislabel}{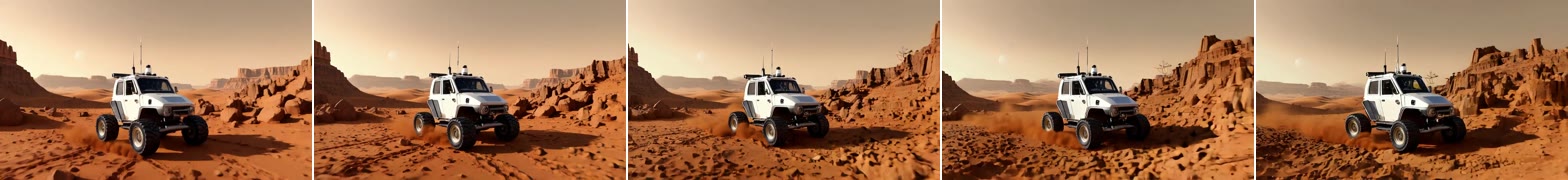}
\visprompt{A Mars rover moving on Mars}
\end{minipage}
}{
\begin{minipage}{\linewidth}
\centering
\visrow{\sfvislabel}{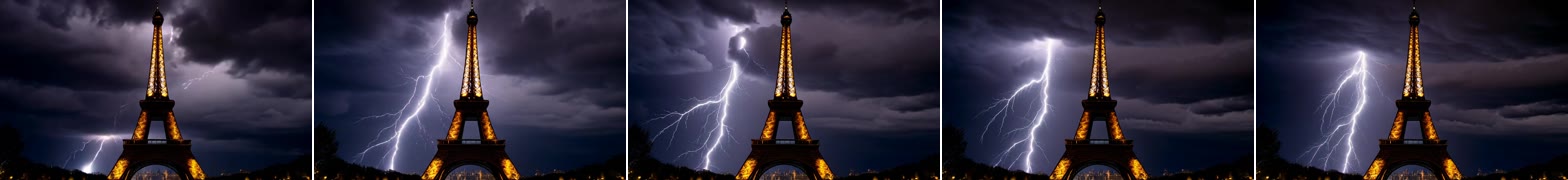}
\visrow{\sfunstepvislabel}{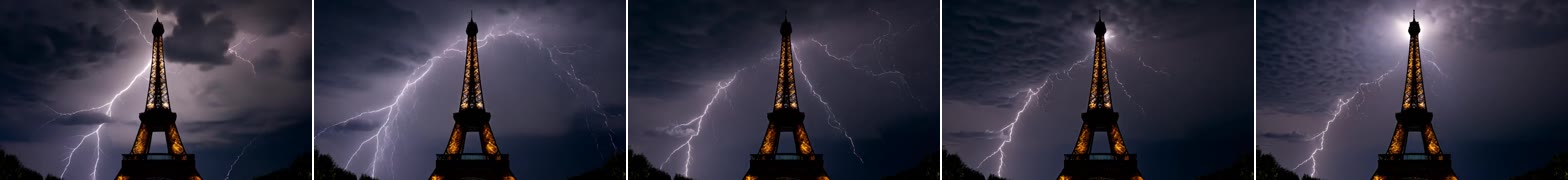}
\visrow{\cfvislabel}{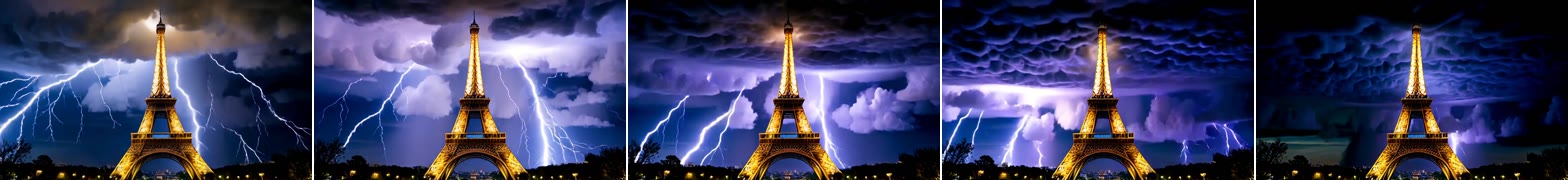}
\visrow{\cfunstepvislabel}{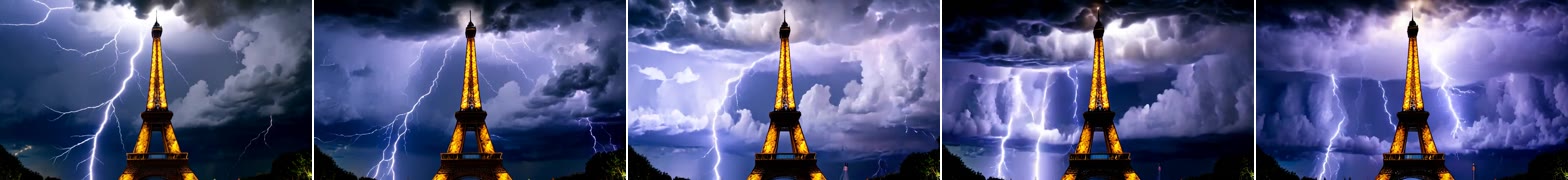}
\visprompt{A lightning striking atop of eiffel tower, dark clouds in the sky}
\end{minipage}
}
\end{figure}

\begin{figure}[p]
\centering
\vispage{
\begin{minipage}{\linewidth}
\centering
\vistimes
\visrow{\sfvislabel}{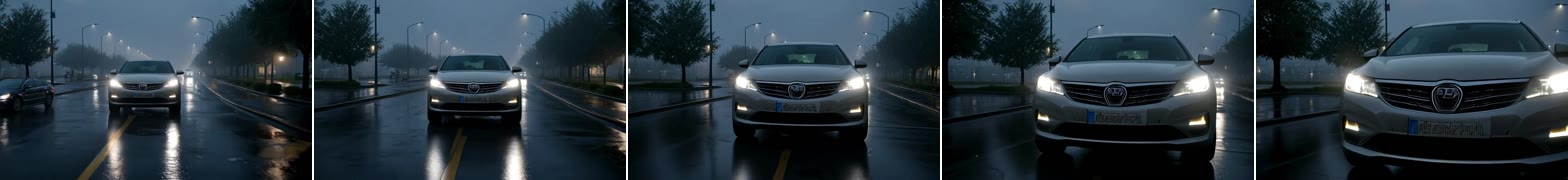}
\visrow{\sfunstepvislabel}{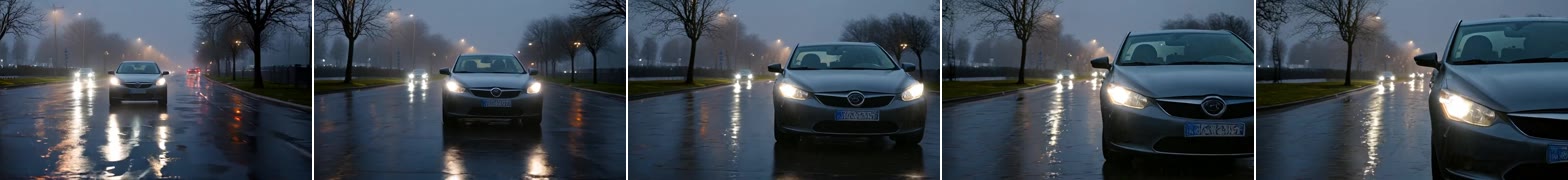}
\visrow{\cfvislabel}{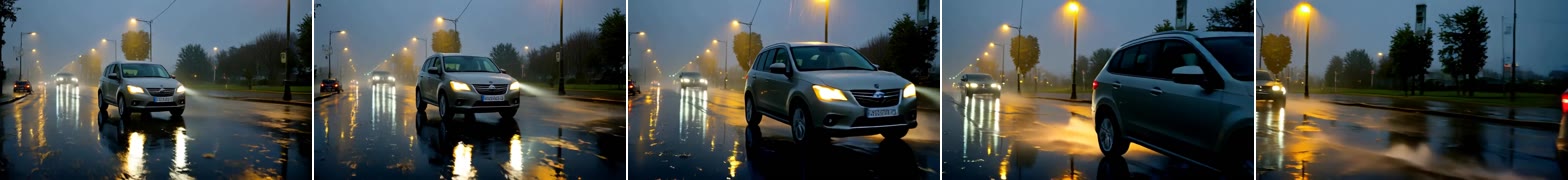}
\visrow{\cfunstepvislabel}{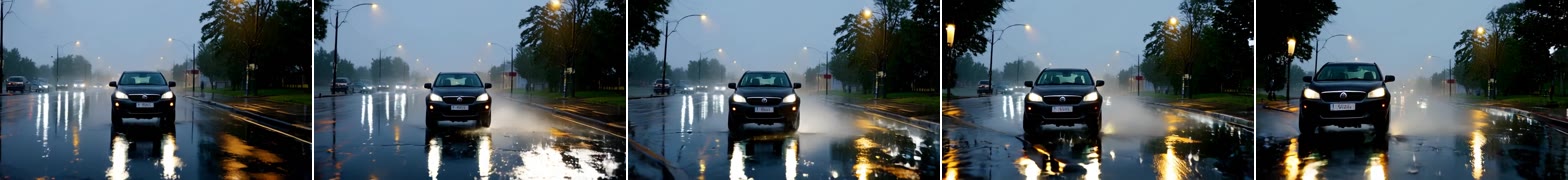}
\visprompt{A car moving slowly on an empty street, rainy evening}
\end{minipage}
}{
\begin{minipage}{\linewidth}
\centering
\visrow{\sfvislabel}{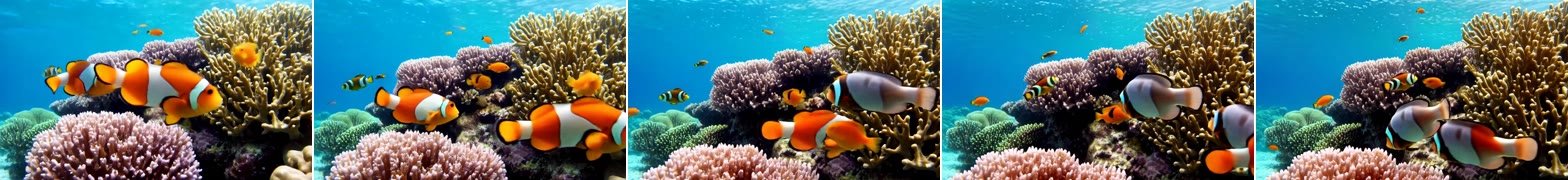}
\visrow{\sfunstepvislabel}{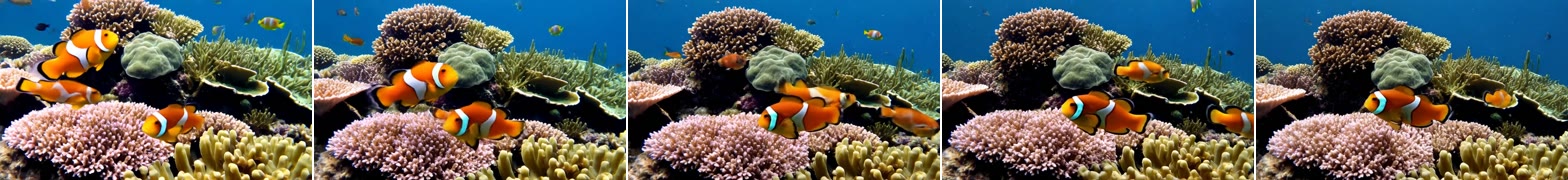}
\visrow{\cfvislabel}{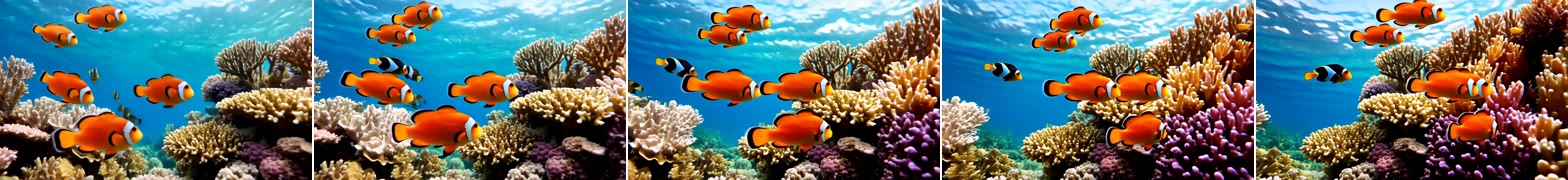}
\visrow{\cfunstepvislabel}{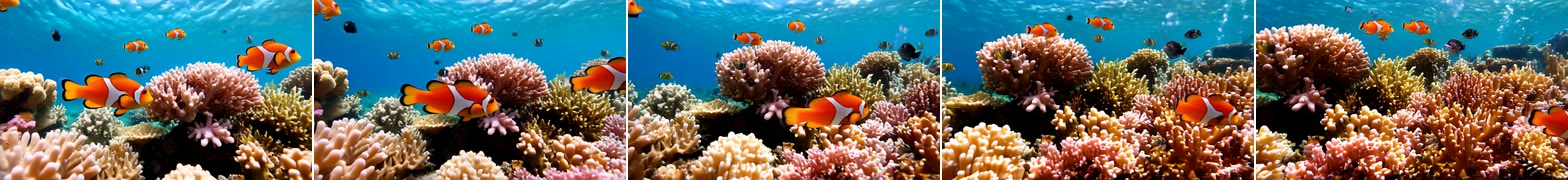}
\visprompt{Clown fish swimming through the coral reef}
\end{minipage}
}{
\begin{minipage}{\linewidth}
\centering
\visrow{\sfvislabel}{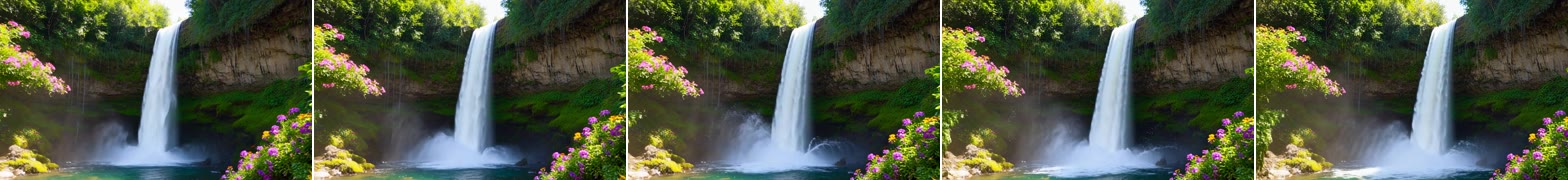}
\visrow{\sfunstepvislabel}{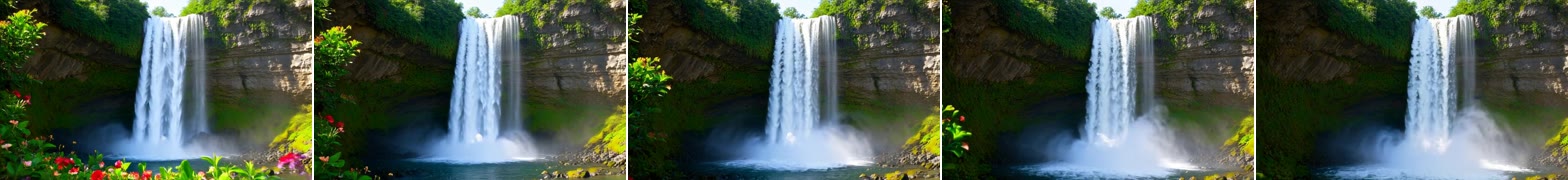}
\visrow{\cfvislabel}{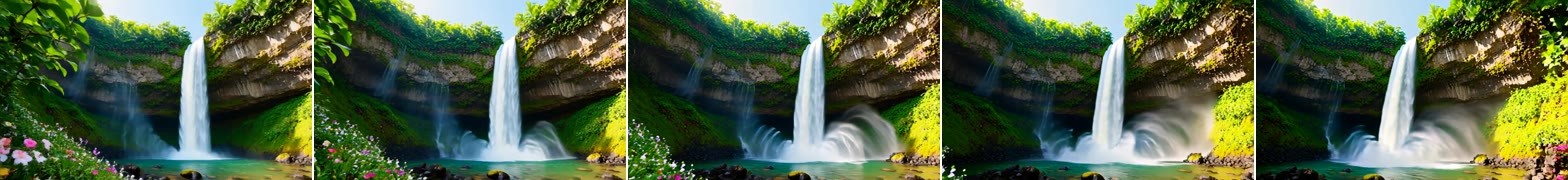}
\visrow{\cfunstepvislabel}{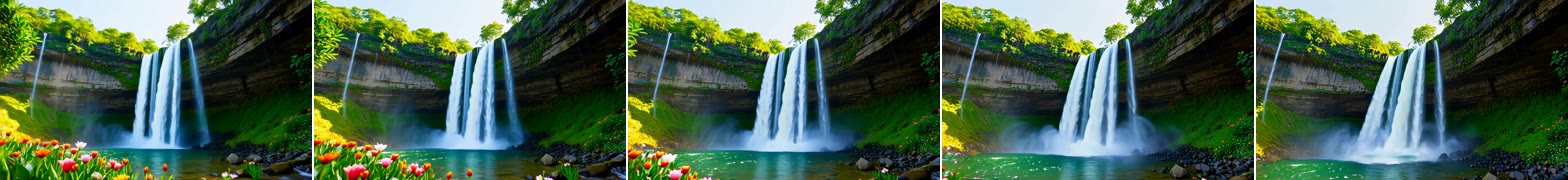}
\visprompt{waterfall}
\end{minipage}
}
\end{figure}

\clearpage
\newcommand{\iTwoVTimes}{%
  \noindent\makebox[\linewidth][c]{%
    \hspace*{\vislabelwidth}\hspace{0.004\linewidth}%
    \makebox[\visframewidth][c]{\scriptsize Input image}\hspace{0.002\linewidth}%
    \makebox[\visframewidth][c]{\scriptsize 1.25 s}\hspace{0.002\linewidth}%
    \makebox[\visframewidth][c]{\scriptsize 2.5 s}\hspace{0.002\linewidth}%
    \makebox[\visframewidth][c]{\scriptsize 3.75 s}\hspace{0.002\linewidth}%
    \makebox[\visframewidth][c]{\scriptsize 5 s}%
  }\par\vspace{0.4pt}%
}

\newcommand{\iTwoVPrompt}[1]{%
  \vspace{2pt}%
  {\small\textit{#1}}\par%
}

\begin{figure}[p]
\centering
\setlength{\visframewidth}{0.93\visframewidth}
\setlength{\visframeheight}{0.93\visframeheight}
\setlength{\visstripwidth}{0.93\visstripwidth}
\begin{tikzpicture}
  \path[use as bounding box] (0,0) rectangle (\linewidth,\textheight);
  \node[anchor=north west,inner sep=0pt,outer sep=0pt] (visintro) at (0,\textheight) {
    \begin{minipage}{\linewidth}
\subsection*{Image-to-video examples}

For image-to-video generation, all four variants receive the same reference image, VBench I2V prompt, and initial noise. The first column shows the supplied reference image, and the remaining columns show generated frames.
    \end{minipage}
  };
  \coordinate (vistop) at ([yshift=-6pt]visintro.south west);
  \node[anchor=north west,inner sep=0pt,outer sep=0pt] at (vistop) {
\begin{minipage}{\linewidth}
\centering
\iTwoVTimes
\visrow{\sfvislabel}{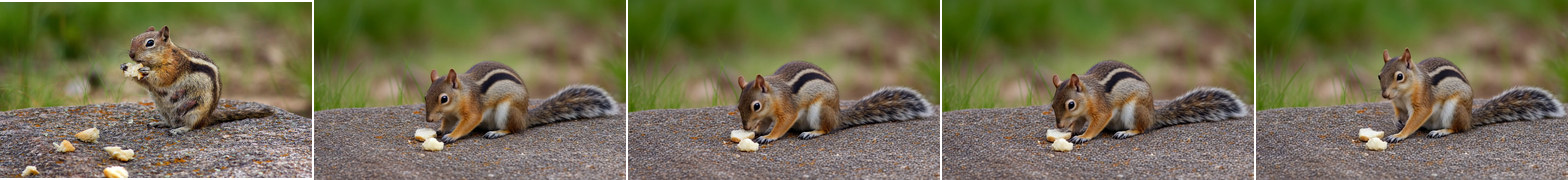}
\visrow{\sfunstepvislabel}{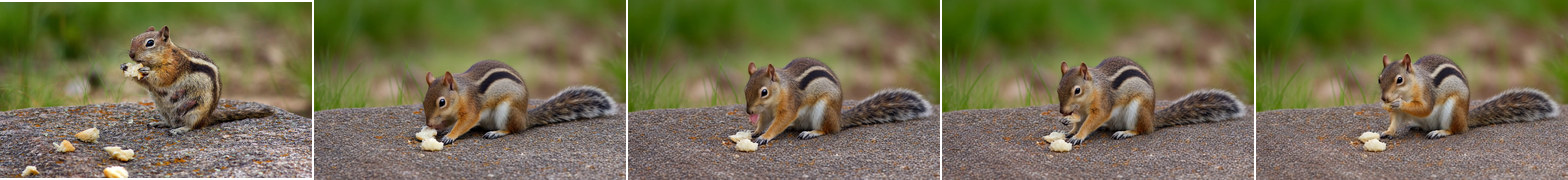}
\visrow{\cfvislabel}{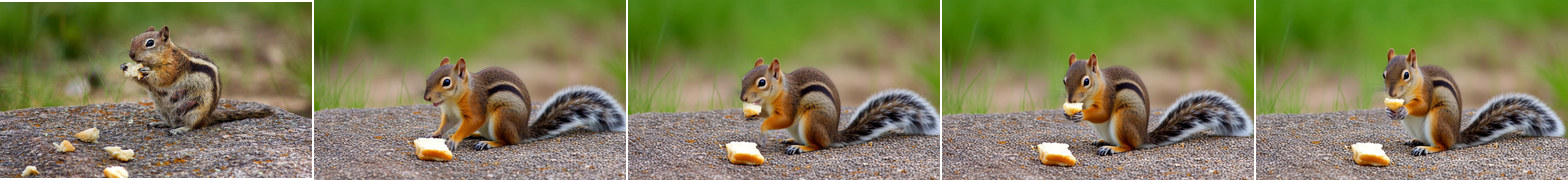}
\visrow{\cfunstepvislabel}{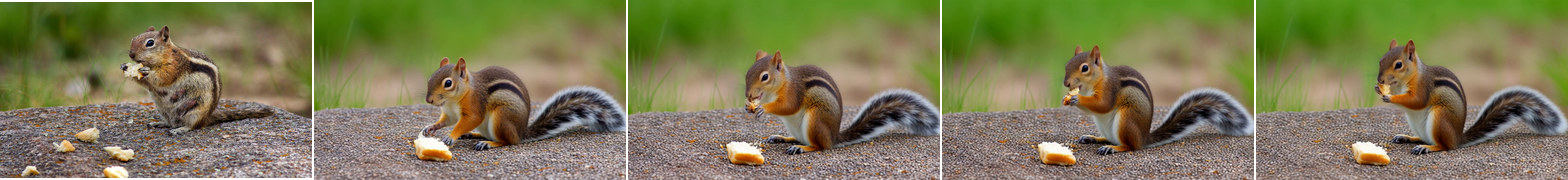}
\iTwoVPrompt{a squirrel sitting on the ground eating a piece of bread}
\end{minipage}
  };
  \node[anchor=center,inner sep=0pt,outer sep=0pt] at ($(vistop)!0.5!(\linewidth,0)$) {
\begin{minipage}{\linewidth}
\centering
\visrow{\sfvislabel}{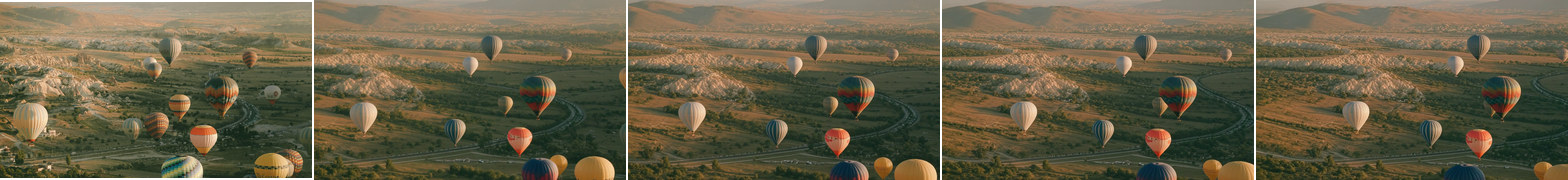}
\visrow{\sfunstepvislabel}{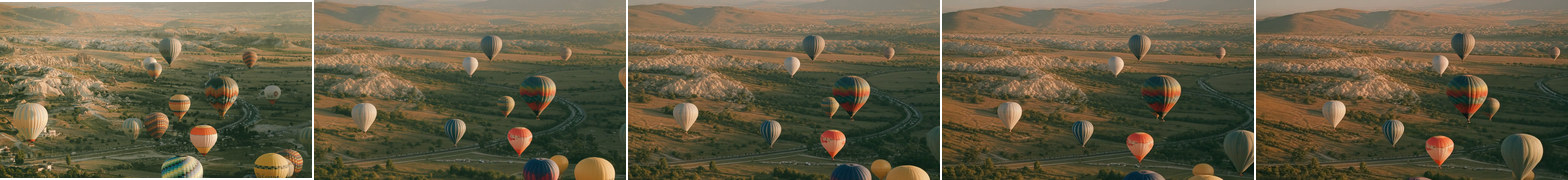}
\visrow{\cfvislabel}{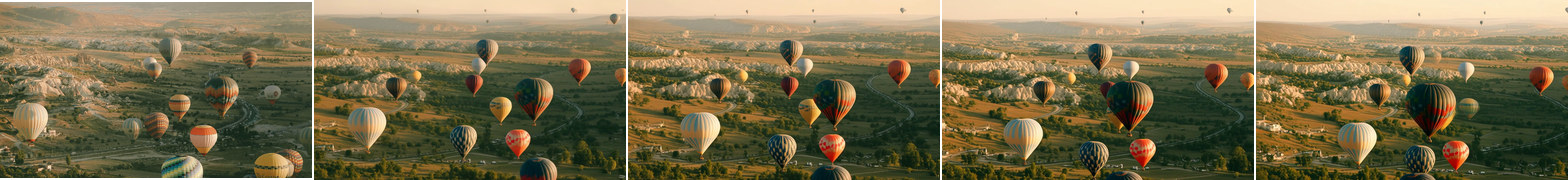}
\visrow{\cfunstepvislabel}{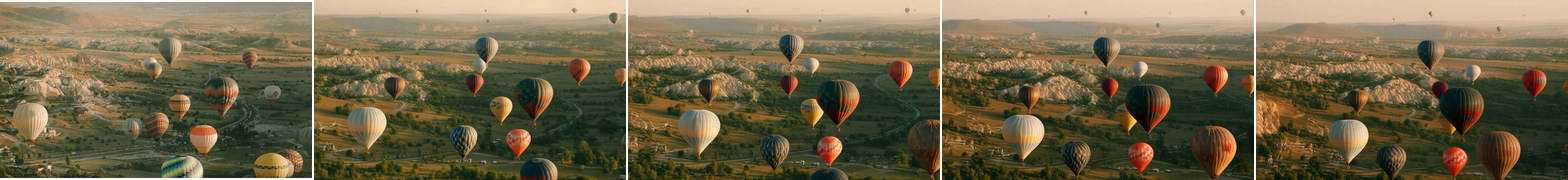}
\iTwoVPrompt{a group of hot air balloons flying over a valley}
\end{minipage}
  };
  \node[anchor=south west,inner sep=0pt,outer sep=0pt] at (0,0) {
\begin{minipage}{\linewidth}
\centering
\visrow{\sfvislabel}{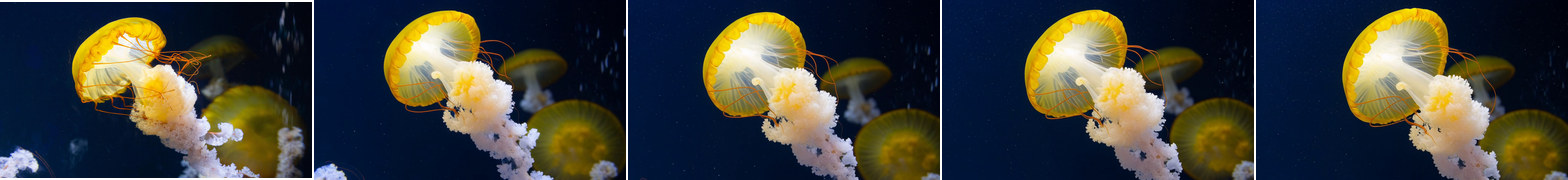}
\visrow{\sfunstepvislabel}{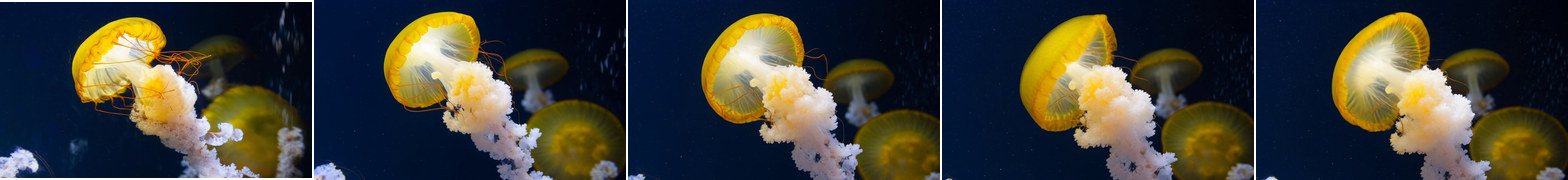}
\visrow{\cfvislabel}{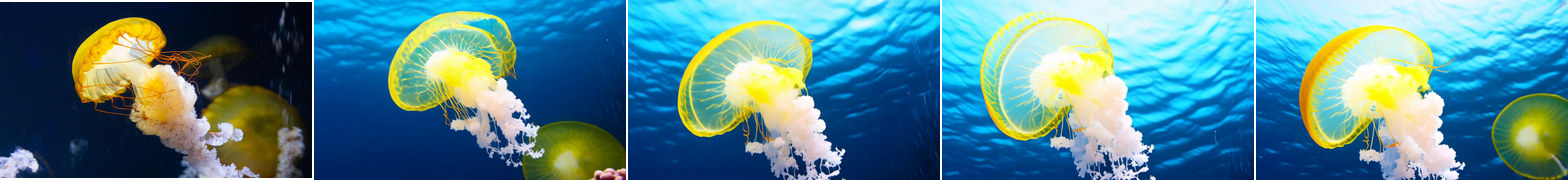}
\visrow{\cfunstepvislabel}{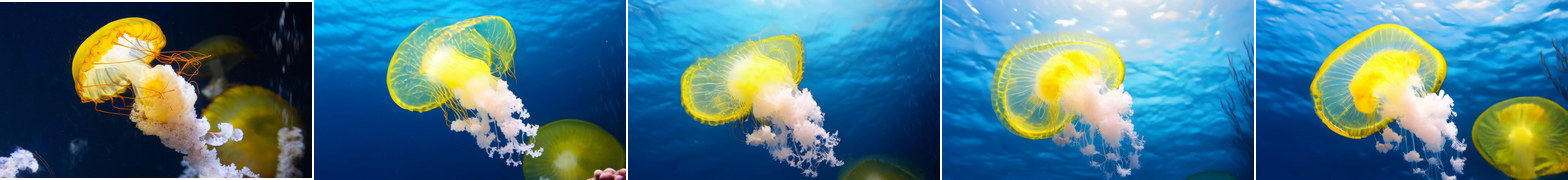}
\iTwoVPrompt{a yellow and white jellyfish is floating in the ocean}
\end{minipage}
  };
\end{tikzpicture}
\end{figure}

\begin{figure}[p]
\centering
\vispage{
\begin{minipage}{\linewidth}
\centering
\iTwoVTimes
\visrow{\sfvislabel}{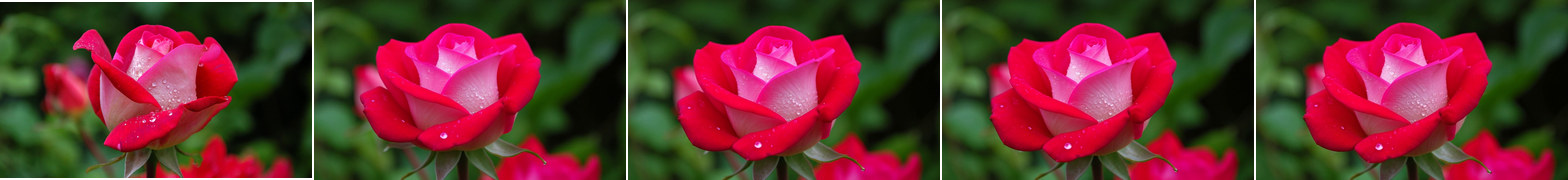}
\visrow{\sfunstepvislabel}{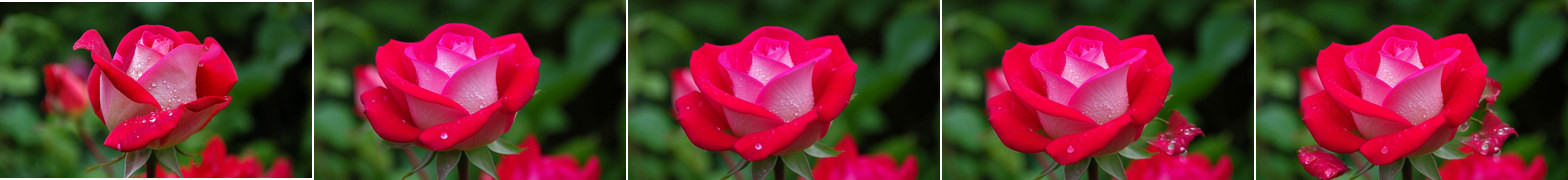}
\visrow{\cfvislabel}{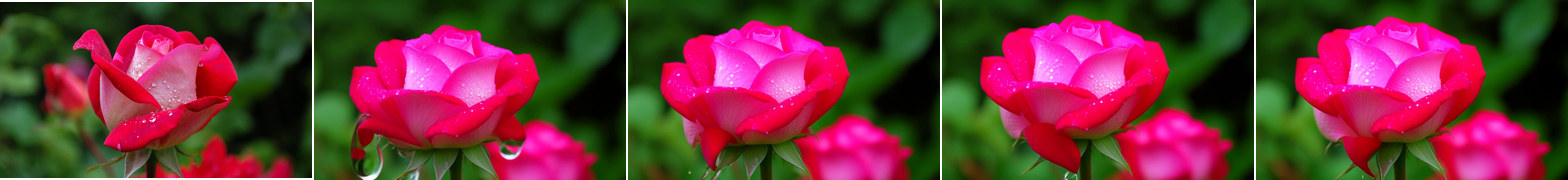}
\visrow{\cfunstepvislabel}{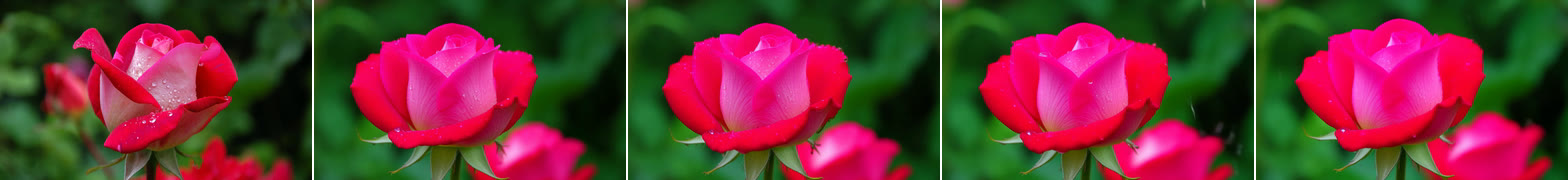}
\iTwoVPrompt{a close-up of a pink rose with water droplets on it}
\end{minipage}
}{
\begin{minipage}{\linewidth}
\centering
\visrow{\sfvislabel}{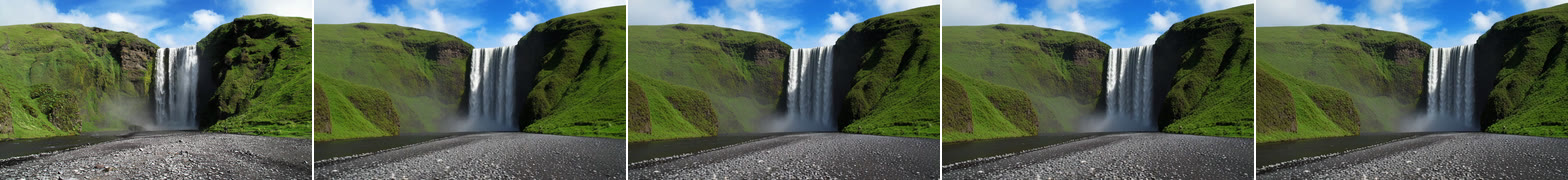}
\visrow{\sfunstepvislabel}{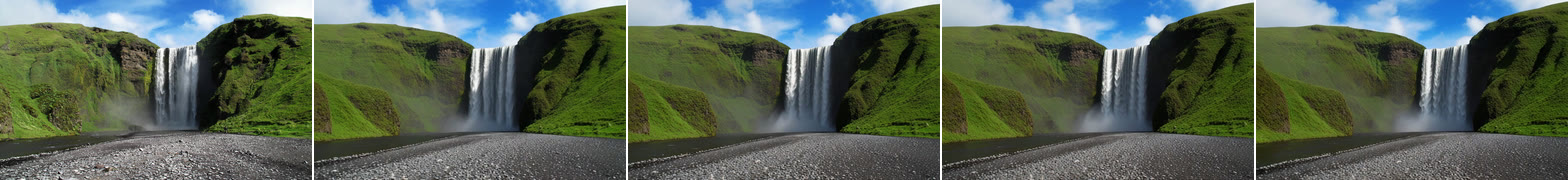}
\visrow{\cfvislabel}{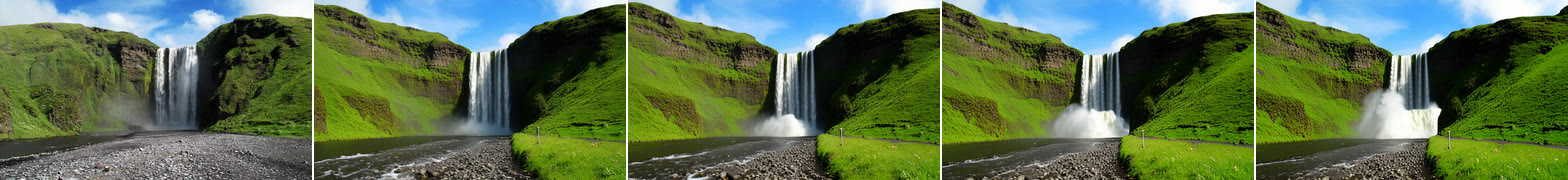}
\visrow{\cfunstepvislabel}{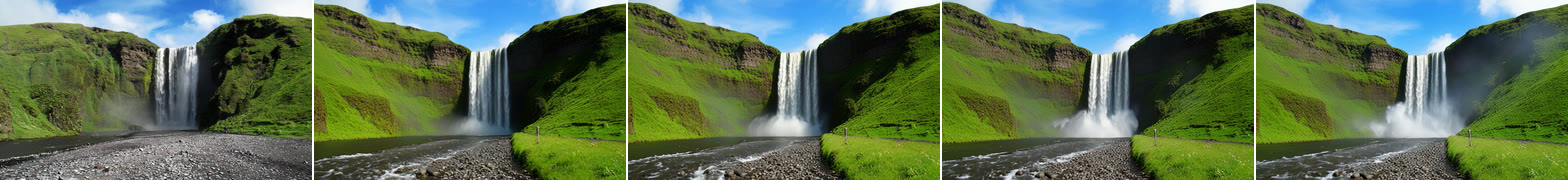}
\iTwoVPrompt{a large waterfall in the middle of a lush green hillside}
\end{minipage}
}{
\begin{minipage}{\linewidth}
\centering
\visrow{\sfvislabel}{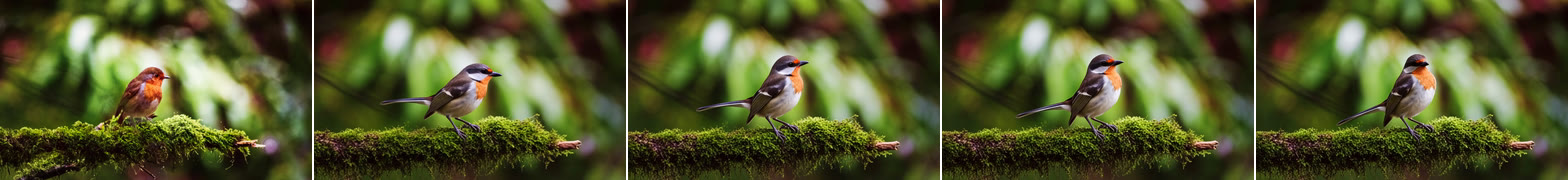}
\visrow{\sfunstepvislabel}{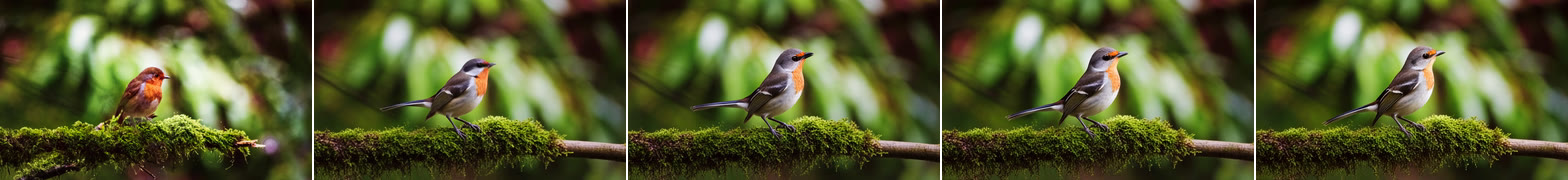}
\visrow{\cfvislabel}{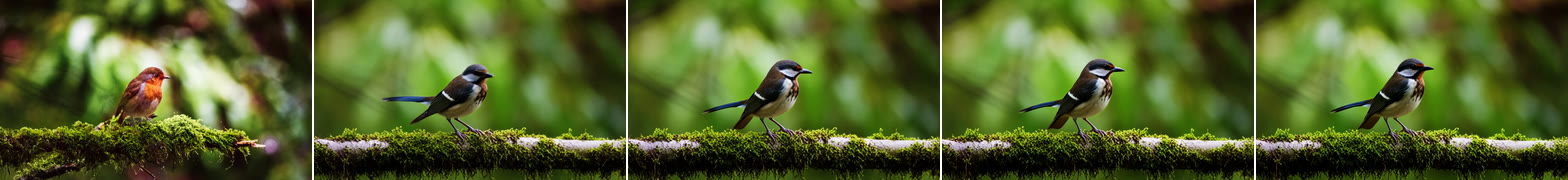}
\visrow{\cfunstepvislabel}{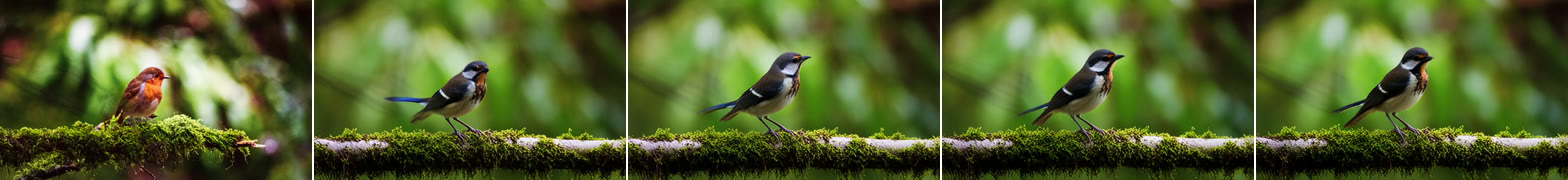}
\iTwoVPrompt{a small bird sits on a moss covered branch}
\end{minipage}
}
\end{figure}